\documentclass[review,10pt]{JMtemplate}
\usepackage{times}
\usepackage{graphicx}
\usepackage{bm}
\usepackage{paralist,algorithmic,algorithm}
\usepackage{amsmath,mathrsfs,amssymb}
\usepackage{url}            
\usepackage{amsfonts}       
\usepackage{nicefrac}       
\usepackage{microtype}      
\usepackage{booktabs}       
\usepackage{longtable}
\usepackage{threeparttable}
\usepackage{multirow}
\usepackage{soul}
\usepackage{subfigure}
\usepackage{wrapfig}
\usepackage{tikz}
\usetikzlibrary{fpu}
\usetikzlibrary{math}
\usepackage[american]{babel}
\usepackage{setspace}
\usepackage{stfloats}
\usepackage{lipsum}
\usepackage{multicol}
\usepackage{pifont}  

\usepackage{natbib}  
\usepackage{caption} 

\newcommand*{\dif}{\mathop{}\!\mathrm{d}}

\newcommand*{\diag}{\mathop{}\!\mathrm{diag}}

\newcommand*{\e}{\mathop{}\!\mathrm{e}}
\newcommand*{\ii}{\mathop{}\!\mathrm{i}}

\newtheorem{theorem}{Theorem}

\newtheorem{corollary}[theorem]{Corollary}
\newtheorem{proposition}{Proposition}

\begin{document}
\begin{frontmatter}
\title{On the Intrinsic Structures of Spiking Neural Networks}

\author{\textbf{Shao-Qun Zhang\textsuperscript{\rm 1}, 
		Jia-Yi Chen\textsuperscript{\rm 1},
		Jin-Hui Wu\textsuperscript{\rm 1}
		Gao Zhang\textsuperscript{\rm 2}, 
		Huan Xiong\textsuperscript{\rm 3}, 
		Bin Gu\textsuperscript{\rm 3},
		Zhi-Hua Zhou\textsuperscript{\rm 1}} \footnote{Zhi-Hua Zhou is the corresponding author.} \\~\\
	\textsuperscript{\rm 1}National Key Laboratory for Novel Software Technology, Nanjing University, China \\
	\texttt{\{zhangsq, chenjy, wujh, zhouzh\}@lamda.nju.edu.cn} \\
	\textsuperscript{\rm 2}Institute of Mathematics, Nanjing Normal University, China \\
	\texttt{gaozhang0810@hotmail.com} \\
	\textsuperscript{\rm 3}Department of Machine Learning, Mohamed bin Zayed University of Artificial Intelligence, UAE \\
	\texttt{\{huan.xiong, bin.gu\}@mbzuai.ac.ae}
}


\begin{abstract}
Recent years have emerged a surge of interest in spiking neural networks (SNNs) owing to their remarkable potential to handle time-dependent and event-driven data. The performance of SNNs hinges not only on selecting an apposite architecture and fine-tuning connection weights, similar to conventional artificial neural networks (ANNs), but also on the meticulous configuration of intrinsic structures within spiking computations. However, there has been a dearth of comprehensive studies examining the impact of intrinsic structures. Consequently, developers often find it challenging to apply a standardized configuration of SNNs across diverse datasets or tasks. This work delves deep into the intrinsic structures of SNNs. Initially, we unveil two pivotal components of intrinsic structures: the integration operation and firing-reset mechanism, by elucidating their influence on the expressivity of SNNs. Furthermore, we draw two key conclusions: (1) the membrane time hyper-parameter is intimately linked to the eigenvalues of the integration operation, dictating the functional topology of spiking dynamics; (2) various hyper-parameters of the firing-reset mechanism govern the overall firing capacity of an SNN, mitigating the injection ratio or sampling density of input data. These findings elucidate why the efficacy of SNNs hinges heavily on the configuration of intrinsic structures and lead to a recommendation that enhancing the adaptability of these structures contributes to improving the overall performance and applicability of SNNs. Inspired by this recognition, we propose two feasible approaches to enhance SNN learning. These involve leveraging self-connection architectures and employing stochastic spiking neurons to augment the adaptability of the integration operation and firing-reset mechanism, respectively. We theoretically establish that (1) both methods promote the expressive property for universal approximation, (2) the incorporation of self-connection architectures fosters ample solutions and structural stability for SNNs approximating adaptive Hamiltonian systems, and (3) the stochastic spiking neuron model aids in constraining generalization with an exponential reduction in Rademacher complexity, in comparison to both conventional ANNs and SNNs. Empirical experiments conducted on various real-world datasets affirm the effectiveness of our proposed methods.
\end{abstract}

\end{frontmatter}

\section{Introduction} \label{sec:intro}


Spiking neural network (SNN) has garnered increasing attention as a bio-inspired neural network model due to its great potential in neuromorphic computing and sparse computation~\citep{maass1997,maass2001}. The SNN building  emulates the information communication mechanism among biological neurons, wherein spiking neurons communicate with each other through sequences of spikes. Within a neuron, a spiking neuron replicates the process of converting information between membrane potentials and spikes, employing the integration-and-fire paradigm. 

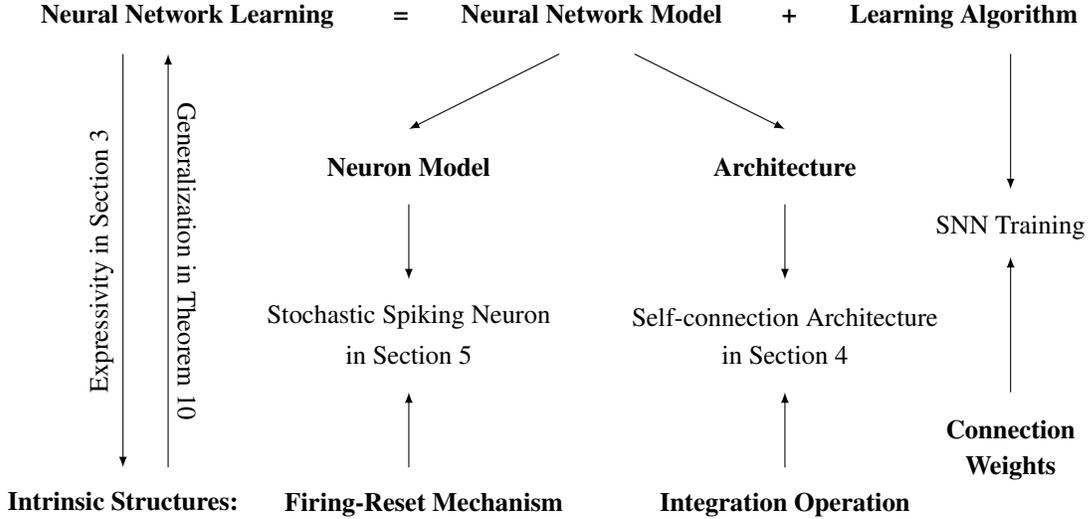
\begin{figure}[t]
	\begin{tikzpicture}
		\tikzmath{ 
			real = \a;
			\a = 4;
			real = \b;
			\b = 9;
			real = \c;
			\c = 12;
			coordinate \learning; 
			\learning = (6,\a);
			coordinate \neuron; 
			\neuron = (\a,2);
			coordinate \architecture; 
			\architecture = (\b,2);
		}
		\node at (\learning) {\bf Neural Network Learning \quad\quad=\quad\quad Neural Network Model \quad\quad+\quad\quad Learning Algorithm}; 
		\draw[-latex] (6,3.5) -- (\a,2.5);
		\draw[-latex] (7,3.5) -- (\b,2.5);
		\node at (\neuron) {\bf Neuron Model};
		\node at (\architecture) {\bf Architecture};
		\draw[-latex] (\a,1.5) -- (\a,0.5);
		\draw[-latex] (\b,1.5) -- (\b,0.5);
		\node at (\a,0) {Stochastic Spiking Neuron};
		\node at (\a,-0.5) {in Section~\ref{sec:stoc}};
		\node at (\b,0) {Self-connection Architecture};
		\node at (\b,-0.5) {in Section~\ref{sec:self-connection}};
		\draw[-latex] (\c,3.5) -- (\c,1.7);
		\node at (\c,1.2) {SNN Training};
		\draw[-latex] (\a,-2) -- (\a,-1);
		\draw[-latex] (\b,-2) -- (\b,-1);
		\node at (4.2,-2.5) {\textbf{Firing-Reset Mechanism}};
		\node at (\b,-2.5) {\textbf{Integration Operation}};
		\draw[-latex] (\c,-1) -- (\c,0.8);
		\node at (\c,-1.5) {\textbf{Connection}};
		\node at (\c,-2) {\textbf{Weights}};
		\node at (0.2,-2.45) {\bf Intrinsic Structures:};
		\draw[-latex] (0.2,3.5) -- (0.2,-2) node[midway, left] {\rotatebox{90}{Expressivity in Section~\ref{sec:inside}}};
		\draw[-latex] (0.8,-2) -- (0.8,3.5) node[midway, right] {\rotatebox{270}{Generalization in Theorem~\ref{thm:generalization_Stoc}}};
	\end{tikzpicture}
	\caption{Overview of this work.}
	\label{fig:overview}
\end{figure}

According to the framework of neural network learning, the performance of SNNs hinges not only on the determination of network architectures and the training of connection weights, akin to artificial neural networks (ANNs) with conventional McCulloch-Pitts (MP) neurons~\citep{mcculloch1943:mp}, but also on the specific configurations of intrinsic structures. These encompass the spiking computations within the integration-and-fire paradigm and their corresponding hyper-parameters, as depicted in Figure~\ref{fig:overview}. Regrettably, there has been a dearth of analyzable studies on exploring the impact of these intrinsic structures.

A consensus regarding the optimal configuration for deploying SNNs across various datasets or tasks remains elusive. Moreover, there is currently no systematic guidance available on the choice between leaky and general formations or on how to fine-tune the associated hyper-parameters. Existing SNNs often adhere to biologically plausible knowledge from neuroscience when configuring these intrinsic structures, occasionally making minor adjustments to hyper-parameters~\citep{carlson2014}. Consequently, developers often grapple with the challenge of discerning which elements of the intrinsic structures are indispensable and how best to adjust the corresponding hyper-parameters for various learning tasks.

In this paper, we embark on a theoretical exploration of the intrinsic structures inherent to SNNs. Initially, we unveil two pivotal components: the integration operation and the firing-reset mechanism, by deconstructing the expressivity of SNNs. Moreover, we draw two key conclusions:
\begin{itemize}
	\item The membrane time hyper-parameter intricately correlates with the eigenvalue of the integration operation, ultimately dictating the functional topology of spiking dynamics. An ill-suited configuration can impede proper SNN learning, elevating the risk of structural instability.
	\item Diverse hyper-parameters within the firing-reset mechanism exert influence over the firing capacity of SNNs, irrespective of the injection ratio or sampling density of input data. An improperly set firing-reset mechanism can obstruct the generation of spiking patterns. This can manifest as an excess of inhibited or even non-responsive neurons, or alternatively, an elevated excitation frequency.
\end{itemize}
In contrast to prior studies, our work challenges the conventional manners of pre-fixing intrinsic structures prior to learning. Instead, we advocate for adaptive settings that respond to the specific data or environmental context. Adopting such adaptable intrinsic structures stands to significantly enhance the performance and applicability of SNNs. Inspired by this recognition, we propose two methods for improving SNN learning, that is, adding the self-connection architecture and incorporating a stochastic spiking neuron model, which correspond to the modifications of the network architecture and spiking neuron model in Figure~\ref{fig:overview}, respectively.

The starting point of adding self-connection architectures is from advanced studies that have characterized typical SNNs as bifurcation dynamical systems~\citep{zhang2021:bsnn}. These investigations have shed light on the potential issues of structural instability arising from simple configurations of SNNs. In contrast, SNNs enhanced with self-connection architectures evolve into adaptive dynamical systems~\citep{zhang2022:SNNsTheory}, as exemplified in Figure~\ref{fig:SS_for_bifurcation}. This insight motivates our exploration of the potential benefits of self-connection architectures, specifically in enhancing the adaptivity of integration operations. Consequently, we theoretically prove that SNNs equipped with self-connection architectures (1) serve as universal approximators, (2) offer an ample array of solutions for approximating adaptive Hamiltonian systems, and (3) exhibit heightened structural stability, as specified by the lower and upper bounds of the maximum number of limit cycles.

The key idea of incorporating the stochastic spiking neuron model is to introduce a level of stochasticity to the firing-reset mechanism. This manner ensures that a spiking neuron maintains a certain excitation probability of being activated even though the integrated membrane potential falls short of the firing threshold. Besides, the inclusion of this probabilistic element engenders an unbiased and non-asymptotic estimator for gradients, enabling gradient calculations. We theoretically prove that the SNN equipped with stochastic spiking neurons possesses (1) the expressive attributes of universal approximation along with approximation complexity advantages over conventional ANNs and SNNs, and (2) the explicit generalization bounds in which the excitation probability exponentially reduces the Rademacher complexity compared to previous studies on ANNs.

\vspace{0.2 cm}
Our main contributions are summarized as follows:
\begin{itemize}
	\item \textbf{Theoretical Investigation of Intrinsic Structures.} We conduct a thorough theoretical investigation into two crucial types of intrinsic structures: the integration operation and the firing-reset mechanism. This analysis leads to two significant conclusions:
	\begin{itemize}
		\item The membrane time hyper-parameter exhibits a close relationship with the eigenvalues of the integration operation. See Subsection~\ref{subsec:integration}.
		\item The firing-reset mechanism fundamentally determines the firing capacity of SNNs. See Subsection~\ref{subsec:firing-reset}.
	\end{itemize}	
	\item \textbf{Advocacy for Adding Self-connection Architectures.} We advocate adding self-connection architecture to improve the adaptivity of the integration operation. This addition has three noteworthy outcomes:
	\begin{itemize}
		\item The SNN with self-connection architectures has the property of universal approximation, as stated in Theorem~\ref{thm:ua_ScSNN}.
		\item The self-connection architecture promotes an abundance of solutions, ranging from polynomial to exponential complexities, for SNNs approximating adaptive Hamiltonian systems, as stated in Theorem~\ref{thm:bifurcation_bound}.
		\item Adding self-connection architectures contributes to greater structural stability of SNNs, as stated in Theorem~\ref{thm:ScSNN_periodic}, Theorem~\ref{thm:ScSNN_lower}, and Proposition~\ref{prop:upper}.
	\end{itemize}
	\item \textbf{Introduction of Stochastic Spiking Neuron Model.} We propose the stochastic spiking neuron model by probabilizing the firing-reset mechanism. We have four significant findings:
	\begin{itemize}
		\item The stochastic spiking neuron model induces an unbiased and non-asymptotic estimator for gradients; thus, SNNs equipped with stochastic spiking neurons allow gradient calculations. See Subsection~\ref{subsec:EBP}.
		\item The SNN equipped with stochastic spiking neurons has the universal approximation property in Theorem~\ref{thm:ua_stoc}.
		\item The SNN equipped with stochastic spiking neurons exhibits the approximation complexity advantages, over conventional ANNs and SNNs, as stated in Theorem~\ref{thm:computational_power}. 
		\item We present the explicit generalization bounds for SNNs equipped with stochastic spiking neurons. Notably, the excitation probability exponentially reduces the Rademacher complexity, offering a notable performance compared to previous studies on ANNs. See Theorem~\ref{thm:generalization_Stoc} and Theorem~\ref{thm:estimation_for_deep_layer}.
	\end{itemize}
	\item \textbf{Experimental Validation.} The experiments conducted on static and neuromorphic datasets demonstrate the effectiveness of our proposed methods.
\end{itemize}

The rest of this paper is organized as follows. Section~\ref{sec:notations} introduces some useful notations. Section~\ref{sec:inside} investigates the expressivity and intrinsic structures of SNNs.  Section~\ref{sec:self-connection} and Section~\ref{sec:stoc} propose the improved methods by modifying the network architecture and spiking neuron model, respectively. Section~\ref{sec:experiments} conducts numerical experiments. Section~\ref{sec:conclusions} concludes this work.



\section{Notations}  \label{sec:notations}
We here introduce some useful terminologies and related notations. Let $[N] = \{1, 2, \dots, N\}$ be an integer set for $N \in \mathbb{N}^+$, and $|\cdot|_{\#}$ denotes the number of elements in a collection, e.g., $|[N]|_{\#} = N$. The symbol $\boldsymbol{x} \preccurlyeq 0$ means that every element $x_i \leq 0$ for any $i \in [|\boldsymbol{x}|_{\#}]$. Let the sphere $\mathcal{S}(r)$ and globe $\mathcal{B}(r)$ be $\mathcal{S}(r) = \{ \boldsymbol{x} \mid \| \boldsymbol{x} \|_2 = r \}$ and $\mathcal{B}(r) = \{ \boldsymbol{x} \mid \| \boldsymbol{x} \|_2 \leq r \}$ for any $r\in\mathbb{R}$, respectively. Given two functions $g,h\colon \mathbb{N}^+\rightarrow \mathbb{R}$, we denote by $h=\Theta(g)$ if there exist positive constants $c_1,c_2$, and $n_0$ such that $c_1g(n) \leq h(n) \leq c_2g(n)$ for every $n \geq n_0$; $h=\mathcal{O}(g)$ if there exist positive constants $c$ and $n_0$ such that $h(n) \leq cg(n)$ for every $n \geq n_0$; $h=\Omega(g)$ if there exist positive constants $c$ and $n_0$ such that $h(n) \geq cg(n)$ for every $n \geq n_0$. 

The general linear group over field $\mathbb{F}$, denoted by $\mathbf{GL}(n, \mathbb{F})$, is the set of $n \times n$ invertible matrices with entries in $\mathbb{F}$. Especially, we define that a special linear group $\mathbf{SL}(n, \mathbb{F})$ is the subgroup of $\mathbf{GL}(n, \mathbb{F})$ that consists of matrices with determinant 1. For any field $\mathbb{F}$, the $n \times n$ orthogonal matrices form the following subgroup
\[
\mathbf{O}(n,\mathbb{F}) = \{ \mathbf{P} \in \mathbf{GL}(n, \mathbb{F}) \mid \mathbf{P}^{\top}\mathbf{P} = \mathbf{P}\mathbf{P}^{\top} = \mathbf{E}_n \}
\]
of the general linear group $\mathbf{GL}(n, \mathbb{F})$, where $\mathbf{E}_n$ is a $n \times n$ identity matrix. Similarly, we can denote a special orthogonal group by $\mathbf{SO}(n,\mathbb{F})$, which consists of all orthogonal matrices of determinant 1 and is a normal subgroup of $\mathbf{O}(n,\mathbb{F})$. Therefore, this group is also called the rotation group. 

Let $\mathcal{C}(K, \mathbb{R})$ be the set of all scalar functions $f: K \rightarrow \mathbb{R}$ which are continuous on $K \subset \mathbb{R}^n$. Given $\boldsymbol{\alpha} = (\alpha_1, \alpha_2, \dots, \alpha_n)^{\top} \in \mathbb{N}^n$, we define 
\[ 
D^{|\boldsymbol{\alpha}|} f(\boldsymbol{x}) = \frac{\partial^{\alpha_1}}{\partial x_1^{\alpha_1}} \frac{\partial^{\alpha_2}}{\partial x_2^{\alpha_2}} \dots \frac{\partial^{\alpha_n}}{\partial x_n^{\alpha_n}} f(\boldsymbol{x}) \ ,
\]
where $\boldsymbol{x} = (x_1, x_2, \dots, x_n) \in K$ and $|\boldsymbol{\alpha}| = \sum_{i \in [n]} \alpha_i$. Further, we define
\[
\mathcal{C}^l(K, \mathbb{R}) = \left\{ f \in \mathcal{C}(K, \mathbb{R}) \mid D^{|\boldsymbol{\alpha}|} f \in \mathcal{C}^0(K, \mathbb{R}) \right\} 
\]
for $\boldsymbol{\alpha} \in \mathbb{N}^n$ with $|\boldsymbol{\alpha}| \leq l$. For $p \geq 1$, we define
\[
\mathcal{L}^p(K,\mathbb{R}) = \left\{  f \in \mathcal{C}(K,\mathbb{R}) ~\left|~ \left\| f \right\|_{p,K} < \infty \right. \right\} \ ,
\]
where
\[
\left\| f \right\|_{p,K} \overset{\mathrm{def}}{=} \left(\int_{K} |f(\boldsymbol{x})|^p \dif\boldsymbol{x}\right)^{1/p} \ .
\]
This work considers the Sobolev space $\mathcal{W}^{l,p}(K,\mathbb{R})$, defined as the collection of all functions $f \in \mathcal{C}^l(K, \mathbb{R})$ and $D^{\boldsymbol{\alpha}} f \in \mathcal{L}^p(K,\mathbb{R})$ for all $|{\boldsymbol{\alpha}}| \in [l]$, that is, 
\[
\| D^{\boldsymbol{\alpha}} f \|_{p,K} = \left( \int_{K} \left| D^{\boldsymbol{\alpha}} f (\boldsymbol{x}) \right|^p \dif\boldsymbol{x} \right)^{1/p} < \infty \ ,
\]
in which we employ $p=2$ as the default throughout this paper.

\section{Spiking Dynamics and Intrinsic Structures of Spiking Neural Networks}   \label{sec:inside}
The computational process of SNNs complies with an integration-and-firing paradigm, which comprises an \emph{integration operation} and a \emph{firing-reset mechanism} as follows.

\vspace{0.2cm}
\noindent\textbf{Integration Operation.}  The integration operation of SNNs is usually formulated as some first-order differential equations. The leaky integration-and-firing (LIF) model is common type of spiking equations, of which the general form is as follows
\begin{equation}  \label{eq:SNN_LIF}
	\quad\text{LIF}:\quad	\tau_m \frac{\dif \boldsymbol{u} (t)}{\dif t} = - \boldsymbol{u}(t) + \tau_r f_{\textrm{agg}} (\mathbf{I}(t)) \ ,
\end{equation}
where $\boldsymbol{u}(t) = (u_1(t), \dots, u_N(t))^{\top}$ indicates the membrane potential vector of $N$ spiking neurons at timestamp $t$, $\mathbf{I}(t) = (\mathbf{I}_1(t), \dots, \mathbf{I}_M(t))^{\top}$ denotes the $M$-dimensional input signals, $\tau_m$ and $\tau_r$ are positive-valued hyper-parameters with respect to membrane time and membrane resistance, respectively. Here, $f_{\textrm{agg}}$ is an aggregation function, usually with the following form
\[
f_{\textrm{agg}} (\mathbf{I}(t)) = \mathbf{W}~ \mathbf{I}(t) \ ,
\]
where $\mathbf{W}$ is the learnable connection matrix. Sometimes, we should normalize the aggregated distributions using some techniques such as batch normalization, which not only modifies the feature space by re-centering but also avoids larger information flows and gradient explosion by re-scaling. For the simplicity of mathematical formation, we omit the normalization techniques throughout this paper except for experiments.

\vspace{0.2cm}
\noindent\textbf{Firing-Reset Mechanism.} The spiking neuron model employs the typical threshold rule, that is, neuron $k$ $(k\in[N])$ fires spikes $s_k(t)$ at time $t$ if and only if $u_k(t) \geq u_{\text{firing}}$ where $u_{\text{firing}}$ indicates the firing threshold. We formulate this procedure using a spike excitation function
\begin{equation} \label{eq:excitation}
	f_e: \mathbb{R} \to \mathbb{R} \ ,
	\quad\text{where}\quad
	s_k(t) = f_e(u_k(t)) \overset{\mathrm{def}{}}{=} \left\lfloor \frac{u_k(t)}{ u_{\text{firing}} }  \right\rfloor \ .
\end{equation}
After firing, the membrane potential is instantaneousaneously reset to a lower value $u_{\text{reset}}$, that is, reset voltage; formally, one has
\[
u(t) = (1 - s(t)) \cdot u(t) + s(t) \cdot u_{\text{reset}} \ .
\]
Note that this work does not consider using absolute refractory periods~\citep{hunsberger2015:refractory} or refractory kernels~\citep{dumont2017:refractory}.
\begin{figure}[t]
	\centering
	\includegraphics[width=1\textwidth]{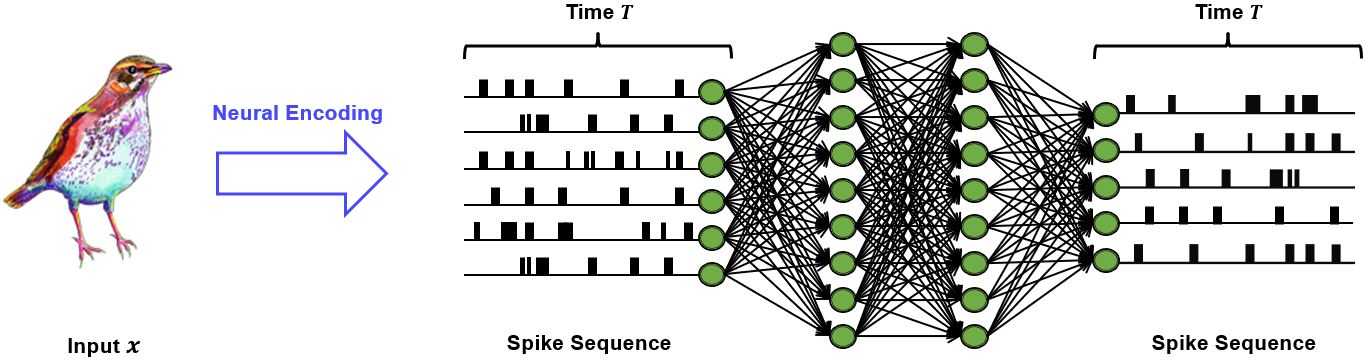}
	\caption{Illustrations for the working flow of SNNs.}
	\label{fig:strings}
\end{figure}

\vspace{0.2cm}
\noindent\textbf{Neural Encoding.} Input signals of SNNs are formal of binary strings or equally spike sequences, i.e., $\mathbf{I}_j(t) \in \{0,1\}$ for $j\in[M]$.  In cases where non-spiking data is provided, it needs to be converted into spike format during pre-processing. This conversion technique is commonly referred to as \emph{neural encoding}. Neural encoding methods can be broadly categorized into two main groups: \emph{timing-based encoding} and \emph{rate-based encoding}. All specialized encoding schemes can be separated into one of these two by answering whether the exact timing and order of spikes are crucial for the information to be submitted~\citep{auge2021:encoding}. 

In timing-based encoding, spike sequence relies on the precise timing of every spike, e.g., encoding with the distance between time instances that fire spikes~\citep{mostafa2017:encoding}. The following displays several key computations of timing-based encoding, in which
\begin{equation} \label{eq:timing_encoding}
	\left\{~\begin{aligned}
		\text{TTFS Timing}:&~ I(t) = t_{\text{spike}} - t^{(0)}_{\text{spike}} \ , \\
		\text{ISI Timing} ~:&~ I(t) = t_{\text{spike}} - t'_{\text{spike}} \ , \\
	\end{aligned}\right.
\end{equation}
where $t_{\text{spike}}$, $t'_{\text{spike}}$, and $t^{(0)}_{\text{spike}}$ denote the timing of current, last, and initial spikes, respectively. In addition, binary encoding is noteworthy, where each spike is associated with a ``1" (or ``0") in a bit stream, indicating the occurrence (or non-occurrence) of a spike within a specified interval or the timing of the spike within that interval. This manner guarantees a consistent presence of spikes, regardless of the specific bit pattern being encoded. Consequently, timing-based encoding can achieve higher information densities and efficiencies. Unfortunately, temporal encoding typically entails more intricate architectures and may lack well-established training methods. 

In rate-based encoding, spike sequence relates to the spike activity over time, e.g., encoding with the count, density, and population of fired spikes within temporal windows~\citep{subbulakshmi2021:encoding}. The following displays the illustration of several key computations of rate-based encoding
\begin{equation} \label{eq:rate_encoding}
	\left\{~\begin{aligned}
		\text{Count Rate}\quad:&~ I(t) = \frac{N_{\text{spike}}(t:t + \Delta t)}{\Delta t} 
		\quad\text{(average over time)} \ , \\
		\text{Density Rate}~:&~ I(t) = \frac{N_{\text{spike}}(t:t + \Delta t)}{N_{\text{runs}} \Delta t}
		\quad\text{(average over several runs)} \ , \\
		\text{Population Rate}:&~ I(t) = \frac{N_{\text{spike}}(t:t + \Delta t)}{N_{\text{neurons}} \Delta t}
		\quad\text{(average over several neurons)} \ , \\
	\end{aligned}\right.
\end{equation}
where $N_{\text{spike}}(t:t + \Delta t)$ denotes the spike count over interval $[t,t + \Delta t]$, $\Delta t$ is the time window,  $N_{\text{runs}}$ and $N_{\text{neurons}}$ indicates the neural activities measured over different simulations. Thus, the rate-based encoding can be convincible through its robustness against fluctuations and noise due to its simplicity. 

It is observed that there is an invertible transformation between the rate-based and timing-based encoding techniques, which is theoretically investigated in Subsection~\ref{subsec:approximation_stoc}. In practice, the rate-based encoding has become the simplest and most popular encoding scheme in SNNs, and researchers usually employ rate-based data encoded by a Poisson distribution~\citep{susemihl2013:encoding} or recorded by a Dynamic Vision Sensor~\citep{quiroga2005:encoding}.

\vspace{0.2cm}
\noindent\textbf{Training Approaches.} The last two decades have witnessed the increasing prominence of SNNs in machine learning and artificial intelligence research, leading to a proliferation of efficient software packages for their training and deployment. The most popular approaches for SNNs training originated from the spike response model scheme~\citep{gerstner1995:srm}, based on which Eq.~\eqref{eq:SNN_LIF} has the following solution with the boundary condition $u_{\text{reset}}=0$
\begin{equation} \label{eq:solution}
	u_k(t) = \int_{t'}^{t} \exp\left( - \frac{t''-t'}{\tau_m} \right) \Delta(t'') \dif t'' 
	\quad\text{with}\quad
	\Delta(t'') = \frac{\tau_r}{\tau_m} \sum_{j\in[M]} \mathbf{W}_{kj} \mathbf{I}_j(t'') \ ,
\end{equation}
where $t'$ denotes the last firing timestamp $t' = \max \{ t'' \mid u_k(t'') = u_{\text{firing}}, ~t'' < t \}$. 

Existing approaches for training SNNs can be roughly divided into two categories. The first category, referred to as conversion approaches between ANNs and SNNs, entails employing a straightforward continuous-valued ANN during the training process and subsequently converting it into an accurate spiking equivalent~\citep{diehl2016:conversion}. Notably, \citet{rueckauer2017:conversion} introduced conversions between SNNs and CNNs, encompassing architectures like VGG-16 and Inception-v3. The second category, known as direct training approaches, involves configuring an SNN to accommodate discontinuous spike activities and then training it using back-propagation through time~\citep{huh2018:BPTT}. Famously, SpikeProp and its variants transfer the information in the timing of a single spike~\citep{bohte2002:SpikeProp,mckennoch2006:SpikeProp}. However, SpikeProps are constrained to single-spike learning, which usually causes the deactivation of numerous neurons, that is, a phenomenon known as ``dead neurons"~\citep{jin2018}. Some researchers attempted to approximate the back-propagation dynamics by some surrogate gradients~\citep{neftci2019:surrogate,li2021:surrogate}. \citet{shrestha2018:slayer} incorporated firing derivatives by incorporating the temporal dependency between spikes; thus, the back-propagated error at a given time step becomes an integration of earlier spike inputs.

\vspace{0.2cm}
\noindent\textbf{Expressivity.} From Eq.~\eqref{eq:solution}, we can formulate the function expressed by a two-layer SNN as follows
\begin{equation}  \label{eq:express_classical}
	\left\{~\begin{aligned}
		f(\cdot,t)  &= \frac{1}{t-t'}~ \sum_{k=1}^{N} w_k 	f^{\textrm{hidden}}_k(\cdot,t) \ , \\
		f^{\textrm{hidden}}_k(\cdot,t) &= f_e \left(  \frac{\tau_r}{\tau_m} \int_{t'}^t \exp\left(  - \frac{t'' - t'}{\tau_m} \right) \mathbf{W}_{k,[M]} \mathbf{I}(t'') \dif t''  \right) \ ,
	\end{aligned}\right.
\end{equation}
in which $f^{\textrm{hidden}}_k(\cdot, t)$ indicates the expressive sub-function related to the $k^{\textrm{th}}$ hidden neuron for $k\in [N]$, $w_k$ and $\mathbf{W}_{k,[M]}$ denote the first-layer and second-layer weights that connect to the $k^{\textrm{th}}$ hidden neuron, respectively, where $\mathbf{W}_{k,[M]}$ is a row vector, i.e., the $i^{\textrm{th}}$ row of matrix $\mathbf{W}$. Here, we add a denominator $t-t'$ so that the expressive function $f(\cdot,t)$ indicates the firing rate of the concerned SNN after the last firing timestamp $t'$. According to the recognition of~\citep[Theorem 1]{zhang2022:SNNsTheory}, the expressive hypotheses with the form of Eq.~\eqref{eq:express_classical} are not dense in continuous function space $\mathcal{C}^l(K,\mathbb{R})$ for $l \in \mathbb{N}^+$.

Notice that the expressive function of the LIF-SNN consists of several key components: the membrane time hyper-parameter $\tau_m$ in the integration operation, the spike excitation function $f_e$ that corresponds to the membrane resistance hyper-parameter $\tau_r$ and firing threshold hyper-parameter $u_{\textrm{firing}}$ in the firing-reset mechanism, and the learnable connection weights $w_k$ and $\mathbf{W}_{k,[M]}$. According to the framework of neural network learning in Figure~\ref{fig:overview}, the former two closely relate to the intrinsic structures within spiking computations. 

Over the past decades, there have been theoretical studies examining the expressivity of SNNs. Maass et al.~\citep{maass1997,maass2001} showed that the designed SNNs can simulate some typical computational models such as Turing machines, random access machines, threshold circuits, etc. \citet{she2021:sequence} showed the universal approximation property of SNNs by leveraging spike propagation paths. \citet{zhang2022:SNNsTheory} introduced an instantaneous firing rate to enable SNNs to approximate dynamical systems. However, systematic analyses of the intrinsic structures of SNNs have been notably absent in prior studies. In the forthcoming subsections, we will take an in-depth analysis of the roles played by these two key components.

\subsection{Eigenvalues of Integration Operations}  \label{subsec:integration}
We first investigate the Hamiltonian system for spiking dynamics led by the integration operation.  Before that, we define the accumulative pulse voltage (i.e., the variable of the pre-synaptic state) $\boldsymbol{v}(t) \in \mathbb{R}^N$ as follows
\begin{equation} \label{eq:SNN_v}
	\frac{\dif \boldsymbol{v}(t)}{\dif t} = f_{\textrm{agg}} (\mathbf{I}(t))  \ .
\end{equation}
For convenience, we specify the mapping $h$ between $v$ and $f_{\textrm{agg}} (\mathbf{I}(t))$ as 
\[
h: v(t) \mapsto f_{\textrm{agg}} (\mathbf{I}(t)) \ .
\]
Thus, Eq.~\eqref{eq:SNN_LIF} becomes
\begin{equation}  \label{eq:SNN_converted}
	\left\{~\begin{aligned}
		\tau_m \frac{\dif \boldsymbol{u} (t)}{\dif t} &= - \boldsymbol{u}(t) + \tau_r \frac{\dif \boldsymbol{v}(t)}{\dif t} \\
		\frac{\dif \boldsymbol{v}(t)}{\dif t} &= h(\boldsymbol{v})  \ .
	\end{aligned}\right.
\end{equation}
According to Pontryagin's Minimum principle~\citep{pontryagin1987mathematical}, the Hamiltonian system for spiking dynamics in Eq.~\eqref{eq:SNN_LIF} is led by
\begin{equation} \label{eq:H_Hamilton}
	\mathcal{H}(\boldsymbol{u}, \boldsymbol{v}, t) = \left\langle \boldsymbol{p}, \frac{\dif \boldsymbol{u}}{\dif t} \right\rangle + \left\langle \boldsymbol{q}, \frac{\dif \boldsymbol{v}}{\dif t} \right\rangle + l(\boldsymbol{u}) \ ,
\end{equation}
where $l(\cdot)$ is the cost function, and $\boldsymbol{p} = (p_1, p_2, \dots, p_N)^{\top} \in \mathbb{R}^N$ and $\boldsymbol{q} = (q_1, q_2, \dots, q_N)^{\top} \in \mathbb{R}^N$ indicate the adjoint state variables that correspond to the membrane voltage $\boldsymbol{u}$ and synaptic current $\boldsymbol{v}$, respectively, in which the dynamics of these adjoint state variables are
\[
\frac{\dif \boldsymbol{p}}{\dif t} = - \frac{\partial \mathcal{H}}{\partial \boldsymbol{u}} 
\quad\text{and}\quad
\frac{\dif \boldsymbol{q}}{\dif t} = - \frac{\partial \mathcal{H}}{\partial \boldsymbol{v}} \ .
\]
Combined with Eq.~\eqref{eq:SNN_converted}, Eq.~\eqref{eq:H_Hamilton} becomes
\[
\begin{aligned}
	\mathcal{H}(\boldsymbol{u}, \boldsymbol{v}, t) &= \left\langle \boldsymbol{p}, -\frac{\boldsymbol{u}}{\tau_m} + \frac{\tau_r}{\tau_m} \frac{\dif \boldsymbol{v}}{\dif t} \right\rangle + \bigg\langle \boldsymbol{q}, h(\boldsymbol{v}) \bigg\rangle + l(\boldsymbol{u})  \\
	&= - \left\langle \boldsymbol{p}, \frac{\boldsymbol{u}}{\tau_m} \right\rangle + \left\langle \boldsymbol{q} + \frac{\tau_r}{\tau_m} \boldsymbol{p}, h(\boldsymbol{v}) \right\rangle + l(\boldsymbol{u})
\end{aligned}
\]
with for $i \in [N]$
\[
\left\{~\begin{aligned}
	\frac{\dif p_i}{\dif t} &= \frac{p_i}{\tau_m} - \left( q_i + \frac{\tau_r}{\tau_m} p_i \right) \left( \sum_{j\in[M]} \mathbf{W}_{ij} \frac{\partial h (\boldsymbol{v}) }{\partial \mathbf{I}_j} \right) - \frac{\dif l(\boldsymbol{u})}{\dif u_i} \\
	\frac{\dif q_i}{\dif t} &= - \left( q_i + \frac{\tau_r}{\tau_m} p_i \right) \frac{\dif h (\boldsymbol{v}) }{\dif v_i} \ .
\end{aligned}\right.
\]
The above formula can be simplified by the following variable conversions
\[
\tilde{\boldsymbol{p}} = \frac{1}{\tau_m} \boldsymbol{p}
\quad\text{and}\quad
\tilde{\boldsymbol{q}} = \boldsymbol{q} + \frac{\tau_r}{\tau_m} \boldsymbol{p} \ ,
\]
which yield
\begin{equation}  \label{eq:H_Hamilton_tilde}
	\left\{~\begin{aligned}
		\mathcal{H}(\boldsymbol{u}, \boldsymbol{v}) &= - \left\langle \tilde{\boldsymbol{p}}, \boldsymbol{u} \right\rangle + \left\langle \tilde{\boldsymbol{q}}, h(\cdot) \right\rangle + l(\boldsymbol{u}) \\
		\tau_m \frac{\dif \tilde{\boldsymbol{p}} }{\dif t} ~ &= \tilde{\boldsymbol{p}} - \mathbf{W} \mathbf{M}  \tilde{\boldsymbol{q}} - \frac{\dif l(\boldsymbol{u})}{\dif \boldsymbol{u}} 
		\quad\text{where}\quad
		\mathbf{M} _{ji} = \frac{\partial h (\boldsymbol{v}) }{\partial \mathbf{I}_j} \\
		\frac{\dif \tilde{\boldsymbol{q}} }{\dif t} \quad &= - \mathbf{M}^{(v)} \tilde{\boldsymbol{q}} + \tau_r \frac{\dif \tilde{\boldsymbol{p}} }{\dif t}  
		\quad\text{where}\quad
		\mathbf{M}^{v}_{ki} = \frac{\dif h (\boldsymbol{v}) }{\dif v_k} \ .
	\end{aligned}\right.
\end{equation}

Intuitively, we have a total ``energy'' defined for the overall network as follows
\begin{equation}  \label{eq:energy_network}
	\mathcal{H}(t) = |\boldsymbol{u}|^2 + \frac{2\tau_r}{\tau_m} \int \left\langle \frac{\dif \boldsymbol{v}}{\dif t},  \boldsymbol{u}(t) \right\rangle  \dif t - \theta \ ,
\end{equation}
where $\theta$ is a universal constant. Correspondingly, the unit energy defined on the $k^{\textrm{th}}$ spiking neuron ($k\in[N]$) becomes
\[
\mathcal{H}_k(t) = u_k^2(t) + \frac{2\tau_r}{\tau_m} \int \sum_{j\in[M]} \mathbf{W}_{kj} \mathbf{I}_j(t) u_k(t) \dif t - \theta_k \ ,
\]
where $\theta_1 + \theta_2 + \dots + \theta_N = \theta$. It is evident that the defined functions above satisfy the Hamiltonian system derived from Eq.~\eqref{eq:H_Hamilton} and Eq.~\eqref{eq:H_Hamilton_tilde}. Alternatively, it is observed that Eq.~\eqref{eq:energy_network} is a Lyapunov-like function that converts SNNs into a Hamiltonian dynamical system. With direct calculations, we can obtain the following derivative 
\[
\frac{\dif \mathcal{H}}{\dif t} = \frac{1}{2} \boldsymbol{u}^{\top} ~\mathbf{M}(\tau_m)~ \boldsymbol{u} \ ,
\]
where $\mathbf{M}(\tau_m)$ is of the quadratic form
\begin{equation}  \label{eq:H_der}
	\mathbf{M}(\tau_m) = \begin{pmatrix}
		-1/\tau_m      & 0    & \dots   & 0 \\
		0 &    -1/\tau_m      & \dots   & 0 \\
		\vdots    & \vdots       & \ddots   & \vdots       \\
		0 & 0   & \dots    &  -1/\tau_m 
	\end{pmatrix}_{N \times N} \ .   
\end{equation}
This derivative ${\dif \mathcal{H}} / {\dif t}$ represents the rate at which the energy function changes, which is determined by the hyper-parameter $\tau_m$.

Thus, we have the following conclusion.
\begin{theorem}  \label{thm:bifurcation_dynamics}
	Provided the initial condition $u(0) = u_{\text{reset}}$ or $u(t') = u_{\text{reset}}$, SNN with LIF neurons in Eq.~\eqref{eq:SNN_LIF} leads to a Hamiltonian system, and $-1/\tau_m$ indicates the \textbf{eigenvalue} of the integration operation, where $-1/\tau_m<0$, $-1/\tau_m=0$, and $-1/\tau_m>0$ correspond to the dissipative, conservative, and energy-diffuse dynamical systems, respectively.
\end{theorem}
Theorem~\ref{thm:bifurcation_dynamics} reveals that the membrane time hyper-parameter $\tau_m$ is relative to the eigenvalue of the integration operation and determines the functional topology of spiking dynamics. This conclusion coincides with the insights of~\citep[Theorem 2]{zhang2021:bsnn}. The proof of Theorem~\ref{thm:bifurcation_dynamics} can be accessed in Appendix~\ref{app:bifurcation_dynamics}.

\vspace{0.2cm}
\noindent\textbf{Drawbacks of pre-fixing eigenvalues of integration operations.} Notice that in conventional SNN learning, the value of membrane time hyper-parameter $\tau_m$ is often pre-determined; thus, the functional topology, i.e., whether the system is dissipative, conservative, or energy-diffuse dynamical systems, of the SNN is determined no matter what system the actual data is drawn from. For example, one undertakes the task of predicting the efficiency of thermal power generation, in which the model takes in the fuel data and outputs the generated electric energy. Setting a positive value for $\tau_m$ to transform the SNN into an energy-diffuse system would be inappropriate, as prior knowledge dictates that there will inevitably be energy loss in the thermal power generation process. Thus, an improper setting of $\tau_m$ can hinder the possibility of proper SNN learning. Furthermore, for most tasks, it is usually challenging to identify in advance which system the task conforms to or the data comes from. In other words, one can hardly know how to set a proper $\tau_m$ before learning. Therefore, the eigenvalues (relative to $\tau_m$) of the integration operation of SNNs must be adaptive to the data.

Besides, the switching of various systems is highly sensitive to the relation between $1/\tau_m$ (relative to eigenvalues) and the critical point at 0; the functional topology of SNNs makes a sudden change when $1/\tau_m$ crosses the critical point. This phenomenon is commonly referred to as a ``bifurcation", as discussed by~\citet{zhang2021:bsnn}. Obviously, the bifurcation may cause an unstable structure within the hypothesis space expressed by SNNs, particularly when SNNs suffer from some functional perturbations. Therefore, the risk of unstable structure is another drawback of pre-fixing eigenvalues of integration operations prior to learning.

\subsection{Depolarization of Firing-Reset Mechanisms}  \label{subsec:firing-reset}
In this subsection, we leverage the effects of the firing-reset mechanism. For convenience, we here consider the univariate system, i.e., $u$ is a scalar variable, and replace the membrane resistance hyper-parameter $\tau_r$ with a gate function $g: \mathbb{R} \to \mathbb{R}$, satisfying that
\begin{itemize}
	\item $g$ is relative to the pulse voltage $v$, where ${\dif v(t)}/{\dif t} = f_{\textrm{agg}} (\mathbf{I}(t)) $,
	\item $g(v)$ is non-negative, i.e., $g(v) \geq 0$ for $v \in \mathbb{R}$,
	\item constant integral on time interval $[t'',t']$, i.e., $\int_{v(t'')}^{v(t')} g(v) \dif v = C_1$, where $t',t''$ are two adjacent firing timestamps and $C_1 > 0$.
\end{itemize}
Hence, Eq.~\eqref{eq:SNN_LIF} becomes
\begin{equation}   \label{eq:gate_LIF}
	\tau_m \frac{\dif u (t)}{\dif t} = - u(t) + g(v) \frac{\dif v(t)}{\dif t}  \ ,
\end{equation}
where $v(t)$ denotes the accumulative pulse capacity from timestamps $t_1$ to $t$. To ensure the neuron excitation, the accumulative pulse capacity should be larger than the firing flux, i.e., $v(t) \geq \left( u_{\textrm{firing}} - u_{\textrm{reset}} \right)$ for $t \in [t_1,t_2] \subseteq [t'', t']$, since the LIF-SNN with $\tau_m>0$ is a dissipative system. We further suppose that $v(t_1) = u_{\textrm{firing}} - u_{\textrm{reset}}$ and $v(t_2) = c \cdot ( u_{\textrm{firing}} - u_{\textrm{reset}} )$ where $ c \geq 1$. If one fixed the values of $t_1$ and $c$, we can quantify the effects of the firing-reset mechanism by investigating the information capacities at pre-synapse and post-synapse, compared to the firing flux between the firing threshold and reset voltage.

One has the time integral of membrane potential $u(t)$
\[
\begin{aligned}
	\int_{t_1}^{t_2} u(t) \dif t &=  \int_{t_1}^{t_2}  \left[ -\tau_m \frac{\dif u (t)}{\dif t} +  g(v) \frac{\dif v(t)}{\dif t} \right] \dif t   \\
	& = - \tau_m \int_{t_1}^{t_2} \frac{\dif u (t)}{\dif t} \dif t  + \int_{t_1}^{t_2}  g(v) \frac{\dif v(t)}{\dif t} \dif t  \\
	&= - \tau_m \int_{u(t_1)}^{u(t_2)}  \dif u  + \int_{v(t_1)}^{v(t_2)}  g(v) \dif v  \\
	& = \tau_m \left[ u(t_1) - u(t_2) \right] + \int_{v(t_1)}^{v(t_2)}  g(v) \dif v  \ ,
\end{aligned}
\]
where the computed integral indicates the membrane capacity (i.e., post-synaptic state variable) of spiking neurons during time interval $[t_1, t_2]$. It is observed that the membrane capacity is dominated by the membrane potential values at endpoints (i.e., timestamps $t_1$ and $t_2$) and the constant that corresponds to $g$. For convenience, we set $u(t_1) = u_{\textrm{reset}}$.

Intuitively, we define
\begin{equation}  \label{eq:gate}
	g(v) \overset{\underset{\mathrm{def}}{}}{=} \frac{C_2}{ t'-t'' } \quad (C_2>0)\ ,
\end{equation}
on interval $v(t) \in [t_1, t_2]$ for $t\in[t'',t']$. It is evident that $g(v)$ is apposite since
\[
\int_{v(t_1)}^{v(t_2)} g(v) \dif v  = \frac{(c-1)C_2}{ (t'-t'')} \left( u_{\textrm{firing}} - u_{\textrm{reset}} \right) \ .
\]
The gate function $g$ limits an excitation area, whose width is the average pulse capacity so that the neuron is activated in a gradual manner when the membrane potential crosses the excitation area. Further, the last term $g(v) \dif v(t)/\dif t$ in Eq.~\eqref{eq:gate_LIF} denotes the voltage that is instantaneously injected into the excitation area, which reduces to the conventional threshold-triggered model in Eq.~\eqref{eq:SNN_LIF} once $g(v)$ defaults as a constant like $\tau_r$.
\begin{figure}[t]
	\centering
	\includegraphics[width=1\textwidth]{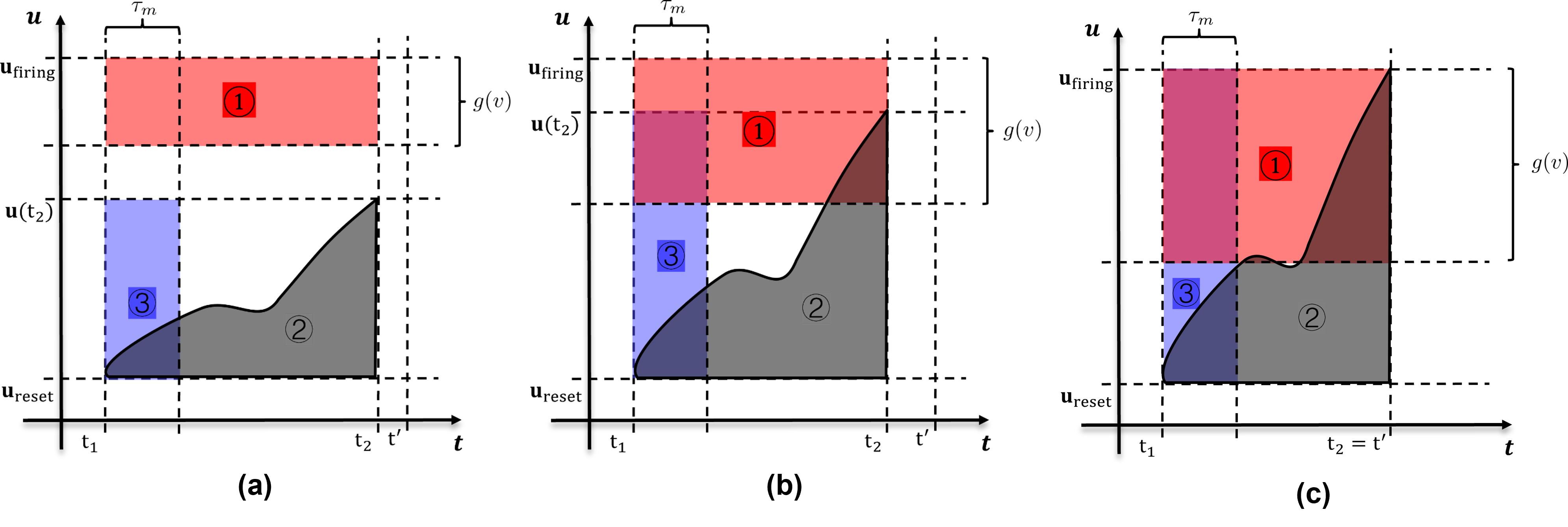}
	\caption{Illustrations for three typical cases of Eq.~\eqref{eq:voltage}.}
	\label{fig:gate}
\end{figure}

To sum up the above, we have
\begin{equation} \label{eq:voltage}
	\underbrace{ \int_{v(t_1)}^{v(t_2)}  g(v) \dif v }_{ \text{\ding{172}} } 
	- 	\underbrace{ \int_{t_1}^{t_2} u(t) \dif t }_{ \text{\ding{173}} } 
	= \underbrace{ \tau_m \left[ u(t_2) - u_{\textrm{reset}}  \right] }_{ \text{\ding{174}} }   \  ,
\end{equation}
where the last term collects the ``leaky" information during time interval $[t_1,t_2]$. Figure~\ref{fig:gate} illustrates three typical cases of Eq.~\eqref{eq:voltage}. For case (a), spike neuron receives less pulses instantaneously as the injection period becomes longer. Thus, the membrane potential cannot arrive at the excitation area. For case (b), spike neuron receives more pulses instantaneously as the injection period becomes shorter. Thus, the membrane potential has entered in the excitation area but does not exceeds the firing threshold due to inadequate pulse capacity. For case (c), the instantaneous injection pulses are enough to activate the spiking neuron, so that the membrane potential crosses the excitation area and exceeds the firing threshold, i.e., $t' = t_2$. In this case, Eq.~\eqref{eq:voltage} has a significant point
\[
(c-1)C \left( u_{\textrm{firing}} - u_{\textrm{reset}} \right) = \tau_m~  u_{\textrm{firing}}~  (t'-t_1) \ .
\]

Notice that when we mentioned ``less" or ``more" pulses, it actually corresponds to a formal description of the injection ratio between ${\dif v(t)}/{\dif t}$ and $u_{\textrm{firing}} - u_{\textrm{reset}} $, that is,
\[
\begin{aligned}
	\text{instantaneous:} &\quad \frac{\dif v(t)}{\dif t}  \cdot \frac{1}{u_{\textrm{firing}} - u_{\textrm{reset}} }  \\
	\text{average \quad:} &\quad \frac{1}{t_2 - t_1} \int_{t_1}^{t_2} \left[ \frac{\dif v(t)}{\dif t}  \cdot \frac{1}{u_{\textrm{firing}} - u_{\textrm{reset}} }  \right] \dif t  = \frac{ g(v)}{u_{\textrm{firing}} - u_{\textrm{reset}} } \ .
\end{aligned}
\]
It is observed that the injection ratio is dominated by the width of the excitation area and the capacity; a larger width as well as a smaller firing flux, results in a larger injection ratio.

The derivation above underscores an intuitive observation that increasing the injection ratio facilitates the excitation of spiking neurons. However, it is crucial to note that the injection ratio is intimately linked with the data, particularly the sampling density of neural encoding, as mentioned earlier. Once the firing-reset mechanism is pre-determined --- meaning, the values of the firing threshold and reset voltage are fixed before receiving any data --- the firing capacity of a spiking neuron is determined, regardless of the injection ratio or sampling density. Hence, even though the entire network adopts a common initialization method, it still remains possible to impede the excitation of spiking neurons or even lead to the occurrence of ``dead neurons". Developers are thus compelled to meticulously fine-tune hyper-parameters when they run SNNs on different datasets or employ diverse neural encoding techniques. For instance, when the injection rate or sampling density is low, one may need to augment the firing flux to circumvent the issue of ``dead neurons". Conversely, when the injection rate or sampling density is high, reducing the firing flux becomes imperative to regulate the excitation frequency of neurons. Therefore, the configuration of the firing-reset mechanism in SNNs must adapt to the specific data at hand.


\vspace{0.2cm}
\noindent\textbf{Summary of Section~\ref{sec:inside}.} In this section, we conducted a systematic investigation into the impact of various components on the expressivity of SNNs, specifically focusing on the intrinsic structures and hyper-parameters of the SNN model itself. Our findings led to two primary conclusions: (1) The membrane time hyper-parameter $\tau_m$ is closely tied to the eigenvalue of the integration operation, dictating the functional topology of spiking dynamics. An improper setting of $\tau_m$ can disable the possibility of proper SNN learning and elevate the risk of structural instability. Therefore, it is imperative that the eigenvalues (in relation to $\tau_m$) of the integration operation of SNNs must be adaptive to the data. (2) The firing-reset mechanism, encompassing hyper-parameters $u_{\textrm{firing}}$, $u_{\textrm{reset}}$, and $\tau_r$, fundamentally governs the firing capacity of SNNs, mitigating the influence of the injection ratio or sampling density of input data. An improper setting of the firing-reset mechanism can hamper the generation of spiking flows, potentially resulting in an abundance of inhibited neuron or even the occurrence of ``dead neurons", or alternatively, leading to a higher excitation frequency. Therefore, the setting of the firing-reset mechanism of SNNs must adapt to the data. 

Based on the above conclusions, we present two feasible ways for improving SNN learning in Section~\ref{sec:self-connection} and Section~\ref{sec:stoc}, respectively.

\section{Self-connection Architecture and Adaptive Eigenvalues}  \label{sec:self-connection}
As discussed in Subsection~\ref{subsec:integration}, it is advantageous for SNN learning to make the eigenvalues of integration operations adaptive to the data or environment. An intuitive approach is to render the membrane time hyper-parameter $\tau_m$ relative to eigenvalues as learnable, replacing the pre-fixed values. Given the loss function $E$, we list the corresponding gradients as follows
\begin{equation} \label{eq:gradients_directly}
	\left\{~\begin{aligned}
		&\nabla_{\tau_m} E \propto \tau_r \sum_{j\in[M]} \mathbf{W}_{kj} \left[ \int_{t'}^{t} \exp\left( \frac{t'-s}{\tau_m} \right) (s-t') \mathbf{I}_j(s) \dif s \right] \ , \\
		&\nabla_{\mathbf{W}_{kj}} E \propto \tau_r \int_{t'}^{t} \exp\left( \frac{t'-s}{\tau_m} \right) \mathbf{I}_j(s) \dif s \ , \\
	\end{aligned}\right.
\end{equation}
where the subscript $k$ denotes the $k^{\textrm{th}}$ spiking neuron. However, this approach presents significant challenges, as elaborated in~\citep{zhang2021:bsnn}. The main impediments are twofold. (1) The existing SNN training methodologies predominantly rely on the spike response model scheme, as depicted in Eq.~\eqref{eq:solution}, where the membrane potential $u_k(t)$ is predominantly influenced by an indirect product interaction of connection weights $\mathbf{W}_{kj}$ and the eigenvalue of integration operations $-1/\tau_m$. Consequently, concurrently optimizing both $\mathbf{W}_{kj}$ and $\tau_m$ using gradient descent is arduous. Moreover, there is a lack of guaranteed convergence for alternating gradient optimization. (2) Errors stemming from $\tau_m$ accumulate over time; thus, the gradients appear to vanish, i.e., $\partial u_k / \partial \tau_m \to 0$ as $t \to t'$, and conversely, the gradients explode over time when $t \gg t'$. Thus, directly training $\tau_m$ can lead to challenges associated with gradient explosion and vanishing. 

Moreover, a feasible approach involves searching for $\tau_m$ using zero-order optimization like Bayesian optimization. The key idea is to regard $\tau_m$ as a group of hyper-parameters drawn from a prior distribution so that the optimization challenges associated with solving for $\mathbf{W}_{kj}$ and $\tau_m$ can be reduced to mature methods. However, this approach succeeds in an apposite initialization, placing greater demands on computation and storage.

In this section, we introduce an alternative approach, i.e., adding the self-connection architecture, to attain adaptive eigenvalues of the integration operations. The key idea is to decouple the relationship between the eigenvalues and the membrane time $\tau_m$, enabling us to achieve adaptive eigenvalues by training self-connection parameters.

\subsection{Mutual Promotion and Back-Propagation of Adding Self-connection Architecture}
We begin with the basic computation of SNNs equipped with self-connection architectures as follows
\begin{equation} \label{eq:ScSNN}
	\frac{\dif \boldsymbol{u} (t)}{\dif t} = -\frac{\boldsymbol{u}(t)}{\tau_m} + \boldsymbol{u}^*(\mathbf{V},t) + \frac{\tau_r}{\tau_m} f_{\textrm{agg}}(\mathbf{I}(t) )  \ .
\end{equation}
In contrast to the computations of LIF equations, Eq.~\eqref{eq:ScSNN} employs an extra vector $\boldsymbol{u}^*(\mathbf{V},t) = (u_1^*, \dots, u_N^*)^{\top}$ portrays the mutual promotion between neurons adjusted by self-connection parameters $\mathbf{V}$. Here, we provide two intuitive implementations for $\boldsymbol{u}^*(\mathbf{V},t)$ and omit $t$ for simplicity
\begin{equation} \label{eq:ScSNN_mutual}
	\left\{~\begin{aligned}
		\text{Linear}\quad:&~ u_k^*(\boldsymbol{\mathbf{V}},t) = \sum_{i \neq k} \mathbf{V}_{ki} u_i + o(|\boldsymbol{u}|) \ , \text{~\citep{zhang2021:bsnn}} \\
		\text{Polynomial}:&~ u^*_k(\mathbf{V},t) = \sum_{i} \mathbf{V}_{ki}^{(1)} u_i + \sum_{p=2}^{n} \left\langle~ \mathbf{V}_{k}^{(p)}, \textrm{P}_p(\boldsymbol{u}) ~\right\rangle + o(|\boldsymbol{u}|^p) \ , 
	\end{aligned}\right.
\end{equation}
for $k\in[N]$, where $o(|\boldsymbol{u}|^p)$ denotes a high-order over the $n$-order polynomial $|\boldsymbol{u}|^p$ for $p \in \mathbb{N}^+$, $\mathbf{V}_{k}^p$ indicates a vector $(\mathbf{V}_{k,j}^{(p)})_{j\in |\Lambda|}$, and $\textrm{P}_p(\boldsymbol{u}) = (u_1^{\alpha_1} u_2^{\alpha_2} \dots u_N^{\alpha_N})_{(\alpha_1,\dots,\alpha_N) \in \Lambda}$ indicates another one in which $|\boldsymbol{\alpha}| = \alpha_1 + \alpha_2 + \dots + \alpha_N = p$. Taking an example of $N=2$ and $n=2$, we have
\begin{equation} \label{eq:ScSNN_planar}
	\left\{~\begin{aligned}
		u_1^*(\mathbf{V},t) 
		&= \mathbf{V}_{11}^{(1)}u_1 + \mathbf\mathbf{V}_{12}^{(1)} u_2 
		+ \mathbf{V}^{(2)}_{11} u_1^2 + \mathbf{V}^{(2)}_{22}  u_2^2 + \mathbf{V}^{(2)}_{12} u_1 u_2 \ , \\
		u_2^*(\mathbf{V},t)
		&= \mathbf\mathbf{V}_{21}^{(1)} u_1 + \mathbf{V}_{22}^{(1)} u_2 
		+ \mathbf{V}^{(2)}_{11} u_1^2 + \mathbf{V}^{(2)}_{22} u_2^2 + \mathbf{V}^{(2)}_{12} u_1 u_2 \ .
	\end{aligned}\right.
\end{equation}

\vspace{0.2cm}
\noindent\textbf{Key Ideas of Adding Self-connection.} The motivation for adding self-connection to SNNs is to enhance the adaptivity of the eigenvalues of integration operations. Essentially, our objective is to guarantee that the integration operation of SNNs possesses dynamically adjustable eigenvalues. One viable way is to endue the integration operation with learnable parameters, as outlined in the subsequent algebraic equation
\begin{equation}  \label{eq:algebraic_ScSNN}
	\frac{\dif \boldsymbol{u}}{\dif t} = \mathbf{M}(\mathbf{V},\tau_m) \boldsymbol{u} + G(\boldsymbol{u},\boldsymbol{\lambda})
	\quad\text{with}\quad
	G(\boldsymbol{u},\boldsymbol{\lambda})=o(|\boldsymbol{u}|) \ ,
\end{equation}
where $\mathbf{M}_1(\tau_m) = \diag\{-1/\tau_m, \dots, -1/\tau_m\}_{N}$ and $\mathbf{M}_2(\mathbf{V})$ is relative to the learnable parameter $\mathbf{V}$. So the eigenvalue $\rho_i$ of $\mathbf{M}(\mathbf{V},\tau_m)$ can be calculated as the sum of that of $\mathbf{M}_1(\tau_m)$ and that of $\mathbf{M}_2(\mathbf{V})$. Suppose that the eigenvalues of the matrix $\mathbf{M}_2(\mathbf{V})$ are $\beta_1,\dots,\beta_N$. Then we have
\begin{equation} \label{eq:ScSNN_eigenvalues}
	\rho_i = -1/\tau_m + \beta_i  \ ,
\end{equation}
in which both $\tau_m$ and $\mathbf{V}$ adjust the eigenvalues. Since the self-connection weights are learnable, the eigenvalues can be adaptive to the changing environment over time even when $\tau_m$ remains constant. Thus, the key idea of adding self-connection is to decouple the strong dependency between the eigenvalues and the membrane time $\tau_m$, allowing us to obtain adaptive eigenvalues through the adjustment of the learnable parameters $\mathbf{V}$. The bounds of Theorem~\ref{thm:bifurcation_bound} in the next subsection further confirm this conjecture. Further, it is promising to manage the risk of unstable structure, which will be investigated in Subsection~\ref{subsec:ScSNN_general}.

\vspace{0.2cm}
\noindent\textbf{Error Back-Propagation.} We here provide a concrete scheme for implementing SNN with self-connection architectures. This work considers $M$ pre-synaptic input channels and $N$ hidden spiking neurons. Formally, we have the following equation for neuron $k\in[N]$
\begin{equation} \label{eq:ScSNN_general_feedforward}
	\frac{\dif u_k(t)}{\dif t} = -\frac{u_k(t)}{\tau_m} + u^*_k(\mathbf{V},t) +  \frac{\tau_r}{\tau_m} \sum_{j =1}^M \mathbf{W}_{kj} \mathbf{I}_j(t) \ ,
\end{equation}
which has two types of learnable parameters, i.e., self-connection weights $\boldsymbol{\mathbf{V}}$ and connection weights $\mathbf{W}$. Akin to the spike response model scheme~\citep{gerstner1995:srm}, Eq.~\eqref{eq:ScSNN_general_feedforward} has a closed-form solution
\begin{equation} \label{eq:ScSNN_general_solution}
	u_k(t) = \sum_{s=t'}^t \exp\left( -\frac{t'-s}{\tau_m} \right)  ~\Delta(s) 
	\quad\text{with}\quad
	\Delta(s) = u^*_k(\mathbf{V},s) +  \frac{\tau_r}{\tau_m} \sum_{j=1}^M \mathbf{W}_{kj} \mathbf{I}_j(s) \ ,
\end{equation}
Finally, the generated spike is transmitted to the next neuron via the spike excitation function $f_e: u \mapsto s$. Provided supervised signals, the proposed model can be optimized via the framework of error back-propagation. The temporal-accumulated error in the discrete-time interval $[1:T]$ can be formulated by
\begin{equation} \label{eq:cost}
	E  = \frac{1}{2} \sum_{t=1}^{T} \sum_{k=1}^N E_k(t) 
	= \frac{1}{2} \sum_{t=1}^{T} \mathscr{L} \left( \boldsymbol{o} (t), \hat{\boldsymbol{o}} (t) \right) \ ,
\end{equation}
where $\mathscr{L}$ indicates the loss function, such as the least square loss and 0-1 loss functions, and $\hat{\boldsymbol{o}} (t)$ denotes the target supervised signal related to the prediction signal $\boldsymbol{o} (t)$. So for time $t$, we have
\begin{equation} \label{eq:BP_ScSNN_1}
	\frac{\partial E_k (t)}{\partial \mathbf{W}_{kj}} 
	= \frac{\partial E_k (t)}{\partial o_k (t)} 
	\frac{\partial o_k (t)}{\partial u_k (t)} 
	\frac{\partial u_k (t)}{\partial \mathbf{W}_{kj}} \ ,
\end{equation}
where the first term is the error back-propagation of the excitatory neurons, the second term is that of the generated spikes with respect to the membrane potential, and the third term denotes that of the basic bifurcation neuron. Plugging Eq.~\eqref{eq:ScSNN_general_solution} and Eq.~\eqref{eq:cost} into Eq.~\eqref{eq:BP_ScSNN_1}, we have
\[
\frac{\partial E_k (t)}{\partial \mathbf{W}_{kj}} = \bigg(o_k (t) - \hat{o}_k (t) \bigg)
f_e'\bigg( u_k (t) \bigg) \frac{\tau_r}{\tau_m}
\left[ \sum_{s = t'}^t \exp\left( -\frac{s-t'}{\tau_m} \right) \mathbf{I}_j(s) \right] \ ,
\]
where $t'$ denotes the last firing time. However, the derivative of the spike excitation function $f_e'(u)$ is a persistent problem for training SNNs with supervised signals. Recently, there have emerged many seminal approaches for addressing this problem, such as the smoothing derivative via the probability density functions~\citep{shrestha2018:slayer}, modified spike excitation functions~\citep{zhang2022:SNNsTheory}, and the stochastic excitation mechanism proposed in Section~\ref{sec:stoc}. Therefore, we obtain the back-propagation pipeline relative to connection weights $\mathbf{W}_{kj}$.

Similarly, the correction with respect to some element $\mathbf{V}$ is given by
\[
\nabla_{\mathbf{V}} E _k = 
\sum_{t=1}^{T} \bigg(o_k (t) - \hat{o}_k (t) \bigg)
f_e'\bigg( u_k (t) \bigg) \frac{1}{\tau_m} \frac{\partial u^{*}_k(\mathbf{V},t) }{\partial \mathbf{V}} 
\exp\left( \frac{-t}{\tau_m} \right)  \ .
\]
If $\partial u^{*}_k(\mathbf{V},t) / \partial \mathbf{V}$ indicates a linear partial derivative, we further have
\[
\nabla_{\mathbf{V}} E _k = 
\sum_{t=1}^T \bigg(o_k (t) - \hat{o}_k (t) \bigg)
f_e'\bigg( u_k (t) \bigg) \frac{s_k (t)}{\tau_m} \exp\left( \frac{-t}{\tau_m} \right) 
\quad\text{when}\quad
n=1 \ .
\]

\subsection{Expressivity of Adding Self-connection Architectures} \label{subsec:approximation_ScSNN}
This subsection shows the expressive powers of SNNs with self-connection. The first conclusion is about the universal approximation.
\begin{theorem} \label{thm:ua_ScSNN}
	Let $K \subset \mathbb{R}^M$ be a compact set, and $K_0$ is a null set. Provided $l\in\mathbb{N}^+$, if the spike excitation function $f_e$ is $l$-times differentiable on $K/K_0$ that satisfies
	\[
	0 < \left| \int_{K/K_0} D^r f_e(u) \dif u \right| < \infty \ ,
	\quad\text{for any $r \in [l]$}   \ ,
	\]
	and $\mathbf{W} \in \mathbb{R}^{N \times M}$, $\boldsymbol{w} \in \mathbb{R}^{N \times 1}$, then there exists some time $t$ such that the set of functions $f(\boldsymbol{\cdot}, t): K \rightarrow \mathbb{R}$ expressed by a SNN with the two-layer self-connection architecture and linear mutual promotion, which is of the form 
	\begin{equation}  \label{eq:express_ScSNN}
		\left\{~\begin{aligned}
			f(\cdot,t)  &= \frac{1}{t-t'}~ \boldsymbol{w}^{\top} \boldsymbol{f}_{\textrm{hidden}}(\cdot, t)  \  , \\
			f_k(\cdot,t) &= f_e \left( \mathbf{W}_{k,[M]} \int_{t'}^t \exp\left(  - \frac{t'' - t'}{\tau_m} \right) \mathbf{I}(t'') \dif t''  -  \frac{1}{\tau_m} \sum_{i\in[N]} \exp\left(  -\frac{t-t'}{\tau_m} \right) \mathbf{V}_{ki} s_i(t') \right) \ ,
		\end{aligned}\right.
	\end{equation}
	where $\boldsymbol{f}_{\textrm{hidden}}= (f_1, \dots,f_N)$ and $k\in [N]$, is dense in $\mathcal{C}^0(K,\mathbb{R})$.
\end{theorem}
Theorem~\ref{thm:ua_ScSNN} shows that the SNN with self-connection architectures is a universal approximator, which provides a solid cornerstone for the expressive power of SNNs. We utilize the invertibility of the Fourier transform on Sobolev space $\mathcal{W}^{l,p}_{\mu}(K,\mathbb{R})$ $(p>1)$, to project the concerned functional space $\mathcal{C}^r(K,\mathbb{R})$ into a characteristic space, and the corresponding objective function is transformed as a single integral over the characteristic space. According to Fubini's theorem, the approximation problem on $\mathcal{C}^r(K,\mathbb{R})$ can be converted into another that uses multiple integrals to construst a single integral on the characteristic space. The subsequent proof can then be completed along the thought lines of the technical exposition given by~\citet{carslaw1931:fourier}. The full proof of Theorem~\ref{thm:ua_ScSNN} can be obtained in Appendix~\ref{app:ua_ScSNN}.

\vspace{0.2cm}
Notice that Theorem~\ref{thm:ua_ScSNN} also holds for SNNs with self-connection architectures and polynomial mutual promotion. The second conclusion is about the adaptive eigenvalues adjusted by $\mathbf{V}$. Correspondingly, we can define a ``legal" energy function and its derivative by
\begin{equation}  \label{eq:energy_network_ScSNN}
	\mathcal{H}(t) = |\boldsymbol{u}|^2 + \frac{2\tau_r}{\tau_m} \int \left\langle \frac{\partial \boldsymbol{v}_{\textrm{sc}}}{\partial t},  \boldsymbol{u}(t) \right\rangle  \dif t - \theta
	\quad\text{with}\quad
	\frac{\partial \boldsymbol{v}_{\textrm{sc}}(t)}{\partial t} = \boldsymbol{u}^*(\boldsymbol{\mathbf{V}},t) + f_{\textrm{agg}} (\mathbf{I}(t)) 
\end{equation}
and
\begin{equation}  \label{eq:H_ScSNN_der}
	\frac{\dif \mathcal{H}}{\dif t} = \frac{1}{2} \boldsymbol{u}^{\top} ~\mathbf{M}(\boldsymbol{\mathbf{V}},\tau_m)~ \boldsymbol{u} \ ,
\end{equation}
respectively, with
\[
\mathbf{M}(\boldsymbol{\mathbf{V}},\tau_m) = \begin{pmatrix}
	\mathbf{V}_{11}-{1}/{\tau_m}      & \mathbf\mathbf{V}_{12}    & \dots   & \mathbf{V}_{1N} \\
	\mathbf\mathbf{V}_{21} &    \mathbf{V}_{22}-{1}/{\tau_m}     & \dots   & \mathbf{V}_{2N} \\
	\vdots    & \vdots       & \ddots   & \vdots       \\
	\mathbf{V}_{N1} & \mathbf{V}_{N(N-1)}    & \dots    &  \mathbf{V}_{NN}-{1}/{\tau_m}
\end{pmatrix} \ .
\]
Based on Eq.~\eqref{eq:energy_network_ScSNN} and Eq.~\eqref{eq:H_ScSNN_der}, we have
\begin{theorem}  \label{thm:bifurcation_bound}
	Provided the initial condition $u(0) = u_{\textrm{reset}}$ or $u(t') = u_{\textrm{reset}}$, the SNN with the self-connection architecture and linear mutual promotion leads to an adaptive Hamiltonian system. Especially there are
	\begin{itemize}
		\item[(i)] at most $2^{N-1}$ solutions if $\mathbf{V}_{ij} \geq 0$ for $i,j \in [N]$ \cite[Theorem 2]{zhang2021:bsnn};
		\item[(ii)] are at least $c N \log N$ solutions ($0 < c \leq 1/2$) .
	\end{itemize}
\end{theorem}
Theorem~\ref{thm:bifurcation_bound} shows that adding self-architectures contributes to enough solutions, bounded between polynomial (lower bound $c N \log N$) and exponential (upper bound $2^{N-1}$) numbers, for approximating an adaptive Hamiltonian system as well as maintaining adaptive eigenvalues. The proof of Theorem~\ref{thm:bifurcation_bound} can be accessed in Appendix~\ref{app:bifurcation_bound}.

\begin{figure}[t]
	\centering
	\includegraphics[width=0.75\textwidth]{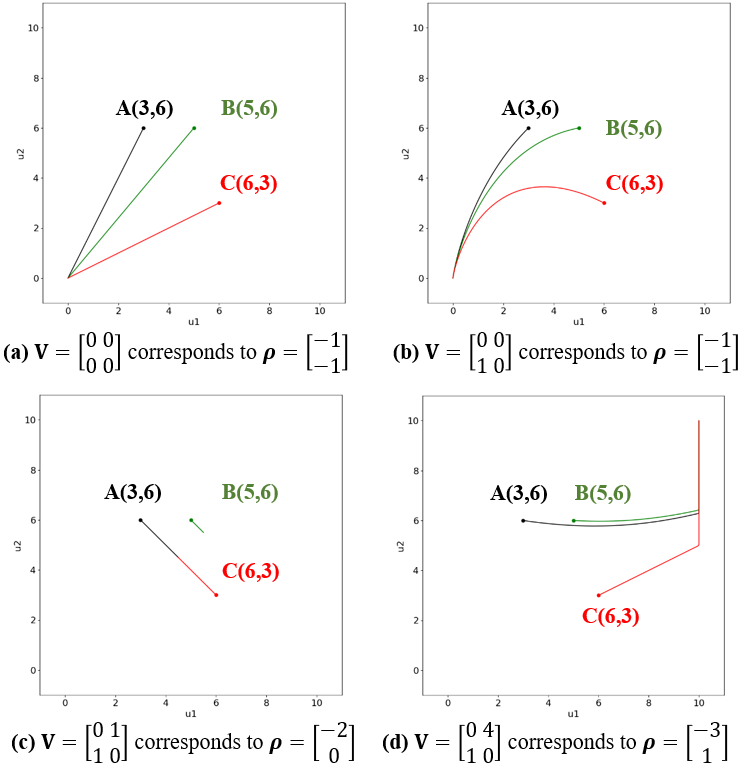}
	\caption{Illustrations for SNNs with/without self-connection conquering bifurcation.}
	\label{fig:LinearScSNNs}
\end{figure}
\vspace{0.2cm}
\noindent\textbf{Simulation Experiment.} Here, we further investigate the effects of $\mathbf{V}$ on the expressivity of adding self-connection architectures. We take a simulation experiment of two self-connection spiking neurons with $\tau_m = \tau_r = 1$, $u_{\textrm{reset}} = 0$, and $u_{\textrm{firing}}=10$ and initialize three points: $A(3,6)$, $B(5,6)$, and $C(6,3)$ at the starting timestamp $t=0$. We conduct four trials with
\[
\mathbf{V} = \begin{bmatrix}
	0 & 0 \\
	0 & 0
\end{bmatrix},\quad \begin{bmatrix}
	0 & 0 \\
	1 & 0
\end{bmatrix},\quad \begin{bmatrix}
	0 & 1 \\
	1 & 0
\end{bmatrix},\quad \begin{bmatrix}
	0 & 4 \\
	1 & 0
\end{bmatrix} \ ,
\] 
where their eigenvalues correspond to 
\[
\boldsymbol{\beta} = \begin{bmatrix}
	0  \\
	0
\end{bmatrix},\quad \begin{bmatrix}
	0  \\
	0 
\end{bmatrix},\quad \begin{bmatrix}
	-1  \\
	1 
\end{bmatrix},\quad \begin{bmatrix}
	-2  \\
	2 
\end{bmatrix} \ .
\] 
According to $\boldsymbol{\rho} = [-1/\tau_m ; -1/\tau_m] + \boldsymbol{\beta}$, we can obtain the eigenvalues of integration operations as follows
\[
\boldsymbol{\rho} = \begin{bmatrix}
	-1  \\
	-1
\end{bmatrix},\quad \begin{bmatrix}
	-1  \\
	-1 
\end{bmatrix},\quad \begin{bmatrix}
	-2  \\
	0 
\end{bmatrix},\quad \begin{bmatrix}
	-3  \\
	1 
\end{bmatrix} \ .
\] 
In Figure~\ref{fig:LinearScSNNs}, the plotted trajectories exhibit distinctive behaviors over time. It is observed that different eigenvalues yield various trends; zero eigenvalue leads to a line parallel to axes, while negative and positive eigenvalues make the curves converge to 0 (corresponding to $u_{\textrm{reset}}$) and 10 (corresponding to $u_{\textrm{firing}}$), respectively.

\begin{figure}[t]
	\centering
	\includegraphics[width=0.95\textwidth]{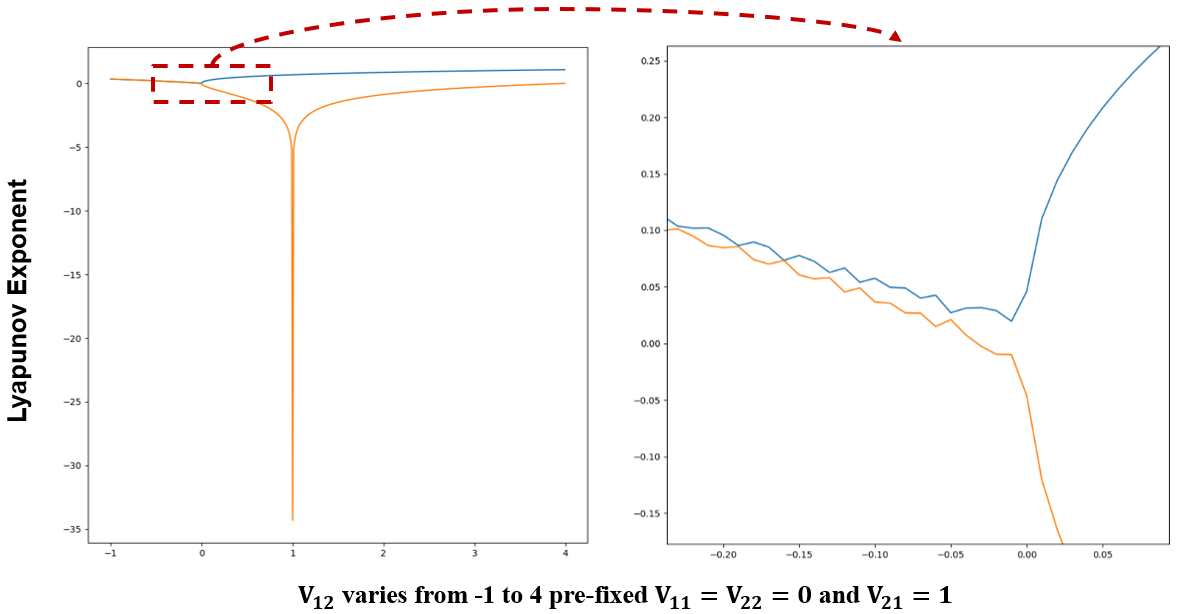}
	\caption{Lyapunov exponent and bifurcation diagram of SNNs with self-connection.}
	\label{fig:Ly}
\end{figure}

To further leverage the effect of $\mathbf{V}$, we fix $\mathbf{V}_{11} = \mathbf{V}_{22} = 0$ and $\mathbf\mathbf{V}_{21} = 1$ and then investigate the bifurcation diagram of SNNs. Figure~\ref{fig:Ly} plots the curves of the Lyapunov exponent as $\mathbf\mathbf{V}_{12}$ varies in the $x$-axis interval of $[-1,4]$. Notice that in the bifurcation diagram given $\mathbf\mathbf{V}_{21}=1$, two critical points emerge: $\mathbf\mathbf{V}_{12}=0$ and $\mathbf\mathbf{V}_{12}=1$, which split the eigenvalue interval as $(-\infty, -1] \cup [-1, 0] \cup [0, +\infty)$. With an increasing $\mathbf\mathbf{V}_{12}$ as well as one of the eigenvalues, the system state is constantly changing, giving rise to intricate dynamic behaviors such as bifurcation, chaos, and limit cycles. It is observed that some periodic states are embedded within the chaotic (i.e., unstable) states.

The related computations of this simulation experiment can be obtained in Appendix~\ref{app:ode_simulation}.

\subsection{Structural Stability of Adding Self-connection Architectures}  \label{subsec:ScSNN_general}
\begin{figure}[t]
	\centering
	\includegraphics[width=0.95\textwidth]{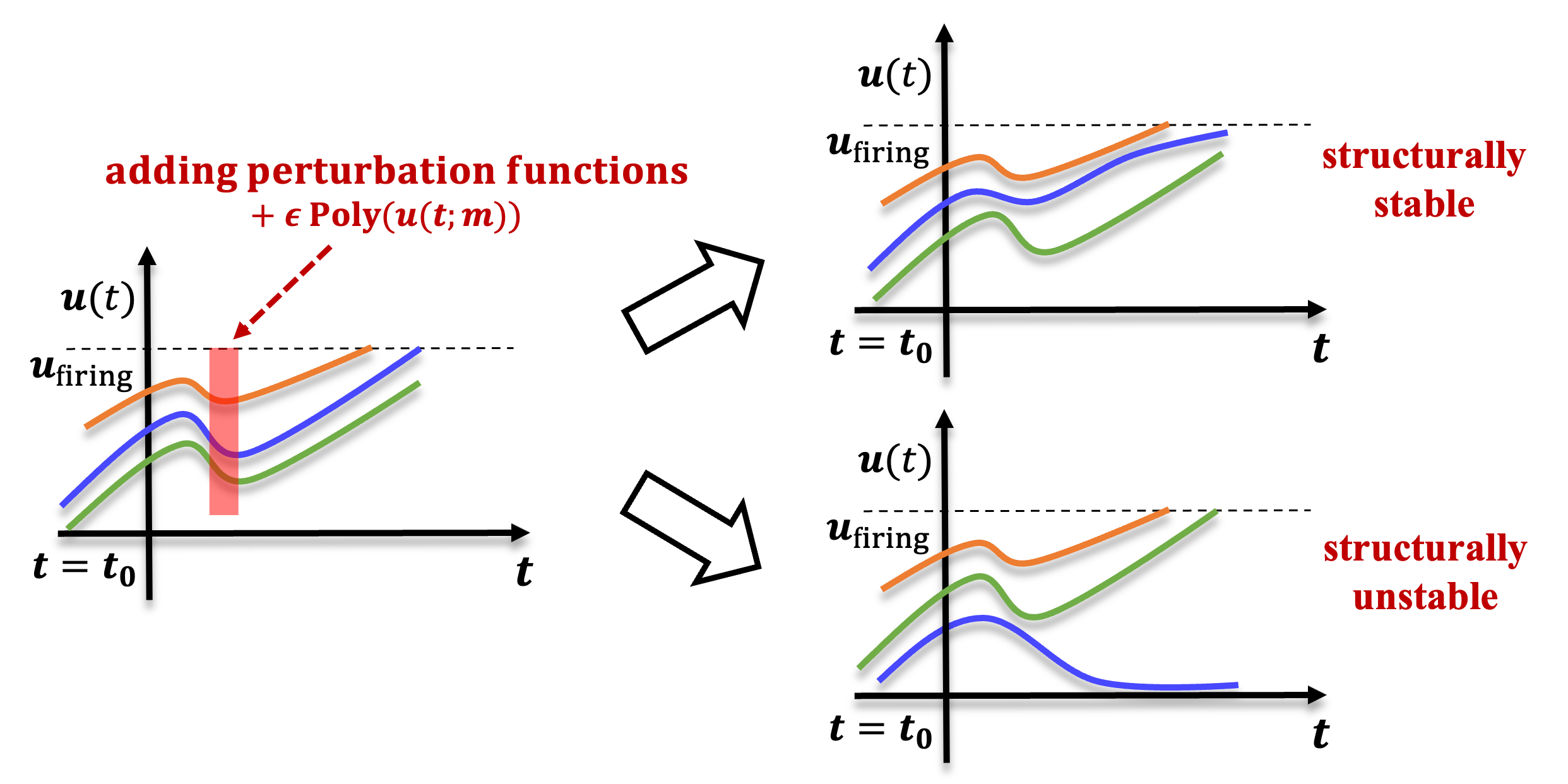}
	\caption{The vector-field plots for bifurcation and structural stability.}
	\label{fig:SS_for_bifurcation}
\end{figure}

In this subsection, we aim to delve into the structural stability associated with the inclusion of self-connection architectures. The fundamental way of identifying structural stability involves adding the (functional) perturbations led by a small parameter $\epsilon$ to a critical point\footnote{This point is sometimes referred to as an equilibrium point or an equilibrium function. Please refer to~\citep{bavzant2000:ss} for details.} of the algebraic equation. The subsequent step is to observe whether the perturbed system bifurcates from the critical point or from some periodic orbits surrounding the critical point; the latter here is called \emph{stable bifurcation solution}, which intrinsically indicates a kind of functional equivalence classes~\citep{kim2021:stability}. This is illustrated in Figure~\ref{fig:SS_for_bifurcation}. Unfortunately, counting the stable bifurcation solutions of multivariate time-varying dynamical systems, at the present state of knowledge, seems to be hopeless~\citep{llibre2015:hilbert}. Here, we simplify this issue by counting (local) limit cycles, which equivalently represent a type of stable bifurcation solution. This choice is justified by the fact that a limit cycle of planar polynomial time-varying dynamical systems in Eq.~\eqref{eq:ScSNN_mutual} essentially constitutes an isolated periodic orbit~\citep{christopher2007limit}. For convenience, we denote $u^*_i(\mathbf{V},t)$ as $u^*_i(\mathbf{V},t) = \textrm{Poly}_i(\boldsymbol{u}(t);n)$ and formally exhibit the algebraic form of Eq.~\eqref{eq:ScSNN} as follows
\begin{equation}  \label{eq:algebraic}
	\frac{\dif u_i(t)}{\dif t} = -\frac{u_i(t)}{\tau_m} + \textrm{Poly}_i(\boldsymbol{u}(t);n)  
\end{equation}
for $ i \in [N]$ and $N\geq 2$. The corresponding perturbed system becomes
\begin{equation}  \label{eq:perturbed}
	\frac{\dif u_i(t)}{\dif t} = -\frac{u_i(t)}{\tau_m} + \textrm{Poly}_i(\boldsymbol{u}(t);n) + \epsilon ~\textrm{Poly}_i(\boldsymbol{u}(t);m) \ ,
\end{equation}
where $\epsilon$ indicates a small parameter that scales the perturbation magnitude of degree $m$. Here, we are interested in the small limit cycles of Eq.~\eqref{eq:perturbed}, which bifurcate at $\epsilon$ from the critical points of Eq.~\eqref{eq:algebraic} as $\epsilon \to 0$. We employ $H(n)$ to denote the maximum number of limit cycles of Hamiltonian systems with polynomials of degree $n$ in Eq.~\eqref{eq:ScSNN_mutual}.

In the rest of this subsection, we first identify the non-negativity of $H(n)$ in Theorem~\ref{thm:ScSNN_periodic}, and then explore the explicit bounds of $H(n)$. However, it is a tricky challenge to tighten $H(n)$ that corresponds to Eq.~\eqref{eq:ScSNN} in confronted of dynamical systems, which coincides with the second part of Hilbert's 16$^{\textrm{th}}$ problem. In the near past, it has not been possible to find uniform upper bounds for $H(n)$, referring to the knowledge of~\citet{llibre2015:hilbert}. Thus, we present a calculable approach for computing the upper bound of $H(n)$ rather than finding the explicit one and then provide a provable lower bound of $H(n)$ in Theorem~\ref{thm:ScSNN_lower}.

\subsubsection{Existence}
Now, we present the existence theorem as follows.
\begin{theorem} \label{thm:ScSNN_periodic}
	Let $\tilde{u}$ be a critical point of system~\eqref{eq:perturbed}. For $\epsilon>0$ sufficiently small, there exists a ($2\pi$-periodic) stable bifurcation solution $f(t,\epsilon)$ of system~\eqref{eq:perturbed} s.t. $f(0,\epsilon) \to \tilde{u}$ as $\epsilon \to 0 $.
\end{theorem}
Theorem~\ref{thm:ScSNN_periodic} shows the existence of ($2\pi$-periodic bifurcation) limit cycles as well as stable bifurcation solutions of the perturbed dynamical system, which implies that $H(n) \geq 0$. This result holds from the following useful lemmas, while the complete proof of Theorem~\ref{thm:ScSNN_periodic} can be accessed in Appendix~\ref{app:periodic}.
\begin{lemma}  \label{lemma:recurrence_upper}
	The perturbed system~\eqref{eq:perturbed} induces a planar differential equation as follows
	\begin{equation}  \label{eq:per_step_1}
		\frac{\partial f(t,\epsilon)}{\partial t} = \sum_{k=0}^K \epsilon^k F_k(t,f) + \epsilon^{K+1} \mathrm{reset}(t,f,\epsilon) 
		\quad\textrm{(normal form)} \ ,
	\end{equation}
	with
	\begin{equation}  \label{eq:O1}
		\int_0^T \mathrm{reset}(s,f,\epsilon) \dif s  = \mathcal{O}(1) \ ,
	\end{equation}
	where $F_k: \mathbb{R} \times \mathbb{K} \to \mathbb{R}$ and $\mathrm{reset}: \mathbb{R}\times \mathbb{K} \times [-\epsilon_0,\epsilon_0] \to \mathbb{R} $ are $\mathcal{C}^k$-continuous functions in which $\mathbb{K}$ denotes the functional space, $k = 0, 1,2 \dots, K$, and $\epsilon_0 \geq 0$.
\end{lemma}
Lemma~\ref{lemma:recurrence_upper} essentially is a Taylor expansion of ${\partial f(t,\epsilon)}/{\partial t}$ where each component $F_k(t,f)$ computes the estimation of degree $k$ for the concerned perturbed system, leading to an equivalent formation of the concerned system~\eqref{eq:perturbed}. The formulas of Lemma~\ref{lemma:recurrence_upper} contribute to the periodic solutions of recursive formations, as shown in the following lemma. Notice

\begin{lemma}  \label{lemma:fundamental}
	Suppose that $\tilde{u}$ and $f(t,\epsilon): [0,T] \times [-\epsilon_0,\epsilon_0] \to \mathbb{R}$ be the critical point and solution of system~\eqref{eq:perturbed}, respectively, satisfying that $f(0,\epsilon) = \tilde{u}$. Then for $t \in [0,T]$, we have
	\[
	\begin{aligned}
		f(t,\epsilon) = \tilde{u} + \int_0^t F_0(s,\tilde{u}) \dif s + \sum_{k=0}^K \epsilon^k G_k(t,\tilde{u}) 
		+ \epsilon^{K+1} \left[ \int_0^t \mathrm{reset}(s, f(s,\epsilon),\epsilon) \dif s + \mathcal{O}(1) \right] \ ,
	\end{aligned}
	\]
	where $G_k$ (for $k = 0, 1, 2 ,\dots, K$) is of recursive form as follows
	\[
	G_k(t,u) = \int_0^t \left[ F_k(s,u) + \mathcal{G}_k \left( F_r(s,u), G_r(s,u) \right) \right] \dif s \ ,   
	\]
	in which
	\[
	\mathcal{G}_k =  \sum_{r=1}^{k-1} \sum_{\boldsymbol{\alpha} \in \mathcal{S}_r} \frac{D^{|\boldsymbol{\alpha}|} F_{k-r}(s,u)}{\alpha_1!(\alpha_2!2!^{\alpha_2}) \dots (\alpha_r!r!^{\alpha_r})} \prod_{l=1}^r G_l(s,u)^{\alpha_j}  \ ,
	\]
	where $\boldsymbol{\alpha} = (\alpha_1, \dots, \alpha_r) \in \mathcal{S}_r$ and $\mathcal{S}_r$ denotes the set of all $r$-tuples of non-negative integers $\{\alpha_j\}_{j\in[r]}$ that satisfies
	\[
	\sum_j j\alpha_j = r \ .
	\]
\end{lemma}
Lemma~\ref{lemma:fundamental} provides the existence and the recursive formation of the concerned periodic solutions (i.e., limit cycles) of system~\eqref{eq:perturbed}.

\subsubsection{Provable Lower Bound}
Now, we present the lower bound theorem as follows.
\begin{theorem}  \label{thm:ScSNN_lower}
	Let $H(n)$ denote the maximum number of limit cycles of dynamical systems with $n$-order polynomial implementation in Eq.~\eqref{eq:ScSNN_mutual}. Then we have $H(n) = \Omega(n^2\ln n)$.
\end{theorem}
Theorem~\ref{thm:ScSNN_lower} shows the lower bound of $H(n)$ of the dynamical system led by Eq.~\eqref{eq:ScSNN}. In detail, we have $H(1) \geq 0$, $H(3) \geq 1$, $H(7) \geq 25$, $H(15) \geq 185$, and $H(31) \geq 1262$.

The complete proof of Theorem~\ref{thm:ScSNN_lower} can be accessed in Appendix~\ref{app:bound}, and its proof idea can be summarized as follows. According to Subsection~\ref{subsec:ScSNN_general}, the lower bound of $H(n)$ coincides with the maximum possible number of limit cycles of the dynamical system led by Eq.~\eqref{eq:ScSNN}. Hence, a key intuition of bounding $H(n)$ is to reformulate the lower bound into a recursive formation. It is observed that equipped with polynomial of degree $n$ in Eq.~\eqref{eq:ScSNN_mutual}, Eq.~\eqref{eq:ScSNN} leads to a Hamiltonian system with perturbation $\epsilon$, 
\begin{equation} \label{eq:H}
	\left\{\begin{aligned}
		\frac{\dif u_k(t)}{\dif t} &= - \frac{\partial \mathcal{H}(u_k, u_{k'})}{\partial u_{k'}} + \epsilon f_{\epsilon}(u_k, u_{k'}) \\
		\frac{\dif u_{k'}(t)}{\dif t} &= \frac{\partial \mathcal{H}(u_k, u_{k'})}{\partial u_k} + \epsilon g_{\epsilon}(u_k, u_{k'}) 
	\end{aligned}\right.  \quad\text{for $k,k' \in [N]$} \ ,
\end{equation}
where $\mathcal{H}(u_k, u_{k'}) = \textrm{Poly}(u_k)^2 + \textrm{Poly}(u_{k'})^2$ indicates the Lyapunov-like energy function, $\epsilon \in (0,\infty)$ denotes the noise amplitude, $f_{\epsilon}(u_k,u_k')$ and $g_{\epsilon}(u_k,u_k')$ are two polynomial functions of degree $2^n -1$ with respect to $u_k$ and $u_k'$. Therefore, $H(n)$ meets a recursive formation as follows:
\[
H(2^{n+1}-1) = 4 H(2^n-1) + (2^n - 2)^2 + (2^n-1)^2 \ .
\]
With straightforward computations, we can conclude that there exists a constant $C>0$ such that
\[
H(n) \geq C (n+1)^2 \ln(n+1) \ .
\]

\subsubsection{Algorithmic Upper Bound}
From Lemma~\ref{lemma:fundamental}, we specify the general solution of which the $k^{\textrm{th}}$ component $G_k(t,u)$ is of a recursive form. So it is feasible to simulate $G_k(t,u)$ algorithmically. Further, we can obtain the limit cycle $f(t,\epsilon)$, and then, the upper bound of $H(n)$ can be calculable easily. Inspired by this recognition, we present the algorithm for calculating the $k^{\textrm{th}}$ component of limit cycles. The procedure is listed in Algorithm~\ref{alg:algorithm_for_upper_bound}, which comprises the following four steps: 
\begin{enumerate}
	\item Simulate Eq.~\eqref{eq:per_step_1} with $K^{\textrm{th}}$ order and $\epsilon$ from the perturbed system~\eqref{eq:perturbed} in \textbf{Procedure 1-3}.
	\item Formulate the exact formula of $F_k(t,\epsilon)$ for $k \in [K]$ in \textbf{Procedure 4}.
	\item Compute the approximation to $G_k(t,\epsilon)$ relative to ${\partial^r F_k}/{\partial \epsilon^r}$ and $\textrm{reset}(t,u,\epsilon)$ for $k \in [K]$ and $r \in [k]$ in \textbf{Procedure 5-12}.
	\item  Calculate the upper bound of $H(n)$ using the number of positive simple critical points of $G_k(t,\epsilon)$ for $k \in [K]$ in \textbf{Procedure 13}.
\end{enumerate}

\begin{algorithm}[t]
	\caption{Algorithmic Calculation for Upper Bounds of $H(n)$.}
	\label{alg:algorithm_for_upper_bound}
	\begin{algorithmic}
		\renewcommand{\algorithmicrequire}{\textbf{Input:}}
		\REQUIRE
		The number of spiking neurons $N=2$, the polynomial degree $n$, the polynomial degree $m$ relative to perturbations, the estimation degree $K$, apposite perturbation $\epsilon$
		\renewcommand{\algorithmicensure}{\textbf{Output:}}
		\ENSURE
		$k^{th}$ component $G_k(t, \epsilon)$ for $k\in[K]$, bifurcation solution $f(t,\epsilon)$
		\renewcommand{\algorithmicrequire}{\textbf{Procedure:}}
		\REQUIRE ~\\
		\STATE
		\textbf{1}: Generate $\textrm{Poly}(\boldsymbol{u}(t);n)$ and $\textrm{Poly}(\boldsymbol{u}(t);K)$ in a feed-forward way
		\STATE
		\textbf{2}: Compute $\frac{\dif u_1(t)}{\dif t}$ and $\frac{\dif u_2(t)}{\dif t}$ from Eq.~\eqref{eq:perturbed}
		\STATE
		\textbf{3}: Convert the perturbed system~\eqref{eq:perturbed} into $\frac{\partial f(t,\epsilon)}{\partial t}$ in Eq.~\eqref{eq:per_step_1}
		\STATE
		\textbf{4}: Compute functions $F_k$ for $k=0$ or $k\in[K]$
		\STATE
		\textbf{5}: Let $\delta F = 0$
		\STATE
		\textbf{6}: for $k$ from 1 to $K-1$ do:
		\STATE
		\textbf{7}: \quad for $r$ from 1 to $K-k$ do:
		\STATE
		\textbf{8}: \quad\quad\quad $\delta F \gets \delta F + \frac{\epsilon^r}{r!} \frac{\partial^r F_k(t,f(t,\epsilon,u))}{\partial \epsilon^r} \big|_{\epsilon=0} $
		\STATE
		\textbf{9}: \quad Re-compute $F_k$ provided $\delta F$ according to $F_k(t,f(t,\epsilon,u)) = F_k(t,f(t,0,u)) + \delta F$
		\STATE
		\textbf{10}: ~ Compute $G_k$ provided $\delta F$ and $F_k$ according to 
		\STATE
		\qquad\qquad $ G_k(t,u) = \int_0^t \left[ F_k(s,u) + \mathcal{G} \left( D^{h(r)} F_r(s,u), G_r(s,u) \right) \right] \dif s $
		\STATE
		\textbf{11}: Compute $\delta R \gets \int_0^{2\pi} \mathrm{reset}(s, u,\epsilon) \dif s$ by sampling $u \in \mathbf{K}$
		\STATE
		\textbf{12}: Compute $f$ provided $G_k$, $F_k$, and $\delta R$ according to 
		\STATE
		\qquad\quad $f(t,\epsilon, u) = u + \int_0^t F_0(s,\tilde{u}) \dif s + \sum_{k=0}^K \epsilon^k G_k(t,u) + \delta R$
		\STATE
		\textbf{13}: \textbf{return} $f$ and $G_k(t, \epsilon)$
	\end{algorithmic}
\end{algorithm}

\begin{proposition}  \label{prop:upper}
	From Algorithm~\ref{alg:algorithm_for_upper_bound}, $H(n)$ can be upper bounded by the number of positive simple critical points of $G_k(t,\epsilon)$ for $k \in [K]$.
\end{proposition}
Notice that the numerator for each $k^{\textrm{th}}$ component $G_k(t,\epsilon)$ is a polynomial function with degree $\lfloor N \cdot T = 2 \cdot 2 \pi \rfloor = 12$. Drawing on the experience of~\citet{huang2023:algorithmic}, we can greatly improve the calculation speed by updating Eq.~\eqref{eq:per_step_1} along with forcing $G_1 \equiv G_2 \equiv \dots \equiv G_{K-1} \equiv 0$.

\vspace{0.2cm}
\noindent\textbf{Concrete example of the algorithmic upper bound.} Here, we consider a simple case as follows:
\begin{equation} \label{eq:example_u}
	\left\{~\begin{aligned}
		\frac{\dif u_1(t)}{\dif t} &= -\frac{u_1(t)}{\tau_m} + u_1^2(t) u_2(t) + \epsilon ~\textrm{Poly}_1(\boldsymbol{u}(t);m)  \\
		\frac{\dif u_2(t)}{\dif t} &= -\frac{u_2(t)}{\tau_m} + u_1(t) u_2^2(t) + \epsilon ~\textrm{Poly}_2(\boldsymbol{u}(t);m) \ ,
	\end{aligned}\right.    
\end{equation}
where we have configurations of $N=2$, $n=3$, $ m= 3$, $K=5$, and
\[
\textrm{Poly}_i(\boldsymbol{u};3) = \beta_{i1} u_1 + \beta_{i2} u_2 + \beta_{i3} u_1^2 + \beta_{i4} u_1u_2 + \beta_{i5} u_2^2 + \beta_{i6} u_1^3 + \beta_{i7} u_1^2u_2 + \beta_{i8} u_1u_2^2 + \beta_{i9} u_2^3 \ ,
\]
for $i\in\{1,2\}$. It is known as a cubic system with $m=3^{\textrm{rd}}$ polynomial perturbations. Provided $K=5^{\textrm{th}}$ component, we have the following conclusion.
\begin{corollary}  \label{coroll:example}
	The maximum number of limit cycles of the concerned system~\eqref{eq:example_u} is at most 3, which can be calculated by the $5^{\textrm{th}}$ components. 
\end{corollary} 
Corollary~\ref{coroll:example} shows that using Algorithm~\ref{alg:algorithm_for_upper_bound} with $m=3$ and $K=5$ enables the upper bound of $H(n)$ with $N=2$ to be calculable. Combining with the lower bound from Theorem~\ref{thm:ScSNN_lower}, we can conclude that $1 \leq H(3) \leq 3$, which is a tight bound and demonstrates the effectiveness of Theorem~\ref{thm:ScSNN_lower}. The detailed materials of this simulation example can be accessed in Appendix~\ref{app:example}.

\subsection{Discussions about Adding Self-connection Architectures}
In this section, we aim to enrich the adaptivity, or equally eigenvalue flexibility of integration operation by adding self-connection architectures. It is worth noting that the utilization of self-connection architectures is not a novel concept within the community of ANNs. A recent study points out that adding self-connection contributes to the robustness and rapid convergence of neural network training~\citep{shahir2023connected}. Besides, \citet{zhang2021:bsnn} resort to the algebraic equation and decoupling principles led by Eq.~\eqref{eq:algebraic_ScSNN} and Eq.~\eqref{eq:ScSNN_eigenvalues} for improving the performance of SNNs. \citet{kim2023sharing} attest to the effectiveness of SNNs equipped with neuron-sharing architectures.

Indeed, it is important to acknowledge a practical consideration. The current implementation of the mutual promotion, exemplified by the Taylor expansion of the self-connection function in Eq.~\eqref{eq:ScSNN_mutual} from neuron $k$ to neuron $i$, inevitably leads to a larger memory consumption, especially when the input spike sequences are high-dimensional and high-frequency. Thus, it is prospective to explore some more practical techniques or modules. Besides, it is crucial to recognize that while adding self-connection architecture presents a promising avenue, it represents just one of the approaches for implementing adaptive eigenvalues in the integration operation of SNNs. While it holds considerable potential, it may not be the definitive solution. Therefore, it is imperative to actively consider and explore other valid approaches to ensure a comprehensive understanding and to address the challenges that arise in this context.

Structural stability is not necessarily related to accuracy. In the case of a structurally unstable system, a bifurcation would lead to a total collapse of its full invariant set. Essentially, the hypothesis space of such a structurally unstable system is flawed; thus, a structurally unstable system potentially hampers the accuracy. Hence, it is imperative for applicants to steer clear of employing structurally unstable systems. This concern stands as a focal point in our work. Hence, our aim is to scrutinize the structural stability of SNNs. Our findings, presented in Theorem~\ref{thm:ScSNN_periodic}, demonstrate that the inclusion of self-connection architectures bolsters structural stability. This stands in stark contrast to conventional SNNs, which, as proven in Theorem~\ref{thm:bifurcation_dynamics}, exhibit structural instability as a bifurcation dynamical system. Consequently, our work establishes theoretically that adding self-connection architecture provides a viable means for SNNs to sidestep the functional perturbations they may encounter. Besides, our theoretical results show the effect of self-connection on withstanding perturbations since the higher-order implementation can be regarded as a Taylor approximation to the complex self-connection function. This reframes the challenge of assessing structural stability in SNNs as a mathematical problem with quantifiable properties.

On the other hand, we investigate the structural stability, or equally qualitative behaviors of bifurcation solutions by adding small perturbations (to be exact polynomial perturbations). Our objective is to quantify stable bifurcation solutions as a metric for a stable structure. Notice that added perturbations here are some functions, as opposed to parameter adjustments~\citep{carlson2013:stability,kim2021:stability}. In other words, when one examines the structural stability of a system, it is better to add perturbation functions to the bifurcation solutions rather than making adjustments to the bifurcation parameters or other factors. It is crucial to differentiate between structural stability, Lyapunov stability (which concerns perturbations in initial conditions for a fixed system), and algorithmic stability (which characterizes perturbations in training sets for specific learning algorithms). Therefore, structural stability is an inherent property of a system, independent of the input. Unfortunately, the upper and lower bounds of $H(n)$ remain elusive for tightening the generalization bounds because these bounds are closely tied to the functional complexity of the SNN model itself, but indirectly linked to the learning procedure. 

In conjunction with the typical learning theory, it is imperative for applicants to both minimize empirical errors and avoid structural instability in the training phase. At present, optimizing the latter during the learning process remains a challenging endeavor. Nevertheless, there is potential in devising technologies (e.g., regularizers or normalizers, etc.) related to structural stability, enabling SNNs to escape from a structurally unstable system. We here present a feasible paradigm as follows
\[
\min \sum_{i\in[N_D]} \mathbb{I}( h(\boldsymbol{x}_i) \neq y_i ) + \lambda~ \mathbb{K}(\mathcal{H}) \ ,
\]
where $(\boldsymbol{x}_i, y_i)$ denotes a pair of training instances and $\mathbb{K}$ is a regularizer that indicates the structural stability. In future work, it is attractive to empirically verify the effects of this paradigm.

\section{Spiking Neuron Model with Stochastic Excitation} \label{sec:stoc}
As discussed in Subsection~\ref{subsec:firing-reset}, it is beneficial for SNN learning to make the firing capacity of firing-reset mechanisms adaptive to the data or environment. An intuitive way is to render $u_{\textrm{firing}}$ learnable, replacing the pre-given and fixed value. Formally, one has
\[
\frac{\partial s_k}{\partial u_{\textrm{firing}}} = \frac{\partial f_e(u_k)}{\partial u_{\textrm{firing}}} \ .
\]
However, as per Eq.~\eqref{eq:excitation}, $f_e(u_k(t))$ is non-differentable at the timestamps where the membrane potential exceeds the firing threshold and is reduced to the reset voltage. This makes direct optimization of $u_{\textrm{{firing}}}$ with gradients challenging. Several efforts have been made to explore alternatives for $u_{\textrm{firing}}$. \citet{huh2018:BPTT} replaced $u_{\textrm{firing}}$ with a gate function that induces gradual areas, enabling differential spiking dynamics. In a similar vein, \citet{rathi2020diet} employed another network model to optimize both membrane leak and firing threshold, allowing $u_{\textrm{firing}}$ to be trained using typical back-propagation algorithms.

This section introduces the stochastic spiking neuron model. In contrast to the conventional studies that take deterministic firing-reset computing, the stochastic spiking neuron fires spike by means of a calculable probability $p(u)$ with respect to the membrane potential $u_k(t)$ and firing threshold $u_{\text{firing}}$ at time $t$, so that the firing-reset mechanism including $u_{\text{firing}}$ can be adjusted by $p(u)$.

\subsection{Stochastic Spiking Neuron Model and Neural Network} \label{subsec:EBP}
Formally, we have the stochastic spiking neuron model as follows
\begin{equation} \label{eq:stoc_spiking}
	\left\{~\begin{aligned}
		\text{Integration} \quad&:~ \tau_m \frac{\partial u_k(t)}{\partial t} =- u_k(t) + \tau_r \sum_{j=1}^{M} \mathbf{W}_{kj}\mathbf{I}_j(t) \\
		\text{Stochastic Excitation} &:~ s_k(t) = f_e^{\textrm{stoc}}(u_k(t)) \sim \text{Bernoulli}(p(u))  \\~\\
		\text{Resetting} \quad&:~ u_k(t) = (1 - s_k(t)) \cdot u_k(t) + s_k(t) \cdot u_{\text{reset}} \ ,
	\end{aligned}\right.
\end{equation}
where $k\in[N]$. The excitation probability $p(u)$ endues the firing-reset mechanism with stochasticity. This work provides various formations for $p(u)$ as follows:
\begin{equation}  \label{eq:p}
	\left\{~\begin{aligned}
		\textrm{Linear} \quad&:~ p_{\textrm{linear}}(u(t)) = \frac{u_{\textrm{firing}} - u(t) }{u_{\textrm{firing}} - u_{\textrm{reset}} } \ , \\
		\textrm{Exponential} &:~ p_{\textrm{exp}}(u(t)) = \exp\left( \frac{u(t) - u_{\textrm{firing}}}{\sigma \cdot ( u(t) - u_{\textrm{reset}} )^q } \right) \ , \\
		\textrm{Heaviside-like} &:~ p_{\textrm{H}}(u(t)) = \left\{\begin{aligned}
			\exp\left( \frac{ u(t) - u_{\textrm{firing}}}{\sigma \cdot ( u(t)-u_{\textrm{reset}} )^q } \right) \ ,&\quad \text{if $u_{\theta} \leq u(t) < u_{\textrm{firing}}$} \ , \\
			\quad\quad\quad\quad 0 \quad\quad\quad\quad\quad\quad  ,&\quad \text{if $u_{\textrm{reset}} \leq u(t) < u_{\theta}$} \ , \\
		\end{aligned}\right.
	\end{aligned}\right.
\end{equation}
in which $\sigma$ is a scaling hyper-parameter concerning $u_{\textrm{reset}}$, the superscript $q >0$ determines the curvature of function $p(u)$, and $u_{\theta}$ induces a truncation where no stochasticity if $u(t)$ is smaller than the truncation threshold $u_{\theta}$. We define an excitation probability threshold $p_{\theta} \in (0,1]$, satisfying that
\[
\sigma \ln p_{\theta} = \frac{ u_{\theta} - u_{ \textrm{firing}} }{ ( u_{\theta} - u_{\textrm{reset}} )^q }  \ .
\]
By regulating $p_{\theta}$, we can limit the firing possibility of stochastic spiking neurons. In this work, we recommend using the Heaviside-like function as $p_{\theta} \to 0$ corresponds to $p^{\textrm{exp}}(t)$ and the whole integration-and-firing process degenerates into the conventional discrete-LIF model when $p_{\theta} = 1$. We will display the recommended values and effect of $p_{\theta}$ in Table~\ref{tab:paras_main} and Figure~\ref{fig:dropout_rate} , respectively, by conducting real-world experiments.

Provided Eq.~\eqref{eq:stoc_spiking}, we can establish a fully-connected feed-forward SNN with stochastic excitation. The feed-forward procedure with $L$ spiking layers can be listed as follows:
\begin{equation} \label{eq:StocSNN}
	\left\{~\begin{aligned}
		\boldsymbol{s}^{(0)}(t) &= \mathbf{I}(t)
		\quad\text{and}\quad
		\boldsymbol{u}^{(0)}(0) = \boldsymbol{0} \ , 
		\quad\text{for}\quad t \in [T] \ ,\quad  l \in [L] \ , \\
		\tau_m \frac{\dif \boldsymbol{u}^{(l)}(t)}{\dif t} &=- \boldsymbol{u}^{(l)}(t) + \tau_r f_{\textrm{agg}} (\boldsymbol{s}^{(l-1)}(t)) \ , \\
		\boldsymbol{s}^{(l)}(t) &\sim \text{Bernoulli} \left( \boldsymbol{p}^{(l)} \left( \boldsymbol{u}^{(l)} \right) \right) \quad\text{with}\quad p^{(l)}_k \left( u_k^{(l)} \right) \sim \textrm{Eq.}~\eqref{eq:p}  \quad\text{for}\quad k \in [N_l] \ , \\
		\boldsymbol{u}^{(l)}(t) &= (\boldsymbol{1} - \boldsymbol{s}^{(l)}(t)) \odot \boldsymbol{u}^{(l)}(t) + \boldsymbol{s}^{(l)}(t) \cdot u_{\text{reset}} \ , \\
		\boldsymbol{o}(t) &= \boldsymbol{s}^{(L)}(t) \ , 
	\end{aligned}\right.
\end{equation}
where $N_l$ denotes the number of spiking neurons in the $l$-th layer, $\mathbf{W}^{(l)}$ is the connection weight matrix in the $l$-th layer, $\boldsymbol{o}$ is the final output vector of the concerned model, and $\odot$ indicates the Hadamard product. Eq.~\eqref{eq:StocSNN} provides a standard procedure for forecasting multivariate spike sequences, i.e., inputting spike sequences and then outputting spike sequences. 

It is worth noting that the proposed stochastic neuron model is a fundamental component of SNN learning; thus, can be used with various architectures, including the self-connection architecture discussed in Section~\ref{sec:self-connection}. The feed-forward and back-propagation procedures follow the respective calculation processes of each module. Besides, if one considers handling the neuromorphic datasets that receive spike sequences but yield a comprehensive prediction, we can decode the spike sequence $\boldsymbol{s}^{(L)}$ or $\boldsymbol{o}$ in the last layer as the prediction~\citep{pillow2005:decoder}, denoted by $y = \text{Decoder}(\boldsymbol{o}(1:T))$. Alternatively, we can adopt a parametric approach, for instance, by counting the output spikes with a Poisson distribution, as introduced in Section~\ref{sec:inside}.

\vspace{0.2cm}
\noindent\textbf{Key Ideas of Stochastic Excitation.} The motivation of stochastic excitation is to enhance the adaptivity of firing-reset mechanisms. We attempt to tackle this challenge by probabilizing the firing-reset mechanism so that a spiking neuron fires spikes by means of a calculable probability function. According to Eq.~\eqref{eq:p}, a lower injection ratio or sampling density would lead to a lower excitation probability value. In this case, the concerned spiking neuron still has a certain excitation probability of being activated even though the integrated membrane potential does not reach the firing threshold. Thus, stochastic excitation is an effective way to avoid the issue of ``dead neurons". 

We also show that the stochastic spiking neuron model allows gradient calculations since the expectation derivatives of spike excitation functions are non-asymptotic and unbiased in the next part. Consequently, developers have the flexibility to apply SNNs with stochastic excitation across a range of datasets and even employ various neural encoding techniques without the need for meticulous hyper-parameter fine-tuning. Subsection~\ref{subsec:approximation_stoc} delves into the expressive power of using a stochastic spiking neuron model. Besides, the excitation probability governs the number of neurons participating in each epoch of training, resulting in multiple spiking subnetworks composed of the neurons that survived throughout training. The ensemble of these spiking subnetworks may serve to counter overfitting and improve the generalization performance of SNNs. We investigate the generalization of the spiking neuron model in Subsection~\ref{subsec:generalization_stoc}.

\vspace{0.2cm}
\noindent\textbf{Stochastic Error Back-Propagation.} We start the stochastic back-propagation algorithm for SNN training on the regression tasks. Let $\boldsymbol{o}(t) \in \{0,1\}$ denote the output vector of an $L$-layer SNN with stochastic excitation at time $t$ for $t\in[T]$, and correspondingly $\hat{\boldsymbol{o}}(t)$ is the target vector at time $t$ provided input spike sequence $\mathbf{I} \in \{0,1\}^{M \times T}$ where $I(t) \in \{0,1\}^{M}$. Thus, the temporal-accumulated error in a discrete-time interval $[1:T]$ can be formulated by
\begin{equation}  \label{eq:loss_regression}
	E = \sum_{t=1}^{T} E (t) =  \sum_{t=1}^{T} \mathscr{L} \left( \hat{\boldsymbol{o}}(t), \boldsymbol{o}(t) \right) \ ,
\end{equation}
where $\mathscr{L}$ denotes the loss function, such as the least square loss and 0-1 loss functions. For $t\in[T]$, we have
\begin{equation} \label{eq:bp_regression}
	\frac{\partial E(t)}{\partial \mathbf{W}^{(l)}} = \frac{\partial \mathscr{L} \left( \hat{\boldsymbol{o}}(t), \boldsymbol{o}(t) \right)}{\partial\mathbf{W}^{(l)}} 
	= \sum_{t'=1}^{t} \underbrace{ \frac{\partial \delta(t) }{\partial \boldsymbol{s}^{(l)}(t')}  }_{\textrm{global}}
	\cdot
	\underbrace{ \frac{\partial \boldsymbol{s}^{(l)}(t')}{\partial\boldsymbol{u}^{(l)}(t') } }_{\textrm{post-synaptic}}
	\cdot
	\underbrace{ \frac{\partial \boldsymbol{u}^{(l)}(t')}{\partial\mathbf{W}^{(l)}} }_{\textrm{pre-synaptic}}  \ ,  
\end{equation}
where $\delta(t) =  \mathscr{L} ( \hat{\boldsymbol{o}}(t), \boldsymbol{o}(t) )$. Notice that the derivative in Eq.~\eqref{eq:bp_regression} consists of three terms, i.e., the global, post-synaptic, and pre-synaptic derivatives.
\begin{itemize}
	\item The global derivative contains two back-propagation pipelines, that is, temporal-wise and layer-wise propulsions that correspond to
	\begin{equation}  \label{eq:global}
		\prod_{t''=t'+1}^{t+1} \frac{\partial \boldsymbol{s}^{(l')}(t'')}{\partial \boldsymbol{s}^{(l')}(t''-1)}
		\quad\text{and}\quad
		\prod_{l'=l+1}^{L} \frac{\partial \boldsymbol{s}^{(l'-1)}(t')}{\partial \boldsymbol{s}^{(l')}(t')} \ ,
		\quad\text{respectively.}
	\end{equation}
	\begin{table}[t]
		\centering
		\caption{Post-synaptic computations.}
		\label{tab:gradients}
		\resizebox{1\textwidth}{!}{%
			\begin{tabular}{c | c | c}
				\toprule
				& Feed-forward & Post-synaptic functions \\
				\midrule
				Surrogate Gradients &			
				$s =\textrm{Heaviside}(u - u_{\textrm{reset}})$
				& smooth functions based on the distance between $u$ and $u_{\text{firing}}$	 \\
				\midrule
				SLAYER &  
				$s = \textrm{Heaviside}(u - u_{\textrm{reset}})$ &		
				$ \begin{aligned}
					& \text{probability density function} \\
					& \rho(u) = \alpha\exp( -\beta | u - u_{\textrm{firing}} | ), ~\text{for $\alpha,\beta \in \mathbb{R}$}
				\end{aligned} $ \\		
				\midrule
				Stochastic Gradients (Ours) & $s \sim \text{Bernoulli}(p(u))$ &
				$p_{\textrm{H}}(u) = \left\{\begin{aligned}
					\exp\left( \frac{ u - u_{\textrm{firing}}}{\sigma \cdot ( u-u_{\textrm{reset}} )^q } \right) \ ,&\quad \text{if $u_{\theta} \leq u < u_{\textrm{firing}}$} \ , \\
					\quad\quad\quad\quad 0 \quad\quad\quad\quad\quad\quad  ,&\quad \text{if $u_{\textrm{reset}} \leq u < u_{\theta}$} \ , \\
				\end{aligned}\right.$ \\
				\bottomrule
		\end{tabular} }
	\end{table}
	
	\item The post-synaptic derivative of Eq.~\eqref{eq:bp_regression} indicates the remediation of the discontinuous and non-differential firing phase. In conventional SNNs training algorithms, the derivative of firing function $\textrm{Heaviside}(u(t)-u_{\textrm{reset}})$ is approximated by a smooth surrogate function~\citep{li2021:surrogate}. Here, we leverage the post-synaptic derivative from the perspective of energy back-propagation. It is observed that the pre-synapse receives $f_{\textrm{agg}}(\boldsymbol{s}^{(l-1)}(t))$ at time $t$, and then the post-synapse fires spikes according to the excitation probability $p(u)$. In the whole procedure, the concerned neuron receives the pre-synaptic signals $f_{\textrm{agg}}(\boldsymbol{s}^{(l-1)}(t))$, consumes the integration operations, and then fires the output according to $p(u)$; the former two correspond to the pre-synaptic derivative and the latter results in the post-synaptic derivative. Further, we obtain the energy rate as
	\[
	\frac{\partial p(u) u_{\textrm{firing}}}{\partial u } = \frac{\partial p(u) }{\partial u } u_{\textrm{firing}} \ . 
	\]
	Inspired by this recognition, we can replace the binary spike $\boldsymbol{s}^{(l)}$ by the corresponding excitation probability $\boldsymbol{p}^{(l)}(\boldsymbol{u}^{(l)})$, and thus the post-synaptic derivative of Eq.~\eqref{eq:bp_regression} becomes
	\begin{equation} \label{eq:post}
		\frac{\partial \boldsymbol{s}^{(l)} \left( \boldsymbol{u}^{(l)} \right)}{\partial\boldsymbol{u}^{(l)} } \leftarrow \frac{\partial \boldsymbol{p}^{(l)} \left(\boldsymbol{u}^{(l)} \right)}{\partial\boldsymbol{u}^{(l)} } \ .
	\end{equation}
	
	\begin{figure}[t]
		\centering
		\includegraphics[width=1\textwidth]{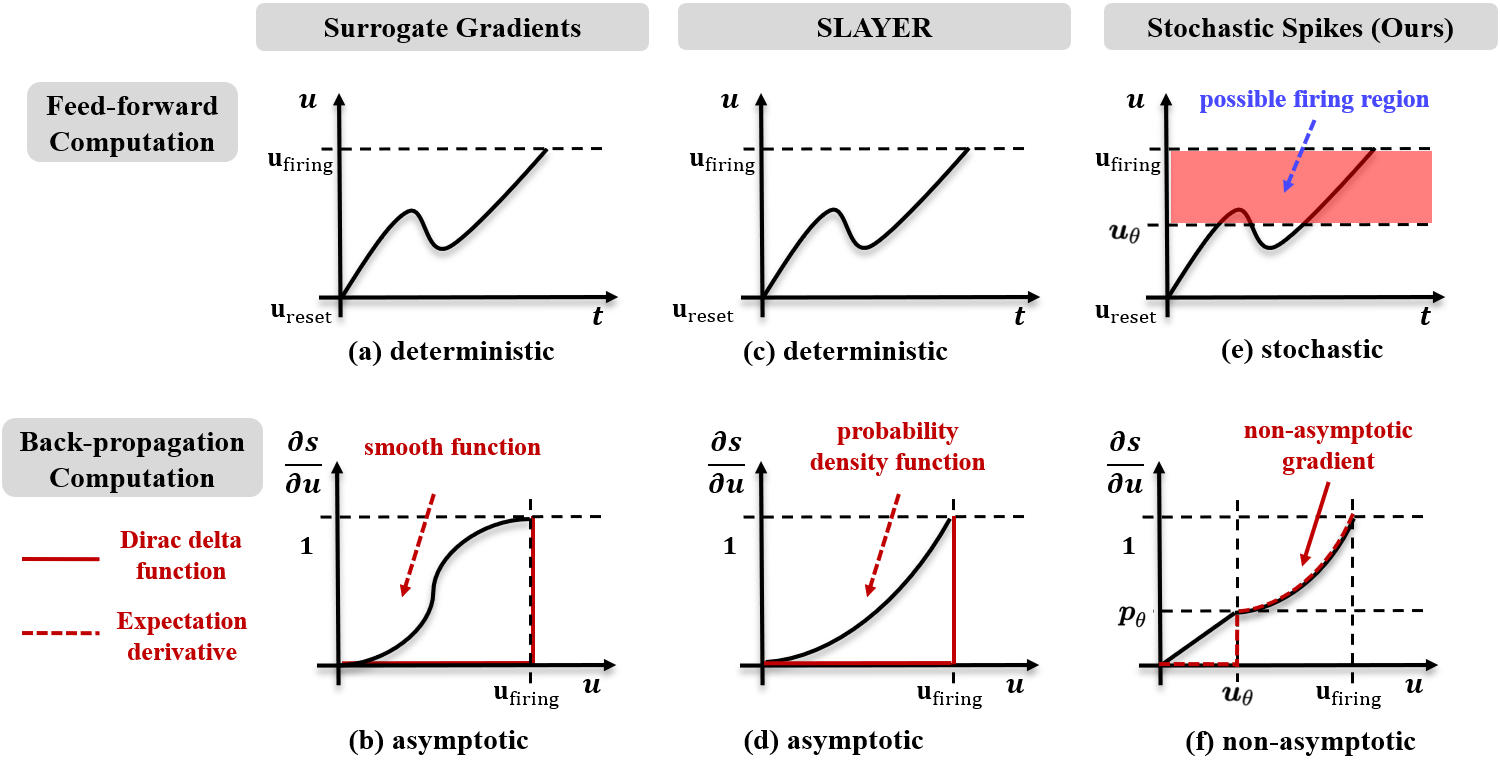}
		\caption{Post-synaptic computations of conventional surrogate gradients, SLAYER, and ours.}
		\label{fig:gradients}
	\end{figure}
	From a computational perspective, we have the following conclusion.
	\begin{theorem}  \label{thm:gradients}
		The post-synaptic derivative of the stochastic spiking neuron model is asymptotic and unbiased as $u_{\theta} = u_{\textrm{reset}}$. For specificity,
		\begin{itemize}
			\item Non-asymptotic. The post-synaptic derivative ${\partial \boldsymbol{p}^{(l)} }/{\partial\boldsymbol{u}^{(l)} }$ is exactly differential as $u_{\theta} = u_{\textrm{reset}}$.
			\item Unbiased. One has
			\[
			\mathbb{E}\left[ \frac{\partial \boldsymbol{s}^{(l)} }{\partial\boldsymbol{u}^{(l)} } \right] = \frac{\partial \boldsymbol{p}^{(l)} }{\partial\boldsymbol{u}^{(l)} } \ .
			\]
		\end{itemize}
	\end{theorem}
	Theorem~\ref{thm:gradients} shows that our proposed stochastic error computation is almost non-asymptotic and unbiased, which provides a theoretical guarantee for the gradient calculations of the stochastic spiking neuron model. We also compare our method with conventional post-synaptic computations. Table~\ref{tab:gradients} lists the post-synaptic function of conventional surrogate gradients~\citep{li2021:surrogate}, SLAYER~\citep{shrestha2018:slayer}, and our proposed stochastic neuron model. Figure~\ref{fig:gradients} illustrates the feed-forward and back-propagation computations. From these charts, conventional surrogate gradients and SLAYER approximate the Dirac delta function (derivatives of non-differentiable firing functions) using the smooth function (asymptotic and biased) and probability density function (asymptotic and unbiased), respectively. However, our proposed stochastic error computation is almost \textbf{non-asymptotic and unbiased}, since the stochastic formulation naturally relaxes the derivatives (red curves) of firing functions from a Dirac delta function to a pseudo-step function (expectation derivative). Limited to space, we move the detailed proof and discussions about Theorem~\ref{thm:gradients} into Appendix~\ref{app:post}.
	
	\item The pre-synaptic derivative of Eq.~\eqref{eq:bp_regression} can be calculated by
	\begin{equation}  \label{eq:pre}
		\frac{\partial \boldsymbol{u}^{(l)}(t)}{\partial\mathbf{W}^{(l)}} = \textrm{REPMAT} \left( \boldsymbol{s}^{(l-1)}(t), N_l \right)^{\top} \ ,
	\end{equation}
	where $\textrm{REPMAT}(\boldsymbol{s},N_l)$ returns a row array containing $N_l$ copies of the column vector $\boldsymbol{s}$. 
\end{itemize} 

The above steps introduce the stochastic back-propagation algorithm for the SNN training on the regression task. Substituting Eqs.~\eqref{eq:global}, ~\eqref{eq:post}, and~\eqref{eq:pre} into Eq.~\eqref{eq:bp_regression} and then summing up Eq.~\eqref{eq:bp_regression} and Eq.~\eqref{eq:loss_regression}, we can obtain the final gradients with respect to $\mathbf{W}^l$.

For training the neuromorphic classification tasks, the error supervised by the target label $y$ can be formulated by 
\[
E_{\textrm{c}} = \mathscr{L} \left( \text{Decoder}(\boldsymbol{o}(1:T)), y \right) \ .
\]
Thus, the gradient with respect to $\mathbf{W}^l$ becomes
\[
\frac{\partial E_{\textrm{c}} }{\partial \mathbf{W}^l} =
\frac{\partial \mathscr{L}\left( \text{Decoder}(\boldsymbol{o}(1:T)), y \right) }{\partial \text{Decoder}(\boldsymbol{o}(1:T))} \left[ \sum_{t=1}^T \frac{\partial \text{Decoder}(\boldsymbol{o}(1:T)) }{\partial \boldsymbol{o}(t)} 
\frac{\partial \boldsymbol{o}(t) }{\partial\mathbf{W}^{(l)}} \right] 
\]
with
\[
\left\{~\begin{aligned}
	&\frac{\partial \mathscr{L}\left( \text{Decoder}(\boldsymbol{o}(1:T)), y \right) }{\partial \boldsymbol{o}(t)} =
	\frac{\partial \mathscr{L} \left( \text{Decoder}(\boldsymbol{o}(1:T)), y \right) }{\partial \text{Decoder}(\boldsymbol{o}(1:T))} 
	~ \frac{\partial \text{Decoder}(\boldsymbol{o}(1:T)) }{\partial \boldsymbol{o}(t)} \ ,  \\
	&\frac{\partial \boldsymbol{o}(t) }{\partial\mathbf{W}^l} = \sum_{t'=1}^{t} \left[ \frac{\partial \boldsymbol{o}(t)}{\partial \boldsymbol{s}^{(l)}(t')}
	\frac{\partial \boldsymbol{s}^{(l)}(t')}{\partial\boldsymbol{u}^{(l)}(t') } 
	\frac{\partial \boldsymbol{u}^{(l)}(t')}{\partial\mathbf{W}^{(l)}} \right] \ . \\
\end{aligned} \right.
\]

For the test time, it is not feasible to randomly excite the spiking neurons since the current weights have been scaled-down versions of the trained weights learned from the stochastic back-propagation algorithm, as well as common-used random algorithms like dropout~\citep{srivastava2014:dropout}, dropconnect~\citep{sakai2019:dropout}, and random ensemble~\citep{zhou2012:ensemble}. Thus, we employ the deterministic model for prediction or classification at the test time.

\subsection{Expressivity of Stochastic Spiking Neurons}  \label{subsec:approximation_stoc}
This subsection investigates the expressive powers of SNNs equipped with stochastic spiking neurons. The fist conclusion is about the universal approximation.
\begin{theorem} \label{thm:ua_stoc}
	Let $K \subset \mathbb{R}^M$ be a bounded set. Provided $l\in\mathbb{N}^+$, if the excitation probability function $p(u)$ is $l$-times differentiable with respective to $u(t)$ on $K$ and satisfies
	\[
	0 < \left| \int_{K} D^r p(u) \dif u \right| < \infty \ ,
	\quad\text{for any $r \in [l]$}   \ ,
	\]
	and $\mathbf{W} \in \mathbb{R}^{N \times M}$, $\boldsymbol{w} \in \mathbb{R}^{N \times 1}$, then the set of functions $f(\boldsymbol{\cdot}, t): K \rightarrow \mathbb{R}$ expressive by a two-layer SNN equipped with stochastic spiking neurons, which is of the form 
	\begin{equation}  \label{eq:express_stoc}
		\left\{~\begin{aligned}
			f(\cdot)  ~&= \mathbb{E}_{\boldsymbol{s}} \left[ \boldsymbol{w}^{\top} \boldsymbol{s}(\cdot, t) \right]  \  ,  \\
			s_k(\cdot,t) &= p (u(\cdot, t), t ) \ , \quad\text{where $k\in [N]$ and $\boldsymbol{s} = (s_1, \dots,s_N)$ } \ , \\
			u(\cdot, t) &= \mathbf{W}_{k,[M]} \int_{t'}^t \exp\left(  - \frac{t'' - t'}{\tau_m} \right) \mathbf{I}(t'') \dif t''  -  \frac{1}{\tau_m} \exp\left(  -\frac{t-t'}{\tau_m} \right) s_k(t')  \ ,
		\end{aligned}\right.
	\end{equation}
	is dense in $\mathcal{C}^0(K,\mathbb{R})$.
\end{theorem}
Theorem~\ref{thm:ua_stoc} shows that a two-layer SNN equipped with stochastic spiking neurons is a universal approximator in the sense of statistical expectation, implying that the proposed stochastic excitation mechanism has a powerful expressive power given apposite or sufficiently large $t$. 

\vspace{0.2cm}
Next, We present the following theorem to show the approximation complexity advantages of the stochastic spiking neuron model over other neuron models.
\begin{theorem}  \label{thm:computational_power}
	For any $\epsilon>0$ and input dimension $d \in \mathbb{N}^+$, there exist a probability measure $\mu$ and a function $g \in \mathcal{C} \left( (\mathbb{R}^+)^M, [0,1] \right)$
	such that the followings hold:
	\begin{itemize}
		\item[(i)] The SNN equipped with only one hidden stochastic spiking neuron $f_{\textrm{stoc}}$ can approximated $g$ well, i.e., $\mathbb{E}_{\mu} \left[ f_{\textrm{stoc}}(\boldsymbol{x}) - g(\boldsymbol{x}) \right]^2 < \epsilon$, where $\boldsymbol{x} \in (\mathbb{R}^+)^M$;
		\item[(ii)] Function $g$ cannot be well approximated by a SNN unless there are at least $\Theta( M^{5/4} )$ hidden LIF neurons;
		\item[(iii)] Function $g$ cannot be well approximated well by a fully-connected feed-forward ANN unless there are at least $(M-6)/2$ hidden sigmoidal neurons.
	\end{itemize}
\end{theorem}
Theorem~\ref{thm:computational_power} shows the parameter complexity advantages of the proposed stochastic spiking neuron over conventional spiking (including LIF) and artificial neuron models for approximating some continuous function on spike timing sequences. In detail, conventional MP and LIF neurons approximate function $g$ well with at least a polynomial number of hidden spiking neurons and a linear number of hidden neurons, respectively, whereas $g$ can be approximated well by the proposed stochastic spiking neurons within a constant number of neurons (only one sometimes). 

The proof sketch is intuitive. The proposed stochastic spike neuron model, according to Eq.~\eqref{eq:p}, can easily approximate the function expressive of the conventional one by regulating $p_{\theta}$. In other words, the whole integration-and-firing process degenerates into the conventional discrete-LIF model when $p_{\theta} \to 0$. Intuitively, the stochastic spiking neuron regulated by the  excitation probability threshold $p_{\theta}$ maintains a stronger approximation ability than the conventional ones.

Notice that $\mathbb{R}^+$ indicates the set of timings of spike timing sequences, by which the original discrete function that works on spike sequences is equivalent to another that maps the timings into an excitation probability vector. We conjecture that there exists an invertible transformation $\phi$ between the spike sequences $\mathbf{X} \in \{0,1\}^{M \times T}$ and its timing sequences $\mathbf{T}_X \in (\mathbb{R}^+)^{M \times T} $. According to~\citet{maass1996noisy}, , we define an element distinctness function $g_{\textrm{EDF}} : (\mathbb{R}^+)^{M \times T} \to \{0, 1\}^M$ by 
\begin{equation} \label{eq:edf}
	g_{\textrm{EDF}} (\mathbf{T}_X) = (g_1, \dots, g_M) \ , \quad
	g_k = \left\{\begin{aligned}
		1~~,  &\quad \text{if $\mathbf{T}_{ki} = \mathbf{T}_{kj}$ for $i \neq j$} \ ; \\
		0~~,  &\quad \text{if $| \mathbf{T}_{ki} - \mathbf{T}_{kj} | \geq c \Delta t$ for $i \neq j$} \ ; \\
		p_{\theta}, &\quad \text{otherwise} \ ,
	\end{aligned}\right. \quad\text{for $k \in [M]$} \ ,
\end{equation}
where $c$ is a scaling constant and $\Delta t$ is a timing threshold. It is obvious that $g_{\textrm{EDF}}$ is an apposite conversion between the rate-based and timing-based encoding, mentioned in Section~\ref{sec:inside}. Thus, the universal approximation issue in Theorem~\ref{thm:ua_stoc} is equivalent to another problem of universally approximating the timing sequences $\mathbf{T}_X \in (\mathbb{R}^+)^{d \times T}$ by regarding $g_{\textrm{EDF}} $ as component of $\phi$. Therefore, $g_{\textrm{EDF}}$ becomes a medium of proving Theorem~\ref{thm:computational_power} when one regards the concerned function $g$ as a component of $g_{\textrm{EDF}}$. The full proof of Theorem~\ref{thm:computational_power} can be accessed from Appendix~\ref{app:proof_approximation}.

\subsection{Generalization Bounds for Stochastic Spiking Neurons}  \label{subsec:generalization_stoc}
This subsection investigates the generalization of SNNs with the stochastic excitation mechanisms. For simplicity, we here focus on binary classification where $\mathcal{Y} = \{-1, +1\}$. As introduced above, SNNs randomly activate spiking neurons according to the possibility indicator $s_j^{(l)}$ that belongs to a Bernoulli distribution with parameter $p_j^{(l)}$. For convenience, we here omit the superscripts and subscripts as possible. Let $\mathcal{W}$ be the connection weight space for SNNs, and $\mathcal{D}$ denotes the underlying joint distribution over input and output space $\mathcal{X} \times \mathcal{Y}$. Thus, we establish the function space as
\[
\mathcal{F}_{\mathcal{W}} = \{ f(\boldsymbol{w}, \mathbf{X} ) \mid \boldsymbol{w} \in \mathcal{W}, \mathbf{X} \in \mathcal{X} \} \ .
\]
Our goal is to find an optimal $\boldsymbol{w}^* \in \mathcal{W}$ so as to minimize the following expected error
\[
E(f) = \mathbb{E}_{(\mathbf{X},y)} \left[ \mathscr{L} \left( f(\boldsymbol{w}, \mathbf{X} ) ,y \right)  \right] \ ,
\]
where $\mathscr{L}$ denotes the loss function, such as the least square loss and 0-1 loss functions. Provided the training data set $S_n = \{(\mathbf{X}_i,y_i) \in \mathcal{X} \times \mathcal{Y} \}_{i\in[n]}$ drawn from $\mathcal{D}$, we define the empirical error as
\[
\widehat{E}(f) = \frac{1}{n} \sum_{i=1}^{n} \mathscr{L} \left( f(\boldsymbol{w}, \mathbf{X}_i) ,y_i \right)  \ ,
\]
where $\widehat{E}(f)$ is an abbreviation of $\widehat{E}(f; S_n, \mathbf{P})$ and the probability matrix
\begin{equation}  \label{eq:PPP}
	\mathbf{P}_{kt} = \left\{\begin{aligned}
		\exp\left( \frac{ u_k(t) - u_{\textrm{firing}}}{\sigma~ (u_k(t)-u_{\text{reset}})^q} \right)  \ ,&\quad \textrm{if $u_{\theta} \leq u(t) < u_{\textrm{firing}}$} \ , \\
		0\quad\quad\quad\quad\quad ,&\quad \text{if $u_{\textrm{reset}} \leq u_k(t) < u_{\theta}$} \ , \\ 
	\end{aligned}\right.
\end{equation}
where $k$ indicates the $k^{\textrm{th}}$ spiking neuron and the membrane threshold $u_{\theta}$ is relative to the excitation probability threshold $p_{\theta} \in (0,1]$. Let $\boldsymbol{p}_k$ is the $k^{\textrm{th}}$ row vector of $\mathbf{P}$.

Here, we study the gap between $E(f)$ and $\widehat{E}(f)$ and present the generalization bound as follows:
\begin{theorem}  \label{thm:generalization_Stoc}
	If the loss function $ \mathscr{L}$ is bounded by $C>0$, then for any $\delta>0$, the following holds with probability at least $1-\delta$
	\begin{subequations}  \label{eq:Rademacher}
		\begin{align}
			E(f) &\leq \widehat{E}(f) + 2 \mathfrak{R}_n( \mathscr{L} \circ \mathcal{F}_{\mathcal{W}}) + C\sqrt{\frac{\ln(2/\delta)}{n}}  \ , \label{eq:Rademacher_a} \\
			E(f) &\leq \widehat{E}(f) + 2 \widehat{\mathfrak{R}}_n( \mathscr{L} \circ \mathcal{F}_{\mathcal{W}}, S_n, \mathbf{P}) + 3C\sqrt{\frac{\ln(2/\delta)}{n}} \label{eq:Rademacher_b} \ , 
		\end{align}
	\end{subequations}
	where $ \mathscr{L} \circ \mathcal{F}_{\mathcal{W}}$ is a composite function space in which $h(\boldsymbol{w}, \mathbf{X}_i, y_i, \boldsymbol{p}_i) =  \mathscr{L} ( f(\boldsymbol{w}, \mathbf{X}_i, \boldsymbol{p}_i) ,y_i)$ for $h \in  \mathscr{L} \circ \mathcal{F}_{\mathcal{W}}$, $\mathfrak{R}_n(\mathscr{L} \circ \mathcal{F}_{\mathcal{W}})$ and $\widehat{\mathfrak{R}}_n( \mathscr{L} \circ \mathcal{F}_{\mathcal{W}}, S_n, \mathbf{P})$ denote the expected and empirical Rademacher complexities of $ \mathscr{L} \circ \mathcal{F}_{\mathcal{W}}$, respectively, provided the Rademacher variable $\epsilon_i$,
	\begin{equation}  \label{eq:general_radermacher}
		\begin{aligned}
			&\mathfrak{R}_n(\mathscr{L} \circ \mathcal{F}_{\mathcal{W}}) = \mathbb{E}_{S_n \in \mathcal{D}, \mathbf{P}} \left[ \widehat{\mathfrak{R}}_n(\mathscr{L} \circ \mathcal{F}_{\mathcal{W}}, S_n, \mathbf{P}) \right]  \ ,  \\
			&\widehat{\mathfrak{R}}_n(\mathscr{L} \circ \mathcal{F}_{\mathcal{W}}, S_n, \mathbf{P}) = \mathbb{E}_{\boldsymbol{\epsilon}} \left[ \sup_{h \in \mathscr{L} \circ \mathcal{F}_{\mathcal{W}}} \left(  \frac{1}{n} \sum_{i=1}^{n} \epsilon_i h(\boldsymbol{w}, \mathbf{X}_i, y_i, \boldsymbol{p}_i) \right) \right] \ .
		\end{aligned}
	\end{equation}
\end{theorem}
Theorem~\ref{thm:generalization_Stoc} shows the generalization bounds concerning the Rademacher complexity. In contrast to the conventional works where the generalization performance is mostly affected by training samples, Theorem~\ref{thm:generalization_Stoc} shows that the generalization bounds are relevant to not only training samples but also the excitation probability in Eq.~\eqref{eq:general_radermacher}. Limited to space, the proof sketch and process of Theorem~\ref{thm:generalization_Stoc} are moved to Appendix~\ref{app:proof_generalization_Stoc}.

Next, we will disclose that the main benefit of introducing stochastic spiking firing into SNNs lies in a sharp reduction on the Rademacher complexity of multi-layer stochastic neurons. Now, we present the second conclusion as follows:
\begin{theorem}  \label{thm:estimation_for_deep_layer}
	Let $\mathcal{F}_{\mathcal{W}}^{\textrm{one}}$ denote the function space of $L$-layer stochastic neurons. If $u_{\textrm{reset}} = 0$, $\| \mathbf{W}^l \|_2 \leq C_l$ for $l \in [L]$, and $\| \mathbf{X} \|_2 \leq C_X$ for $\mathbf{X} \in \mathcal{X}$, we have
	\begin{equation} \label{eq:generalization_bounds}
		\mathfrak{R}_n (\mathscr{L} \circ \mathcal{F}_{\mathcal{W}})
		\leq
		C_n~ \mathfrak{R}_n(\mathcal{F}_{\mathcal{W}}) 
		\leq
		\frac{ (C_n )^L C_X  }{\sqrt{n}} \left( \prod_{l\in[L]}C_l \right) (p_{\textrm{max}})^{(L+1)/2} \ ,
	\end{equation}
	where $p_{\textrm{max}} = \max_{i\in[n],l \in [L]} \{ 1- \max \{ p^{(l,i)}, p_{\theta} \} \} \in (0,1)$ and $C_n$ is a universal constant.
\end{theorem}
Theorem~\ref{thm:estimation_for_deep_layer} shows that the Rademacher complexity of stochastic neurons can be upper bounded by an exponential function relative to the excitation probability, implying that it is promising to reduce the Rademacher complexity $\mathfrak{R}_n (\mathscr{L} \circ \mathcal{F}_{\mathcal{W}})$ and $\mathfrak{R}_n(\mathcal{F}_{\mathcal{W}})$ exponentially by exploiting random algorithms led by stochastic excitation in Eq.~\eqref{eq:stoc_spiking}. The sharp reduction of Rademacher complexity is caused by random algorithms related to the excitation probability function in Eq.~\eqref{eq:stoc_spiking}. Besides, Eq.~\eqref{eq:generalization_bounds} is dependent on the number of training samples, the norm of connection weights, network depth (number of layers), and  excitation probability threshold, but irrelevant to the number of weights, input dimension, and network width (number of units). If one sets $p_{\theta} = 1$, that is, any membrane potential may generate spikes, then our bound can be relaxed to the conventional studies in ANNs~\citep{wan2013:dropout}. Limited to space, the proof sketch and process of Theorem~\ref{thm:estimation_for_deep_layer} are moved to Appendix~\ref{app:proof_estimation_for_deep_layer}.

Notice that the combination of Theorem~\ref{thm:generalization_Stoc} and Theorem~\ref{thm:estimation_for_deep_layer} provides the \textbf{first explicit generalization bound} for SNNs, to the best of our knowledge. Recall the bound in Eq.~\eqref{eq:generalization_bounds}, $p_{max}$ is closely related to $\mathbf{P}_{kt}$ as well as spike dynamics. If one considers the sparsity of $\mathbf{P}_{kt}$, the upper bound may be further reduced. It is an attractive issue to be studied in the future.

\subsection{Discussions about Stochastic Excitation Mechanisms}  \label{subsec:discussions_stoc}
The integration of stochastic elements into the spiking neuron model is an attractive topic confronted in machine learning and neural computation, which stems from two fundamental recognitions: (1) the inherent stochastic nature of the brain where there is certain randomness in the opening or closing of membrane channels due to various neural factors~\citep{markram1996:bio}; (2)  the capacity of stochasticity to empower neuromorphic systems in solving creative problems~\citep{habenschuss2013:noise}. Conventionally, one common approach for introducing stochasticity in SNNs is by incorporating noise into the spiking neuron model~\citep{maass2014:bio}. This allows SNNs to engage in probabilistic inference via sampling~\citep{faisal2008:noise}. The inspiration behind this approach originally arose from the observation that synaptic vesicles are released even without a presynaptic spike~\citep{markram1996:bio}, providing an additional source of inherent noise. From the computational perspective, noisy SNNs work like marginalizing the noise to yield a regularizer. 
Some valuable consequences paved the way for studying the computational ability and learning performance of neuromorphic systems guided by noisy SNNs~\citep{vandesompele2019:noise}. An alternative implementation is to build a generative model by deriving a synaptic update rule that optimizes the likelihood of post-synaptic firing by gradient ascent at firing times~\citep{pfister2006:stoc} or other gradient estimators~\citep{rezende2014:stoc,kajino2021:stoc}. However, both probabilistic inference through sampling and likelihood optimization usually result in considerable storage and computation consumption, which contradicts the high-performance computing capabilities that SNNs are proud of.

This section explores an alternative approach for infusing stochasticity into SNNs, i.e., spike excitation by means of a calculable probability. In contrast to adding noise, the proposed neuron model utilize the stochasticity inside spiking neurons to enrich the flexibility of the firing-reset mechanism and improve the learning procedure of SNN. This idea also comes from the intuition of preventing the complex co-adaptation of feature detectors and encouraging the cortex to be active. There are many benefits.

\vspace{0.2cm}
\noindent\textbf{(1) Coinciding with Biological Facts.} The conventional spiking neuron model establishes upon the post synaptic potential assumption that the post synapse would integrate the membrane potential modified by the neurotransmitters, and the membrane channels would deterministically open only if the integrated potential exceeds a threshold. However, in neuroscience, stochasticity is prevalent in both spike generation and transmission processes, which often corresponds to the unreliability of synapses caused by the inherently stochastic processes on the molecular level, trial-to-trail variability, noise in receiving spike sequences, etc. Studies have intriguingly observed that there is a certain randomness in the opening or closing of membrane channels due to various neural factors~\citep{maass2014:bio}, which means that the neuron may be activated even though the integrated membrane potential has not exceeded the firing threshold. Besides, we also found that a spike of a pre-synapse causes a release of a vesicle filled with neurotransmitters with a relatively low probability around 0.1~\citep{zhang2021:ft}. Thus, it is reasonable to conjecture that stochasticity is inherent in the process of spike transmission and firing within neurons. A clear signal is that there is a mixture of determinism and randomness in the firing mechanism, i.e., the membrane potential will accumulate at the post-synapse and fire after reaching a certain threshold, while there are also cases of stochastic excitation. Inspired by this insight, we resort to the stochastic excitation mechanism by exploiting an excitation probability function.

\vspace{0.2cm}
\noindent\textbf{(2) Enhancing Approximation Ability.} The stochastic excitation function operates on the firing process of spiking neurons over a time period, thus making it possible for the neuron to transmit spikes even if the membrane potential has not yet reached the firing threshold. This manner prompts the cortex to be more excited to learn knowledge and mitigates the occurrence of ``dead neurons"~\citep{bohte2002:SpikeProp} in the training procedure. As shown in Figure~\ref{fig:rasters}, there is a notable $25.64\%$ increase on the number of excitation spikes from stochastic spiking neurons to the discrete-LIF neurons. In addition, by utilizing the excitation probability threshold $p_{\theta}$ (refer to Table~\ref{tab:paras_main} for recommended values), the level of cortical activity can be controlled. Consequently, the proposed SNNs with stochastic excitation can adeptly approximate the expressive functions of the conventional neural network models by regulating $p_{\theta}$. Subsection~\ref{subsec:approximation_stoc} theoretically shows the approximation properties and advantages of SNNs with stochastic excitation over classical SNNs and even ANNs.

\vspace{0.2cm}
\noindent\textbf{(3) Mitigating Overfitting.} The stochastic excitation function temporarily removes the spiking neurons from the network, along with the corresponding incoming and outgoing connections, by means of a calculable possibility. Prospectively, this manner leads to a lot of spiking subnetworks that consist of the neurons that survived the training time. The ensemble of these spiking subnetworks may work against overfitting, thus improving the generalization performance of SNNs. Relatively, during testing, we employ the deterministic model for prediction or classification, with the current weights being scaled-down versions of the trained weights. This workflow of stochastic spiking neurons is reminiscent of the dropout~\citep{srivastava2014:dropout} and random ensemble~\citep{zhou2012:ensemble} techniques in ANNs. Figure~\ref{fig:dropout_rate} will showcase a similar effect on the performance of SNN-Dropout~\citep{sakai2019:dropout} and SNNs equipped with stochastic spiking neurons on N-MNIST. Additionally, the combination of Theorem~\ref{thm:generalization_Stoc} and Theorem~\ref{thm:estimation_for_deep_layer} confirms this conjecture, wherein the sharp reduction of Rademacher complexity is attributed to random algorithms driven by stochastic excitation in Eq.~\eqref{eq:stoc_spiking}. Therefore, the stochastic excitation mechanism provides a computationally cheap and remarkably effective method to reduce overfitting and improve generalization performance in SNNs.

\vspace{0.2 cm}
\noindent\textbf{The Explicit Generalization Bounds.} A comprehensive understanding of SNNs necessitates insights into their expressive power (including approximation ability, computational efficiency, etc.) and generalization performance. Past decade has emerged some studies on the expressivity and computational efficiency of SNNs~\citep{tang2017:sparse,chou2018algorithmic,zhang2022:SNNsTheory}. However, to our knowledge, there is no theoretical investigation to assess the explicit generalization bounds of SNNs, i.e., whether and to what extent a trained SNN performs well on data that has never been seen before. This is a challenging endeavor for two main reasons. Firstly, it is intuitively clear that there are great differences in information processing between ANNs and SNNs, as different computations usually lead to different expressivity and generalization abilities. Thus, there is limited applicability from classical ANN learning experiences. Secondly, existing generalization studies of ANNs have largely been established on representation learning. This primarily revolves around training a neural network model from a pertinent feature space, followed by attempts to bind the complexity of the hypothesis space through the construction of relevant subset representations using specific network architectures and connection weights. The representational progresses work on the norm-based~\citep{neyshabur2015:norm}, kernel-based~\citep{jacot2018:ntk,zhang2022:nngp}, and margin-based~\citep{lyu2022:margin} capacity control. This work advocates the construction of the feature space itself of neural networks rather than the training of a neural network model from a concerned feature space. Following this thought, we probablize the spiking neuron model and formally describe the expressive hypotheses of SNNs in Theorem~\ref{thm:ua_stoc}. This opens the door to establishing a generalization bound for SNNs via the feature space construction driven by the probability excitation function.

\vspace{0.2cm}
\noindent\textbf{The Uncertainty of Stochastic Excitation.} The uncertainty is inherent for SNNs despite the deterministic modeling of spiking neurons. The common-used source of predictive uncertainty of SNNs arises not only from noisy data but also from the randomness and incompleteness of neural encoding techniques discussed in Section~\ref{sec:inside}. While encoding non-spiking data, such as static images, is crucial for training, it can introduce information loss and encoding order-induced randomness, leading to what is commonly referred to as \emph{data uncertainty} or \emph{encoding uncertainty} outside the model. To address this, analysts often resort to employing practical techniques such as adversarial learning, statistical regularization, or rudely extending sequence lengths.

In contrast, our focus here lies on the \emph{model uncertainty}, which investigates the stochastic nature of the model itself. The model uncertainty is led by the stochastic excitation mechanism controlled by the excitation probability $p(u)$ in Eq.~\eqref{eq:stoc_spiking}. Thus, the generalization bound as well as the Rademacher complexity in Eq.~\eqref{eq:generalization_bounds} is relative to excitation probability $p(u)$. Informally, the randomness brought by $p(u)$ leads to hypothesis discrepancy, which coincides with the intuition of preventing overfitting. In Eq.~\eqref{eq:generalization_bounds} of Theorem~\ref{thm:estimation_for_deep_layer}, we roughly measure the upper bound corresponding to the worst case of the hypothesis complexity caused by $p(u)$ via truncating the distribution tail of $p(u)$. We conjecture that this bound may be further tightened by exploiting the spiking computation that is sparse in time and space, which is attractive to be further studied in the future.

\vspace{0.2cm}
\noindent\textbf{Drawbacks.} One drawback of the stochastic spiking neuron model lies in its increased training complexity. Empirically, it typically requires 20-30 more training epochs than the discrete LIF neuron. We conjecture that this circumstance may be attributed to redundant weight updates since the stochasticity in the model can lead to inapposite gradients even with opposite directions at each training epoch. A feasible way to mitigate this issue is to add regularization or normalization techniques. 

Another consideration pertains to the length of neural encoding. While recent years have seen significant progress in few-shot neural encoding for SNNs~\citep{kheradpisheh2018:few}, training an SNN equipped with stochastic spiking neurons still requires a relatively long sequence of spikes, around 300-400 ms, as indicated in Table~\ref{tab:paras_main}. This is due to two primary reasons. Firstly, when one converts a static image into a spike sequence, there exists an inherent gap between the target static image and the resulting spike sequence. A longer encoding length may help alleviate the encoding loss; thus, it is still an argument between the longer and few-shot encoding. Secondly, the stochastic excitation mechanism introduces model uncertainty. A longer spike sequence enables each stochastic spiking neuron to fire a significant number of spikes during training, facilitating ample weight updates while reducing variance attributable to stochasticity.
.

\vspace{0.2cm}
Lastly, it is imperative to emphasize that the stochastic formation of the proposed spiking neuron model shown in Eq.~\eqref{eq:stoc_spiking} is not confined to its current implementation. This encompasses various aspects, including the feed-forward architecture, the error back-propagation through time algorithm, the Heaviside-like possibility function, etc. Exploring alternative and viable schemes remains a valuable avenue for further research.

\section{Experiments}  \label{sec:experiments}
This section conducts experiments on several datasets to evaluate the performance of the proposed methods. For convenience, we employ the post-fixes, that is, `SNN$^*$' and `StocSNN', to denote SNNs equipped with the self-connection architecture and stochastic excitation mechanism, respectively. Besides, the symbol of `StocSNN$^*$' denotes the SNN equipped with both proposed modes. Since the self-connection architectures have relatively larger parameter complexity. Due to this, we here refrain from deploying self-connection architectures with large-scale parameters.

\begin{table*}[!htb]
	\centering
	\caption{Hyper-parameter setting of the proposed SNNs on image recognition.}
	\label{tab:paras_main}
	\resizebox{1\textwidth}{!}{%
		\begin{tabular}{c | c c c c c c c c}
			\toprule
			\textbf{Hyper-parameters Value} & \textbf{MNIST} & \textbf{Fashion-MNIST} & \textbf{EMNIST} & \textbf{CIFAR-10} & \textbf{CIFAR-100} & \textbf{N-MNIST} & \textbf{CIFAR10-DVS}  & \textbf{DVS128-Gesture} \\
			\midrule
			Batch Size                & 32 & 32 & 32 & 128 & 128 & 64 & 64 & 32 \\
			Encoding Length $T$       & 300 & 400 & 400 & 300 & 300 & 300 & 400 & 300 \\
			Expect Spike Count (True)  & 100 & 100 & 140 & 100 & 150 & 80 & 100 & 100 \\
			Expect Spike Count (False) & 10 & 10 & 0 & 10 & 0 & 5 & 10 & 10 \\
			Firing Threshold          & 10 & 10 & 10 & 10 & 10 & 10 & 10 & 10 \\
			Learning Rate $\eta$      & 0.01 & 0.01 & 0.01 & 0.01 & 0.01 & 0.01 & 0.01 & 0.01 \\
			Excitation Probability Threshold $p_{\theta}$ & 0.5 & 0.55 & 0.55 & 0.5 & 0.5 & 0.6 & 0.6 & 0.55 \\
			Maximum Time              & 300 ms & 400 ms & 400 ms & 400ms & 400ms & 300 ms & 400ms & 400ms \\
			Membrane Time $\tau_m$    & 0.2s & 0.2s & 0.2s & 0.2s & 0.2s & 0.1s & 0.1s & 0.1s \\
			Time Constant of Synapse $\tau_s$ & 8 ms & 8 ms & 8 ms & 8 ms & 8 ms & 8 ms & 8 ms & 8 ms \\
			Time Step $\tau_s$        & 1 ms & 1 ms & 1 ms & 1 ms & 1 ms & 1 ms & 1 ms & 1 ms \\
			\bottomrule
	\end{tabular} }
\end{table*}
\subsection{Configurations}
Here, we utilize the architectures of MLP~\citep{shrestha2018:slayer} with 500-500 hidden neurons and VGG-16~\citep{simonyan2014:VGG} with 4096-1024-1024 hidden neurons, denoted as -500-500- and VGG-16 (4096-1024-1024), respectively. For the regression paradigm, we employ the nonlinear mapping $\text{Decoder}(\boldsymbol{o}(1:T)) = \textrm{sigmoid}( \boldsymbol{o}(1:T) \times \boldsymbol{w}_o )$ as the decoder of SNNs, where $\boldsymbol{o}(1:T) \in \mathbb{R}^{n_{\textrm{label}} \cdot T}$, $n_{\textrm{label}}$ denotes the number of labels, and $\boldsymbol{w}_o \in \mathbb{R}^{T \times 1}$ is the weighted vector. For the classification paradigm, the output  (i.e., classification label) of SNNs is the one with the greatest spike count. Notice that we do not add the refractory period to SNNs. The typical configuration values of the proposed SNNs for the conducted datasets are listed in Table~\ref{tab:paras_main}. The contenders comprise the conversion-based SNNs~\citep{kugele2020} and direct-training algorithms~\citep{zhang2019}, including surrogate gradients~\citep{li2021:surrogate} and random algorithms~\citep{sakai2019:dropout}.

The conducted datasets can be divided into two categories: \textbf{(1) The neuromorphic data, used for regression tasks.} (1a) The Neuromorphic-MNIST (N-MNIST) data set\footnote{\url{https://www.garrickorchard.com/datasets/n-mnist}}~\citep{orchard2015} is a spiking version of the original frame-based MNIST data set. Each example in N-MNIST was converted into a spike sequence by mounting the ATIS sensor on a motorized pan-tilt unit and having the sensor move while it views MNIST examples on an LCD monitor. It consists of the same 60,000 training and 10,000 testing samples as the original MNIST data set and is captured at the same visual scale as the original MNIST data set ($28 \times 28$ pixels) with both ``on" and ``off" spikes. (1b) The CIFAR10-DVS data set~\citep{li2017cifar10} is an event-stream conversion of CIFAR-10 by converting 10,000 frame-based images into 10,000 event streams using a dynamic vision sensor (DVS) and a repeated closed-loop smooth (RCLS) movement of frame-based images. Unlike the conversion of frame-based images by moving the camera, the RCLS image movement generates rich local intensity changes in continuous time, which are quantized by each pixel of the DVS camera to generate events. (1c) The DVS128-Gesture data set~\citep{amir2017:gesture} comprises 1,342 instances of a set of 11 hand and arm gestures, which are grouped in 122 trials collected from 29 subjects under 3 different lighting conditions; the gestures include hand waving (both arms), large straight arm rotations (both arms, clockwise and counterclockwise), forearm rolling (forward and backward), air guitar, air drums, and an ``Other'' gesture invented by the subject, where each gesture lasts about 6 seconds; the 3 lighting conditions are combinations of natural light, fluorescent light, and LED light, which was selected to control the effect of shadows and fluorescent light flicker on the DVS128 camera. During each trial, one subject stood against a stationary background and performed all 11 gestures sequentially under the same lighting condition. \textbf{(2) The static images with neural encoding are used for classification tasks.} (2a) The MNIST handwritten digit data set\footnote{\url{http://yann.lecun.com/exdb/mnist/}} comprises a training set of 60,000 examples and a testing set of 10,000 examples in 10 classes, where each example is centered in a $28 \times 28$ image. (2b) The Fashion-MNIST data set\footnote{\url{https://www.kaggle.com/zalando-research/fashionmnist}} consists of a training set of 60,000 examples and a testing set of 10,000 examples. Each example is a $28\times 28$ grayscale image associated with a label from 10 classes. (2c) The Extended MNIST-Balanced (EMNIST)~\citep{cohen2017} data set is an extension of MNIST to handwritten, which contains handwritten upper \& lower case letters of the English alphabet in addition to the digits, and comprises 112,800 training and 18,800 testing samples for 47 classes. (2d) The CIFAR-10 data set~\citep{krizhevsky2009:CIFAR} consists of 60000 $32 \times 32$ color images in 10 classes, with 50000 training images and 10000 test images. (2e) The CIFAR-100 data set is just like the CIFAR-10, except it has 100 classes that are grouped into 20 super-classes, and each class contains 600 (500 training and 100 testings) images. Each image comes with a ``fine'' label (the class to which it belongs) and a ``coarse'' label (the superclass to which it belongs). Similar to~\citep{susemihl2013:encoding,zhang2021:bsnn}, each static image is transformed as a spike sequence using Poisson Encoding. For example, we produce a list of spike signals with a formation of $784 \times T$ binary matrices corresponding to the static MNIST image, where $T$ denotes the encoding length, and each row represents a spike sequence at each pixel.

\begin{table}[!htb]
	\centering
	\caption{The comparative performance of the conducted SNNs on neuromorphic data.}
	\label{tab:Stoc_neuromorphic}
	\resizebox{0.95\textwidth}{!}{%
		\begin{tabular}{@{}cccc@{}}
			\toprule
			\textbf{Datasets} & \textbf{Models} & \textbf{Architectures} (\%)
			& \textbf{Accuracy} \\
			\midrule
			\multirow{10}{*}{N-MNIST}
			& SNN-BP & $2\times28\times$28-800-10 
			& 98.78  \\
			& SKIM~\citep{cohen2017} & $2\times28\times$28-10000-10 
			& 92.87  \\
			& HM2-BP & $2\times28\times$28-800-10 
			& 98.84 $\pm$ 0.02  \\
			& SLAYER & $2\times28\times$28-500-500-10 
			& 98.89 $\pm$ 0.06  \\
			& SNN-Dropout~\citep{sakai2019:dropout} & MLP with 40\% drop rate & 98.17  \\
			& CIFAR-Net & Small CIFAR-Net without NeuNorm
			& 99.44 \\
			& CIFAR-Net & Small CIFAR-Net with NeuNorm
			& 99.53 \\
			& SNN$^*$ & $2\times28\times$28-500-500-10, $n=1$
			& 99.17 $\pm$ 0.12  \\
			& StocSNN  & $2\times28\times$28-500-500-10 
			& \textbf{99.57 $\pm$ 0.10}  \\
			& StocSNN$^*$  & $2\times28\times$28-500-500-10 , $n=1$
			& 99.33 $\pm$ 0.09`  \\
			\midrule
			\multirow{10}{*}{CIFAR10-DVS}
			& Gabor-SNN~\citep{sironi2018hats} & -- & 24.50  \\
			& HAT~\citep{sironi2018hats} & -- & 52.40 \\
			& LIAF~\citep{wu2021liaf} & -- & 71.70  \\
			& CIFAR-Net & Small CIFAR-Net without NeuNorm
			& 58.10 \\
			& CIFAR-Net & Small CIFAR-Net with NeuNorm
			& 60.50 \\
			& SNN-Dropout & CNN with 40\% drop rate & 64.33 \\
			& STBP-tdBN & ResNet-19 & 67.80 \\
			& Dspike~\citep{li2021:surrogate} & ResNet-18 & 75.40 $\pm$ 0.05 \\
			& conversion-based SNN~\citep{kugele2020} & -- & 66.75 $\pm$ 0.22 \\
			& StocSNN  & VGG-16 (4096-1024-1024)
			& \textbf{79.27 $\pm$ 0.17}  \\
			\midrule
			\multirow{4}{*}{DVS128-Gesture}
			& SNN on TrueNorth~\citep{amir2017:gesture} & -- & 94.59 \\
			& STBP-tdBN & ResNet-19 & 96.87 \\
			& conversion-based SNN & -- & 95.68 $\pm$ 0.32 \\
			& StocSNN  & VGG-16 (4096-1024-1024)
			& \textbf{97.12 $\pm$ 0.38}  \\
			\bottomrule
		\end{tabular}
	}
\end{table}

\begin{figure}[t]
	\centering
	\includegraphics[width=1\textwidth]{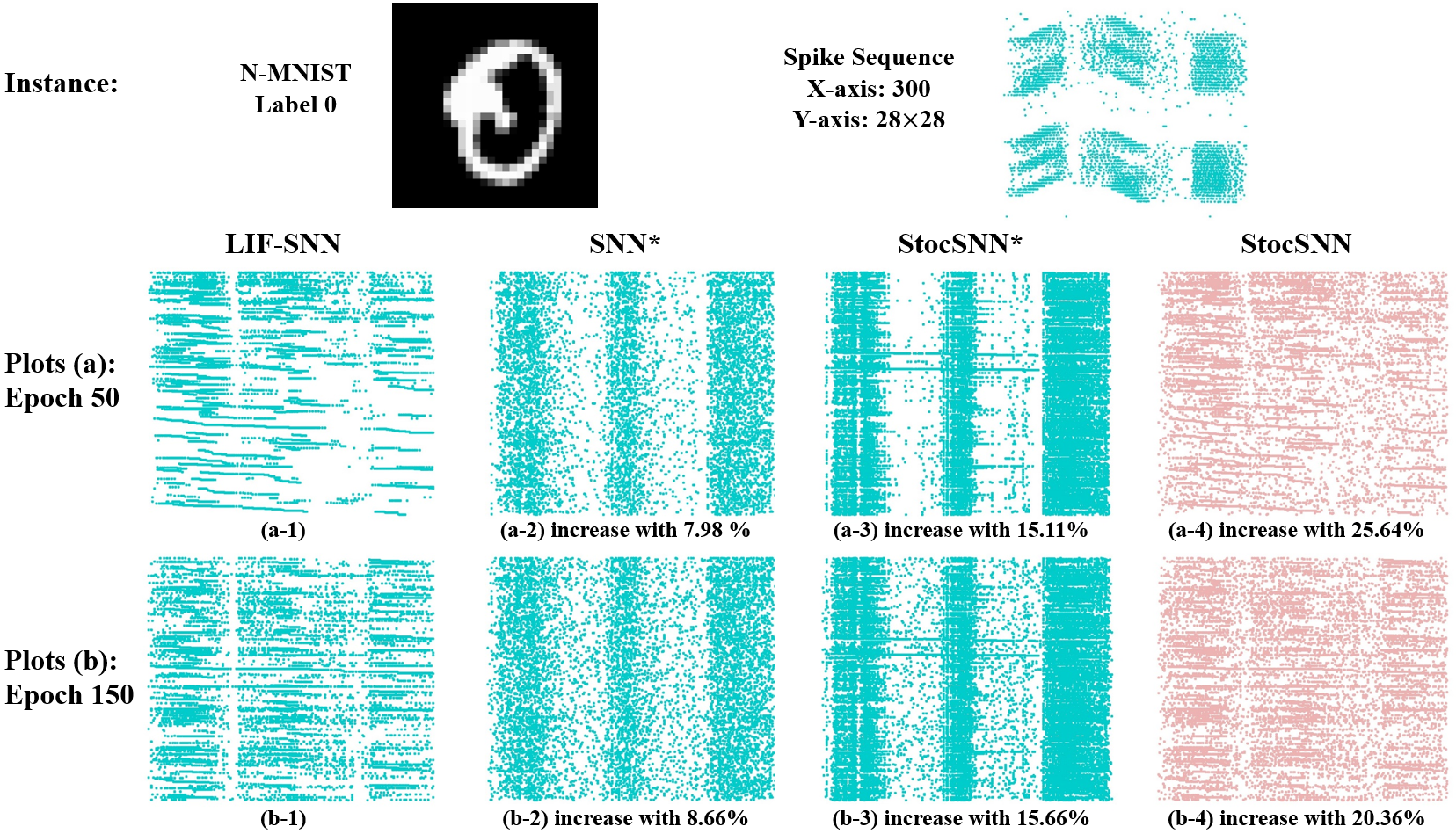}
	\caption{Spike raster illustrations of the general LIF-SNN, SNN$^*$, StocSNN$^*$, and StocSNN for handling N-MNIST.}
	\label{fig:rasters}
\end{figure}
\begin{figure}[t]
	\centering
	\includegraphics[width=1\textwidth]{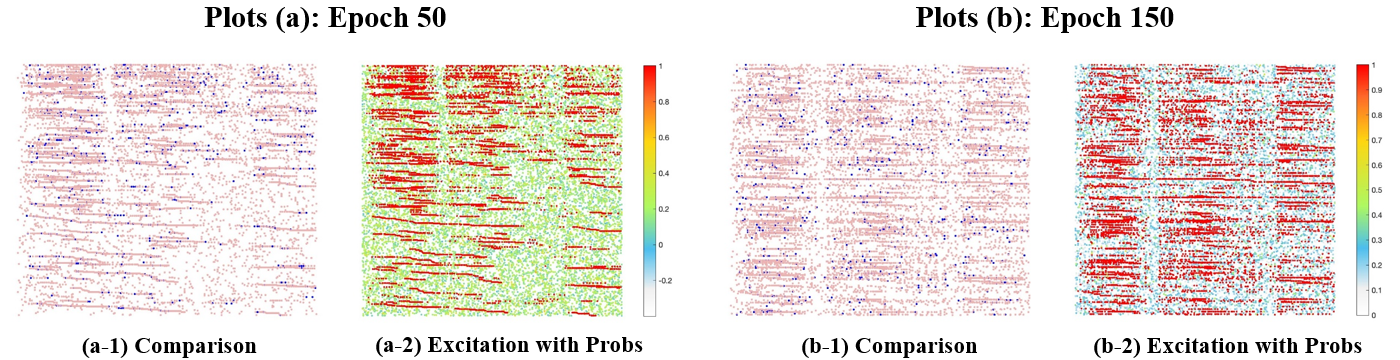}
	\caption{Excitation comparisons between the general LIF-SNN and StocSNN.}
	\label{fig:probs}
\end{figure}
\subsection{Experimental Results on Neuromorphic Data}
Table~\ref{tab:Stoc_neuromorphic} lists a comprehensive comparison of the performance and configurations of the investigated SNNs on neuromorphic datasets. The top-performing models and their performance are highlighted in bold. It is observed that SNNs featuring self-connection architectures exhibit superior accuracy compared to their fully-connected feed-forward counterparts. Additionally, SNNs equipped with stochastic spiking neurons outperform those without in terms of accuracy. The proposed StocSNN stands out as the most effective among the competing approaches, achieving the highest testing accuracy. It is a laudable result for SNNs.

\begin{table}[!htb]
	\centering
	\caption{The comparative performance of the conducted SNNs on static image recognition.}
	\label{tab:Stoc_neuromorphic_static}
	\resizebox{0.98\textwidth}{!}{%
		\begin{tabular}{@{}cccc@{}}
			\toprule
			\textbf{Datasets} & \textbf{Models} & \textbf{Architectures} (\%)
			& \textbf{Accuracy} \\
			\midrule
			\multirow{10}{*}{MNIST} 
			& SNN-BP~\citep{lee2016} & 28$\times$28-800-10 
			& 98.71 \\
			& SNN-EP~\citep{o2019} & 28$\times$28-500-10 
			& 97.63 \\
			& HM2-BP~\citep{jin2018} & 28$\times$28-800-10 
			& 98.84 $\pm$ 0.02 \\
			& SLAYER~\citep{shrestha2018:slayer} & 28$\times$28-500-500-10 
			& 98.53 $\pm$ 0.04  \\
			& SNN-Dropout~\citep{sakai2019:dropout} & MLP with 50\% drop rate & 98.23  \\
			& SNN-DropConnect~\citep{sakai2019:dropout} & MLP with 50\% drop rate & 98.20 \\
			& SNN-L~\citep{lotfi2020} & 28$\times$28-1000-R28-10 
			& 98.23 $\pm$ 0.07  \\ 
			& SNN$^*$  & 28$\times$28-500-500-10, $n=1$
			& 99.02 $\pm$ 0.04 \\ 
			& StocSNN & 28$\times$28-500-500-10 
			& \textbf{99.11 $\pm$ 0.07}  \\
			& StocSNN$^*$  & 28$\times$28-500-500-10, $n=1$
			& 98.91 $\pm$ 0.03 \\
			\midrule
			\multirow{6}{*}{Fashion-MNIST}
			& HM2-BP & 28$\times$28-400-400-10 
			& 88.99 $\pm$ 0.02  \\
			& SLAYER & 28$\times$28-500-500-10 
			& 88.63 $\pm$ 0.12 \\
			& ST-RSBP~\citep{zhang2019} & 28$\times$28-400-R400-10 
			& 90.10 $\pm$ 0.06  \\
			& StocSNN & 28$\times$28-500-500-10, $n=1$
			& 91.22 $\pm$ 0.06  \\
			& StocSNN & 28$\times$28-500-500-10 
			& \textbf{91.37 $\pm$ 0.13}  \\
			& StocSNN$^*$  & 28$\times$28-500-500-10, $n=1$
			& 91.31 $\pm$ 0.07  \\
			\midrule
			\multirow{7}{*}{EMNIST}
			& eRBP~\citep{neftci2017} & 28$\times$28-200-200-47 
			& 78.17  \\
			& HM2-BP & 28$\times$28-400-400-47 
			& 84.43 $\pm$ 0.09  \\
			& SLAYER & 28$\times$28-500-500-47 
			& 85.73 $\pm$ 0.16 \\
			& SNN-L & 28$\times$28-1000-R28-47 
			& 83.75 $\pm$ 0.15  \\
			& SNN$^*$  & 28$\times$28-500-500-47, $n=1$
			& 87.51 $\pm$ 0.23  \\
			& StocSNN & 28$\times$28-500-500-47 
			& \textbf{88.17 $\pm$ 0.18}  \\
			& StocSNN$^*$  & 28$\times$28-500-500-47, $n=1$
			& 87.54 $\pm$ 0.20  \\
			\midrule
			\multirow{10}{*}{CIFAR-10}
			& Converted SNN~\citep{hunsberger2016} & VGG-16 (4096-1024-1024)
			& 87.46 \\
			& TSSL~\citep{zhang2020} & VGG-16 (4096-1024-1024)
			& 91.41 \\
			& DIET-SNN~\citep{rathi2020diet}  & VGG-16 (4096-1024-1024) &  92.43   \\
			& CIFAR-Net~\citep{wu2019direct} & Small CIFAR-Net without NeuNorm
			& 89.83 \\
			& CIFAR-Net & Small CIFAR-Net with NeuNorm
			& 93.16 \\
			& SNN-Dropout & CNN with 40\% drop rate & 96.33 \\
			& SNN-DropConnect & CNN with 40\% drop rate & 97.23 \\
			& STBP-tdBN & ResNet-19 & 72.22 $\pm$ 0.03 \\
			& Dspike~\citep{li2021:surrogate} & ResNet-18 & 93.66 $\pm$ 0.05 \\
			& StocSNN & VGG-16 (4096-1024-1024)
			& \textbf{93.74 $\pm$ 0.07} \\
			\midrule
			\multirow{7}{*}{CIFAR-100}
			& DIET-SNN  & ResNet-20 &  64.07   \\
			& DIET-SNN  & VGG-16 (4096-1024-1024) &  69.67   \\
			& Hybrid-SNN~\citep{rathi2019enabling}   & -- & 67.87  \\
			& RMP-SNN~\citep{han2020rmp}  &  --  &  70.09  \\
			& STBP-tdBN & ResNet-19 & 72.22 $\pm$ 0.03 \\
			& Dspike & ResNet-18 & 73.35 $\pm$ 0.14 \\
			& StocSNN  & VGG-16 (4096-1024-1024)
			& \textbf{74.11 $\pm$ 0.54} \\
			\bottomrule
		\end{tabular}
	}
\end{table}
Figure~\ref{fig:rasters} displays the spike raster plots of the general LIF-SNN, SNN$^*$, StocSNN$^*$, and StocSNN on an N-MNIST instance with label 0 at the $50^{\textrm{th}}$ and $150^{\textrm{th}}$ epochs. Due to space limitations, we present plots for the first hidden layer (consisting of 500 hidden spiking neurons) along with the corresponding excitation probability images. The x-axis and y-axis denote the time stamps and dimensions (input channels and neurons), respectively. As indicated in Table~\ref{tab:paras_main}, we here set $p_{\theta} = 0.6$ for handling N-MNNIST. Limited to the space, we here only show the first-hidden-layer (500 hidden spiking neurons) plots and the corresponding excitation probability pictures. 

In the $50^{\textrm{th}}$ epoch, StocSNN generates 8307 spikes, representing around a 25.64\% increase compared to the 6612 spikes fired by the LIF-SNN. Adding self-connection architectures promotes more spikes than the LIF-SNN, where SNN$^*$ and StocSNN$^*$ produce 7140 and 7611 spikes, respectively, reflecting increases of 7.98\% and 15.11\% over the LIF-SNN. In the case of the $150^{\textrm{th}}$ epoch, StocSNN generates 9770 spikes, surpassing the 8117 spikes of the LIF-SNN by approximately 20.36\%. SNN$^*$ and StocSNN$^*$ produce 8820 and 9388 spikes, respectively, indicating increases of 8.66\% and 15.66\% over the LIF-SNN. 

Furthermore, we compare the generated spikes and corresponding excitation probability pictures of the general LIF-SNN and StocSNN in Figure~\ref{fig:probs}. Red points record the spikes generated by StocSNN, while blue points denote the spikes unique to the LIF-SNN, distinct from those in StocSNN. It is evident that StocSNN generates more spikes than the discrete LIF-SNN with the same connection weights (8307 versus 6612, approximately a 25.64\% increase).

Additionally, it is observed that the excitation probability values are roughly divided into two intervals: deterministic spikes (marked in red) and stochastic spikes with excitation probabilities concentrated around 20\% to 2\% (marked in green). This coincides with the aforementioned increase rate (approximately 25.64\%). The fluctuation of excitation probability values over training epochs is also notable, with the ratio of deterministic spikes to stochastic spikes, as well as the stochastic probability values, diminishing as the training epoch progresses. This aligns with the original intention behind the stochastic formation, as discussed in detail in Subsection~\ref{subsec:discussions_stoc}.

Finally, it is essential to make a clarification that while our proposed methods plausibly lead to an increased firing count, it has not been confirmed whether more spikes necessarily result in higher accuracy.

\subsection{Experimental Results on Static Images}
Table~\ref{tab:Stoc_neuromorphic_static} presents a comparative analysis of performance (accuracy) and configurations (architectures) for the investigated SNNs on static images in various classification tasks. The most outstanding performance is highlighted in bold. Combined with the results in Table~\ref{tab:Stoc_neuromorphic}, it is evident that the proposed StocSNN consistently outperforms other competing approaches, excelling in both static image and neuromorphic dataset tasks. This achievement signifies a noteworthy milestone for SNNs.

\begin{table}[!htb]
	\centering
	\caption{The ``mean-variance" of trained connection weights on N-MNIST.}
	\label{tab:beyond_acc}
	\resizebox{0.9\textwidth}{!}{%
		\begin{tabular}{@{}ccc@{}}
			\toprule
			\textbf{Models} & \textbf{(mean, var) of the $1^{\textrm{rt}}$ hidden layer} & \textbf{(mean, var) of the $2^{\textrm{nd}}$ hidden layer} \\
			\midrule
			LIF-SNN & (-0.0038, 0.0562) & (0.0073, 0.0562) \\
			CNN + LIF-SNN & (-0.0979,1.7063) &  (-0.1806,1.6888) \\
			SNN$^*$ & (-0.0036, 0.0059)  & (0.0077, 0.0580) \\
			StocSNN & (-0.0037, 0.0060)  & (0.0079, 0.0566) \\
			StocSNN$^*$ & (-0.0036, 0.0059)  & (0.0078, 0.0578) \\
			\bottomrule
		\end{tabular}
	}
\end{table}
\subsection{Beyond Accuracy} \label{subsec:beyond_accuracy}
From Figure~\ref{fig:rasters}, it is observed that StocSNN exhibits a higher spike frequency compared to the general LIF-SNN. This observation leads us to conjecture that the modification of intrinsic structures encourages an incremental firing rate. Relatively, the firing rates of SNN$^*$ and StocSNN$^*$ in the same layer are significantly different. This suggests that the adaptive eigenvalues of integration operations play a crucial role, where negative eigenvalues hinder spike excitation, positive ones promote it, and eigenvalues of zero denote a conservative system.

To verify these conjectures, we compute the mean-variance of connection weights for four trained SNNs: the general LIF-SNN, CNN+LIF-SNN, SNN$^*$, StocSNN, and StocSNN$^*$. The results are listed in Table~\ref{tab:beyond_acc}, where it is believed that similar mean and variance imply the similar distributions of trained connection weights. We make two significant observations based on the table.
\begin{itemize}
	\item The means of both two-layer connection weights of LIF-SNN are smaller than those of CNN+LIF-SNN, which employs the same neural model and different architecture with LIF-SNN. However, the variances of both two-layer connection weights in LIF-SNN are considerably larger than those in CNN+LIF-SNN. This observation demonstrates the substantial influence of network architectures on connection weights, where the distributions (i.e., mean and variance) of trained connection weights are relatively different even using the same neural operations. 
	\item The four investigated models (i.e., LIF-SNN, SNN$^*$, StocSNN, and StocSNN$^*$) showcase similar distributions of trained connection weights. This suggests that the effects of the trained connection weights of the four SNN models are almost equivalent. Hence, we can conclude that the improved performance is primarily attributed to the modification of intrinsic structures, namely, the addition of the self-connection architecture and the utilization of the stochastic excitation mechanism.
\end{itemize}

\begin{figure}[t]
	\centering
	\includegraphics[width=0.75\textwidth]{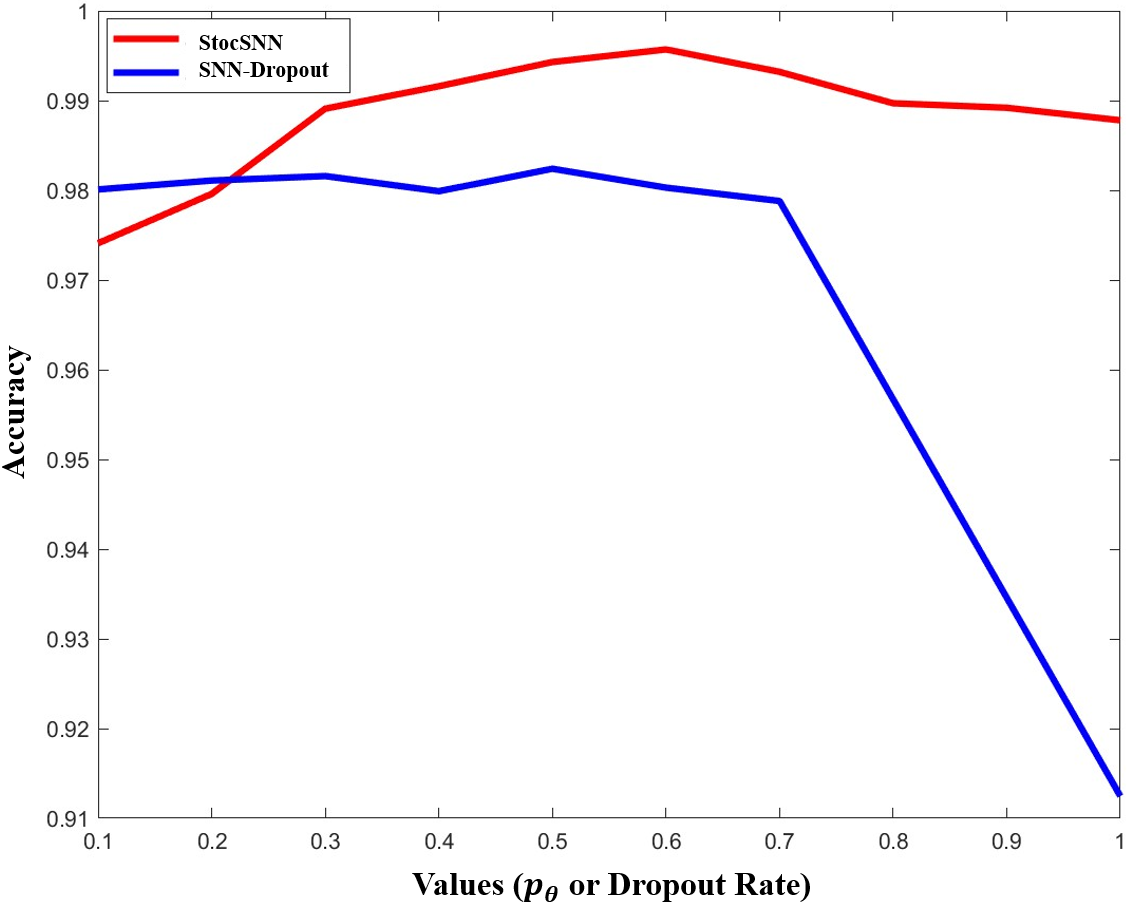}
	\caption{The effect of dropout rate and $p_{\theta}$ on the performance of SNN-Dropout~\citep{sakai2019:dropout} and StocSNN on N-MNIST.}
	\label{fig:dropout_rate}
\end{figure}
\subsection{About the  excitation probability threshold $p_{\theta}$}
The stochastic spiking neuron model works with an extra hyper-parameter, i.e., the  excitation probability threshold $p_{\theta} \in [0,1]$, which regulates the intensity of the firing possibilities of spiking neurons where $p_{\theta} = 0$ implies no stochasticity in SNNs and $p_{\theta} = 1$ represents the highest firing possibility even though the membrane potential has not yet exceeded the pre-defined firing threshold. As mentioned above, the calculable probability function $p(u)$ enables spiking neurons to work with stochastic connections. It would prompt spike excitation since spiking neurons are likely to fire before the membrane potential reaches the firing threshold. Using $p_{\theta}$ in $p(u)$ would ensure that low membrane potential does not cause firing to avoid a massive gap between cumulative membrane potential and firing spikes. Thus, the gap caused by the non-differential post-synaptic computations can be bridged by random algorithms with an adjustable hyper-parameter $p_{\theta}$.

On the one hand, a higher value of $p_{\theta}$ means more activities and more activities enable more effective weight updates, as discussed above. On the other hand, a lower value of $p_{\theta}$ implies that there are more subnetworks in the temporal and layer dimensions. Thus, the setting of $p_{\theta}$ is a trade-off that relies on empirical expertise about the specific tasks. It is an observation that the choice of $p_{\theta}$ is coupled with the number of spiking neurons; smaller $p_{\theta}$ requires bigger networks (with a larger number of spiking neurons), which slow down the training efficiency, and larger $p_{\theta}$ may be beneficial to prevent overfitting. Table~\ref{tab:paras_main} lists the recommended values of $p_{\theta}$ in real-world experiments. Figure~\ref{fig:dropout_rate} plots the effect of dropout rate and $p_{\theta}$ on the performance of SNN-Dropout~\citep{sakai2019:dropout} and StocSNN on N-MNIST, where the x-axis and y-axis denote the values and accuracy, respectively. The accuracy sequence was generated by SNN-Dropout and StocSNN with a dropout rate or $p_{\theta}$ ranging from 0.1 to 1 in increments of 0.1. It is observed that the trends in red and blue curves are consistent, with model shaving dropout rates lower than 0.7 producing flat results, followed by a decline in performance.

\section{Conclusions}  \label{sec:conclusions}
This work provided a theoretical framework for investigating the intrinsic structures of SNNs. By deconstructing the expressivity of SNNs, we unveil two pivotal components of intrinsic structures: the integration operation and firing-reset mechanism. We further conclude that the membrane time hyper-parameter intricately relates to the eigenvalues of the integration operation, thereby dictating the functional topology of spiking dynamics. Additionally, the firing-reset mechanism fundamentally governs the firing capacities of a whole SNN. These insights provide systematic understanding of the impact of intrinsic structures and lead to a crucial recommendation: enhancing the adaptivity of intrinsic structures significantly contributes to improving the performance and universality of SNNs. Inspired by this recognition, we further proposed two feasible methods for improving SNN learning, that is, adding self-connection architectures and building stochastic spiking neuron models by modifying the integration operation and firing-reset mechanism, respectively. We theoretically prove that (1) both two methods promote the expressive property of universal approximation, (2) adding self-connection architectures encourages enough solutions and bolsters structural stability for SNN approximating adaptive Hamiltonian systems, and (3) the stochastic spiking neuron model facilitates bounding the generalization with an exponential reduction in Rademacher complexity, compared to conventional ANNs and SNNs. Empirical validation on various real-world datasets attests to the effectiveness of our proposed methods.

\section*{Acknowledge}
This research was supported by the National Science Foundation of China (62176117,61921006), the Natural Science Foundation of Jiangsu Province (BK20230782), and the research project (“Energy-based probing for Spiking Neural Networks”, Contract No. TII/ARRC/2073/2021) in collaboration between Technology Innovation Institute (TII, Abu Dhabi) and Mohamed bin Zayed University of Artificial Intelligence (MBZUAI, Abu Dhabi).

\clearpage
\appendix
\noindent\textbf{\Large Appendix}

\vspace{0.2 cm}
\noindent This appendix provides the supplementary materials for this work, constructed according to the corresponding sections therein.

\section{Full Proof for Theorem~\ref{thm:bifurcation_dynamics}} \label{app:bifurcation_dynamics}
According to~\citep{zhang2021:bsnn}, the algebraic formulation of a system of LIF equations can be formulated as
\[
\frac{\partial \boldsymbol{u}(t)}{\partial t} = - \frac{1}{\tau_m} \boldsymbol{u}(t) \ ,
\quad\text{when}\quad
\tau_m \neq 0 \ .
\]
Furthermore, we can conclude that $-1/\tau_m$ is the eigenvalues of the LIF-integration operation. Recall the total energy $\mathcal{H}$ and its derivative in Eq.~\eqref{eq:energy_network}, we have
\[
\left\{~\begin{aligned}
	\mathcal{H}(t) &= |\boldsymbol{u}|^2 + \frac{2\tau_r}{\tau_m} \int \left\langle \frac{\partial \boldsymbol{v}}{\partial t},  \boldsymbol{u}(t) \right\rangle  \dif t - \theta \ ,  \\
	\frac{\dif \mathcal{H}}{\dif t} &= \frac{1}{2} \boldsymbol{u}^{\top} ~\mathbf{M}(\tau_m)~ \boldsymbol{u} \ ,
\end{aligned}\right.
\]
where matrix $\mathbf{M}(\tau_m)$ is of the quadratic form
\[
\mathbf{M}(\tau_m) = \begin{pmatrix}
	-1/\tau_m      & 0    & \dots   & 0 \\
	0 &    -1/\tau_m      & \dots   & 0 \\
	\vdots    & \vdots       & \ddots   & \vdots       \\
	0 & 0   & \dots    &  -1/\tau_m 
\end{pmatrix}_{N \times N} \ .   
\]
This derivative ${\dif \mathcal{H}} / {\dif t}$ represents the rate at which the energy function changes, which is determined by the hyper-parameter $\tau_m$. Thus, Eq.~\eqref{eq:SNN_LIF} typically induces a bifurcation dynamical system in which $1/\tau_m$ or $-1/\tau_m$ is the corresponding bifurcation hyper-parameter. Furthermore, it is obvious that
\begin{itemize}
	\item For the case of $-1/\tau_m >0$, one has a energy-increasing system according to ${\dif \mathcal{H}} / {\dif t} > 0$.
	\item For the case of $-1/\tau_m =0$, the system works with invariant energy, that is, ${\dif \mathcal{H}} / {\dif t} = 0$.
	\item For the case of $-1/\tau_m <0$, we have ${\dif \mathcal{H}} / {\dif t} < 0$, thus leading to a energy-decreasing system.
\end{itemize}
As concluded, $-1/\tau_m$ indicates the eigenvalue of the LIF-integration operation, where $1/\tau_m>0$, $1/\tau_m=0$, and $1/\tau_m<0$ correspond to the dissipative, conservative, and energy-diffuse dynamical systems, respectively. This completes the proof. $\hfill\square$

\section{Full Proof for Theorem~\ref{thm:ua_ScSNN}} \label{app:ua_ScSNN}
Similar to the thought line of~\citep[Theorem 1]{zhang2022:SNNsTheory}, we here should prove that the function $f_k(\boldsymbol{\cdot}, t)$ expressive by the $k^{\textrm{th}}$ spiking neuron is a ``well-defined" basis function.  We start this proof with an abbreviation from $f_k(\boldsymbol{\cdot}, t)$ to $f(\boldsymbol{\cdot})$ for simplicity. For $r \in [l]$, we have
\begin{equation} \label{eq:D_r_f}
	\begin{aligned}
		D^r f(\boldsymbol{x}) &= \int_{\mathbb{R}^M} \widehat{D^rf} ( \boldsymbol{y}) \exp\left( 2 \pi \ii \boldsymbol{y}^{\top} \boldsymbol{x} \right)  \dif\boldsymbol{y} \\
		&= \int_{\mathbb{R}^M} \widehat{D^r f} ( \beta\boldsymbol{y}) \exp\left( 2 \pi \beta \ii \boldsymbol{y}^{\top} \boldsymbol{x} \right)  \dif ( \beta \boldsymbol{y}) \\
		&= \int_{\mathbb{R}^M} (2\pi\beta\ii\boldsymbol{y})^r \widehat{f}( \beta\boldsymbol{y} ) \exp\left( 2 \pi \beta \ii \boldsymbol{y}^{\top} \boldsymbol{x} \right) |\beta|^M  \dif\boldsymbol{y} \\
		&= \int_{\mathbb{R}^M} \left[ \boldsymbol{y}^r |\beta|^M \widehat{f}( \beta\boldsymbol{y} ) \right] \left[ (2\pi\beta\ii)^r \exp\left( 2 \pi \beta \ii \boldsymbol{y}^{\top} \boldsymbol{x} \right) \right] \dif\boldsymbol{y} \\
		&= \int_{\mathbb{R}^M} \frac{ \boldsymbol{y}^r |\beta|^M \widehat{f}( \beta\boldsymbol{y} ) }{ \widehat{f_e}(\beta) } \left[ \widehat{D^rf_e}(\beta) \exp\left( 2 \pi \beta \ii \boldsymbol{y}^{\top} \boldsymbol{x} \right) \right] \dif\boldsymbol{y} \\
		&= \int_{\mathbb{R}^M} \frac{ \boldsymbol{y}^r |\beta|^M \widehat{f}( \beta\boldsymbol{y} ) }{ \widehat{f_e}(\beta) } \left[ \int_{\mathbb{R}} D^r f_e(\alpha) \exp\left( -2 \pi \ii \beta \alpha \right) \dif\alpha \right] \exp\left( 2 \pi \beta \ii \boldsymbol{y}^{\top} \boldsymbol{x} \right) \dif\boldsymbol{y} \ ,
	\end{aligned}
\end{equation}
where $\alpha, \beta \in \mathbb{R}$, and the above equations hold from the Fourier transforms and some of their properties. By taking the real part of Eq.~\eqref{eq:D_r_f}, we have
\begin{equation} \label{eq:f_step_one}
	D^r f(\boldsymbol{x}) = \int_{\mathbb{R}^M} \int_{\mathbb{R}} \boldsymbol{y}^r D^r f_e(\alpha) \mathcal{K}(\alpha,\beta,\boldsymbol{y}) \dif\alpha \dif\boldsymbol{y} \ ,
\end{equation}
where
\[
\mathcal{K}(\alpha,\beta,\boldsymbol{y}) = \frac{ |\beta|^M \widehat{f}( \beta\boldsymbol{y} ) \exp\left[ 2 \pi \beta \ii (\boldsymbol{y}^{\top} \boldsymbol{x} - \alpha) \right]}{ \widehat{f_e}(\beta) } \ .
\]
In this proof, we set
\begin{equation} \label{eq:ua_proof}
	\alpha = \boldsymbol{y}^{\top} \boldsymbol{x} + \hbar ,
	\quad
	\boldsymbol{y} = \mathbf{W}_{k,[M]}^{\top} ,
	\quad
	\hbar = -\frac{1}{\tau_m} \sum_{i\in[N]} \exp\left( -\frac{t-t'}{\tau_m} \right) \mathbf{V}_{ki} \boldsymbol{s}_i(t')   \ ,	
\end{equation}
and the $k^{\textrm{th}}$ element of vector $\boldsymbol{x}$ equals to a temporal-weighted average of $\mathbf{I}_k(t)$ at time interval $[t',t]$
\[
\boldsymbol{x}_k = \int_{t'}^{t} \exp\left( -\frac{s-t'}{\tau_m} \right) \mathbf{I}_k(s) \dif s .
\] 
Thus, we have
\[
\mathcal{K}(\alpha,\beta,\boldsymbol{y}) = \frac{ |\beta|^M \widehat{f}( \beta\boldsymbol{y} ) \exp\left( 2 \pi \beta \hbar \ii \right)}{ \widehat{f_e}(\beta) } \overset{\mathrm{def}}{=} \mathcal{K}_{\beta}(\hbar,\boldsymbol{y})
\quad\text{and}\quad
\sup_{\boldsymbol{x} \in K}|\boldsymbol{x}| \leq C_x \ .
\]
Based on Eq.~\eqref{eq:f_step_one}, we can construct a family of approximation functions of the form
\begin{equation} \label{eq:f_kappa}
	f_{\kappa}( \boldsymbol{x} ) = \int_{\mathcal{B}_1} \int_{\mathcal{B}_2} f_e(\boldsymbol{y}^{\top} \boldsymbol{x} + \hbar) \mathcal{K}_{\beta}(\hbar,\boldsymbol{y}) \dif\hbar \dif\boldsymbol{y} \ ,
\end{equation}
where $\mathcal{B}_1 = \{\boldsymbol{x} \mid \boldsymbol{x} \preccurlyeq \kappa \}$ and $\mathcal{B}_2 = \{x \mid x \preccurlyeq (C_x M+1) \kappa \}$. Thus, we have
\begin{equation} \label{eq:Df_kappa}
	D^r f_{\kappa}( \boldsymbol{x} ) = \int_{\mathcal{B}_1} \int_{\mathcal{B}_2} \boldsymbol{y}^r D^r f_e(\boldsymbol{y}^{\top} \boldsymbol{x} + \hbar) \mathcal{K}_{\beta}(\hbar,\boldsymbol{y}) \dif\hbar \dif\boldsymbol{y} \ .
\end{equation}
It suffices to prove that $D^r f_{\kappa} \rightarrow D^r f$ uniformly on $K$, as ${\kappa} \to \infty$. Now
\[
\begin{aligned}
	D^r f_{\kappa} (\boldsymbol{x}) - D^r f (\boldsymbol{x}) =&~ \int_{\mathbb{R}^M / \mathcal{B}_1} \int_{\mathcal{R}} \boldsymbol{y}^r D^r f_e(\boldsymbol{y}^{\top} \boldsymbol{x} + \hbar) \mathcal{K}_{\beta}(\hbar,\boldsymbol{y}) \dif\hbar \dif\boldsymbol{y} \\
	&+ \int_{\mathcal{B}_1} \int_{\mathbb{R} / \mathcal{B}_2} \boldsymbol{y}^r D^r f_e(\boldsymbol{y}^{\top} \boldsymbol{x} + \hbar) \mathcal{K}_{\beta}(\hbar,\boldsymbol{y}) \dif\hbar \dif\boldsymbol{y} \\
	\overset{\mathrm{def}}{=}&~ \mathcal{R}_1 + \mathcal{R}_2 \ .
\end{aligned}
\]
For $R_1$, one has
\[
\begin{aligned}
	|\mathcal{R}_1| &= \left| \int_{\mathbb{R}^M / \mathcal{B}_1} \int_{\mathcal{R}} \boldsymbol{y}^r D^r f_e(\boldsymbol{y}^{\top} \boldsymbol{x} + \hbar) \mathcal{K}_{\beta}(\hbar,\boldsymbol{y}) \dif\hbar \dif\boldsymbol{y} \right| \\
	&\leq \int_{\mathbb{R}^M / \mathcal{B}_1} \left| \boldsymbol{y}^r \right| \left| \int_{\mathcal{R}} D^r f_e(\boldsymbol{y}^{\top} \boldsymbol{x} + \hbar) \mathcal{K}_{\beta}(\hbar,\boldsymbol{y}) \dif\hbar \right| \dif\boldsymbol{y} \\
	&\leq \int_{\mathbb{R}^M / \mathcal{B}_1} \left| \boldsymbol{y}^r \right| \left| \int_{\mathcal{R}} D^r f_e(\boldsymbol{y}^{\top} \boldsymbol{x} + \hbar) \dif\hbar \right| \left| \frac{ |\beta|^M \widehat{f}( \beta\boldsymbol{y} ) }{ \widehat{f_e}(\beta) } \right|  \dif\boldsymbol{y} \\
	&\leq \left\| D^r f_e(\boldsymbol{y}^{\top} \boldsymbol{x} + \hbar) \right\|_{1,\mathbb{R}} \int_{\mathbb{R}^M / \mathcal{B}_1} \left| \frac{ |\beta|^M \boldsymbol{y}^r  \widehat{f}( \beta\boldsymbol{y} ) }{ \widehat{f_e}(\beta) } \right|  \dif\boldsymbol{y} \\
	&\leq \left\| D^r f_e(\boldsymbol{y}^{\top} \boldsymbol{x} + \hbar) \right\|_{1,\mathbb{R}} \int_{\mathbb{R} / \tilde{\mathcal{B}}_1} \left| \frac{ |\beta \boldsymbol{y}|^r \widehat{f}( \beta\boldsymbol{y} ) }{ \widehat{f_e}(\beta) |\beta|^r } \right| \dif(\beta\boldsymbol{y})  \\
	&\leq \frac{\left\| D^r f_e(\boldsymbol{y}^{\top} \boldsymbol{x} + \hbar) \right\|_{1,\mathbb{R}}}{\left| \widehat{f_e}(\beta) |\beta|^r \right|} \int_{\mathbb{R} / \tilde{\mathcal{B}}_1} \left| \boldsymbol{y}^r \widehat{f}(\boldsymbol{y}) \right| \dif\boldsymbol{y} \ ,
\end{aligned}
\]
where $\tilde{\mathcal{B}}_1 = \{ \beta\boldsymbol{x} \mid \beta\boldsymbol{x} \preccurlyeq \beta \kappa \} $. For $R_2$, one has
\[
\begin{aligned}
	|\mathcal{R}_2| &= \left| \int_{\mathcal{B}_1} \int_{\mathbb{R} / \mathcal{B}_2} \boldsymbol{y}^r D^r f_e(\boldsymbol{y}^{\top} \boldsymbol{x} + \hbar) \mathcal{K}_{\beta}(\hbar,\boldsymbol{y}) \dif\hbar \dif\boldsymbol{y} \right| \\
	&\leq \int_{\mathcal{B}_1} \left| \int_{\mathbb{R} / \mathcal{B}_2} D^r f_e(\boldsymbol{y}^{\top} \boldsymbol{x} + \hbar) \dif\hbar \right| \left| \frac{ |\beta|^M \boldsymbol{y}^r \widehat{f}( \beta\boldsymbol{y} ) }{ \widehat{f_e}(\beta) } \right| \dif\boldsymbol{y} \\
	&\leq \int_{\mathbb{R} / \tilde{\mathcal{B}}_2} \left| D^r f_e(\mu) \right| \dif\mu \cdot
	\int_{\tilde{\mathcal{B}}_1} \left| \frac{ |\beta\boldsymbol{y}|^r \widehat{f}( \beta\boldsymbol{y} ) }{ \widehat{f_e}(\beta) |\beta|^r } \right| \dif(\beta\boldsymbol{y}) \\
	&\leq \int_{\mathbb{R} / \tilde{\mathcal{B}}_2} \left| D^r f_e(\mu) \right| \dif\mu
	~\frac{\left\| D^r f_e(\mu) \right\|_{1,\tilde{\mathcal{B}}_1}}{\left| \widehat{f_e}(\beta) |\beta|^r \right|} \ ,
\end{aligned}
\]
where $\mu = \boldsymbol{y}^{\top} \boldsymbol{x} + \hbar$ and $\tilde{\mathcal{B}}_2 = \{x \mid x \preccurlyeq \kappa \}$ since $|\mu| \geq \kappa$. Summing up the inequalities above, we have
\[
\begin{aligned}
	\sup_{\boldsymbol{x} \in K} \left| D^r f_{\kappa} (\boldsymbol{x}) - D^r f (\boldsymbol{x}) \right| &\leq \frac{ C_{\kappa}^1 + C_{\kappa}^2}{\left| \widehat{f_e}(\beta) |\beta|^r \right|}
\end{aligned}
\]
with
\[
C_{\kappa}^1 = \left\| D^r f_e(\boldsymbol{y}^{\top} \boldsymbol{x} + \hbar) \right\|_{1,\mathbb{R}} \int_{\mathbb{R} / \tilde{\mathcal{B}}_1} \left| \boldsymbol{y}^r \widehat{f}(\boldsymbol{y}) \right| \dif\boldsymbol{y}
\quad\text{and}\quad
C_{\kappa}^2 = \left\| D^r f_e(\mu) \right\|_{1,\tilde{\mathcal{B}}_1} \int_{\mathbb{R} / \tilde{\mathcal{B}}_2} \left| D^r f_e(\mu) \right| \dif\mu \ ,
\]
which tends to 0 as $\kappa \to \infty$. Given $\kappa$, it suffices to construct a series of approximations to $f_{\kappa}$ in Eq.~\eqref{eq:f_kappa}. Formally, we define
\[
\tilde{f}_{\kappa}^n(\boldsymbol{x}) = \sum_{\boldsymbol{\mu} \in \mathcal{U}} \tilde{\beta} f_e( \tilde{\boldsymbol{y}}^{\top} \boldsymbol{x} + \tilde{\hbar} ) \ ,
\]
where
\[
\left\{\begin{aligned}
	& \boldsymbol{\mu} = (\mu_1,\mu_2,\dots,\mu_M)^{\top} \quad\text{with}\quad \mu_i \in [-n,n] \cap \mathbb{Z} \quad\text{for}\quad i \in [M] ,\\
	& \tilde{\beta} = (C_x M +1) (\kappa/n)^{M+1} \mathcal{K}_{\beta}(\tilde{\hbar},\tilde{\boldsymbol{y}}) \ ,\\
	& \tilde{\boldsymbol{y}} = \boldsymbol{\mu} \kappa/n \ ,\\
	& \tilde{\hbar} = \mu^* (C_x M +1) \kappa/n \quad\text{with}\quad \mu^* \in [-n,n] \cap \mathbb{Z} \ .\\
\end{aligned} \right.
\]
It is observed that $\tilde{f}_{\kappa}^n$ belongs to the set of expressive functions, and
\begin{equation} \label{eq:Df_kappa_tilde}
	D^r \tilde{f}_{\kappa}^n (\boldsymbol{x}) = \sum_{\boldsymbol{\mu} \in \mathcal{U}} (C_x M +1)  (\kappa/n)^{M+1}~ \tilde{\boldsymbol{y}}^r D^r f_e( \tilde{\boldsymbol{y}}^{\top} \boldsymbol{x} + \tilde{\hbar} ) ~\mathcal{K}_{\beta}(\tilde{\hbar},\tilde{\boldsymbol{y}}) \ .
\end{equation}
Next, we are going to prove that $D^r \tilde{f}_{\kappa}^n \rightarrow D^r f_{\kappa}$ uniformly on $K$, as $n \to \infty$. For simplicity, we define the following function
\[
G_{\beta}(\boldsymbol{x},\boldsymbol{y},\hbar) = \boldsymbol{y}^r D^r f_e(\boldsymbol{y}^{\top} \boldsymbol{x} + \hbar) \mathcal{K}_{\beta}(\hbar,\boldsymbol{y}) \ .
\]
Thus, Eq.~\eqref{eq:Df_kappa} and Eq.~\eqref{eq:Df_kappa_tilde} become
\[
D^r f_{\kappa} (\boldsymbol{x}) = \sum_{\boldsymbol{\mu} \in \mathcal{U}} \int_{\mathcal{B}_3} G_{\beta}(\boldsymbol{x},\boldsymbol{y},\hbar) \dif\hbar \dif\boldsymbol{y}
\quad\text{and}\quad
D^r \tilde{f}_{\kappa}^n (\boldsymbol{x}) = \sum_{\boldsymbol{\mu} \in \mathcal{U}} \int_{\mathcal{B}_3} G_{\beta}(\boldsymbol{x},\tilde{\boldsymbol{y}},\tilde{\hbar}) \dif\hbar \dif\boldsymbol{y}  \ ,
\]
respectively, where $\cup_{\boldsymbol{\mu} \in \mathcal{U}} \mathcal{B}_3 = \{ (x_0, x_1, \dots, x_M) \mid x_0 \in \mathcal{B}_2, ~(x_1, \dots, x_M)^{\top} \in \mathcal{B}_1 \} \subset \mathbb{R}^{M+1} $. Hence, one has
\[
\mathop{\sup}_{(\hbar,\boldsymbol{y}), (\tilde{\hbar},\tilde{\boldsymbol{y}}) \in \mathcal{B}_3}
\left| G_{\beta}(\boldsymbol{x},\boldsymbol{y},\hbar) - G_{\beta}(\boldsymbol{x},\tilde{\boldsymbol{y}},\tilde{\hbar}) \right| < \infty \ .
\]
Let
\[
C_{\kappa}^n (\delta) \quad \overset{\mathrm{def}}{=} \mathop{\sup}_{(\hbar,\boldsymbol{y}), (\tilde{\hbar},\tilde{\boldsymbol{y}}) \in \mathcal{B}_3
	\atop
	|(\hbar,\boldsymbol{y}) - (\tilde{\hbar},\tilde{\boldsymbol{y}})| \leq \delta^{m+1}}
\left| G_{\beta}(\boldsymbol{x},\boldsymbol{y},\hbar) - G_{\beta}(\boldsymbol{x},\tilde{\boldsymbol{y}},\tilde{\hbar}) \right| .
\]
Thus, we have
\[
\begin{aligned}
	\left| D^r \tilde{f}_{\kappa}^n (\boldsymbol{x}) - D^r f_{\kappa} (\boldsymbol{x}) \right|
	&\leq \sum_{\boldsymbol{\mu} \in \mathcal{U}} \int_{\mathcal{B}_3} \left| G_{\beta}(\boldsymbol{x},\boldsymbol{y},\hbar) - G_{\beta}(\boldsymbol{x},\tilde{\boldsymbol{y}},\tilde{\hbar}) \right| \dif\hbar \dif\boldsymbol{y} \\
	&\leq \sum_{\boldsymbol{\mu} \in \mathcal{U}} \int_{\mathcal{B}_3} C_{\kappa}^n (\kappa / n) \dif\hbar \dif\boldsymbol{y} \\
	&\leq C_{\kappa}^n (\kappa / n) \sum_{\boldsymbol{\mu} \in \mathcal{U}} \int_{\mathcal{B}_3} \dif\hbar \dif\boldsymbol{y} \\
	&\leq C_{\kappa}^n (\kappa / n) ~(2n)^{M+1}~ (C_x M +1) (\kappa / n)^{M+1} ,\\
\end{aligned}
\]
where the last inequality holds from
\[
\int_{\mathcal{B}_3} \dif\hbar \dif\boldsymbol{y} = (C_x M +1) (\kappa / n)^{M+1}
\quad\text{and}\quad
|\mathcal{U}|_{\#} = (2n)^{M+1} .
\]
Further, we can obtain
\[
\sup_{\boldsymbol{x} \in K} \left| D^r \tilde{f}_{\kappa}^n (\boldsymbol{x}) - D^r f_{\kappa} (\boldsymbol{x}) \right| \leq (C_x M +1) (2 \kappa)^{M+1} ~C_{\kappa}^n (\kappa / n) \ , 
\]
which tends to 0 as $n \to \infty$.

Therefore, we can prove that the set of concerned functions is dense in $\mathcal{C}^r(K,\mathbb{R})$ for all $r \in [l]$, by taking double limits $n \to \infty$ before $\kappa \to \infty$. Besides, $\mathcal{C}^r(K,\mathbb{R})$ is dense in $\mathcal{C}^0(K,\mathbb{R})$. According to the transitivity of dense operations, we can finish this proof. $\hfill\square$

\section{Full Proof for Theorem~\ref{thm:bifurcation_bound}} \label{app:bifurcation_bound}
Following the proof of \citep[Theorem 2]{zhang2021:bsnn}, we have the algebraic representation of linear ScSNNs.
\[
\frac{\dif \boldsymbol{u}}{\dif t} = \mathbf{L}_{\boldsymbol{v}}\boldsymbol{u} + G(\boldsymbol{u},\mathbf{V}) \quad\text{with}\quad \mathbf{L}_{\boldsymbol{v}} = \mathbf{A} + \mathbf{B}_{N} \quad\text{and}\quad G(\boldsymbol{u},\mathbf{V})=o(|\boldsymbol{u}|) ,
\]
where
\[
\mathbf{A} = \begin{pmatrix}
	- 1 / \tau_m &  & \\
	& \ddots & \\
	&   & - 1 / \tau_m
\end{pmatrix}_{N \times N} \quad\text{and}\quad \mathbf{B}_{N} = \begin{pmatrix}
	0      & \mathbf{V}_{1,2}    & \dots   & \mathbf{V}_{1,N} \\
	\mathbf{V}_{2,1} &    0      & \dots   & \mathbf{V}_{2,N} \\
	\vdots    & \vdots       & \ddots   & \vdots       \\
	\mathbf{V}_{N,1} & \mathbf{V}_{N,(N-1)}     & \dots    & 0
\end{pmatrix} \ .
\]
Suppose that the eigenvalues of the matrix $\mathbf{B}_{N}$ are $\beta_1,\dots,\beta_N$. So the eigenvalue $\rho_i$ of $\mathbf{}_{\boldsymbol{v}}$ can be calculated as the sum of that of $A$ and that of $\mathbf{B}_N$, that is, $\rho_i = 1 / \tau_m + \beta_i$ for $i \in [N]$. Zhang et al.~\citep{zhang2021:bsnn} has elucidated the bifurcation solutions relative to the eigenvalues. Identifying the number of indefinite eigenvalues can be 

To compute the lower bound, we follow the idea of ``Divide and Conquer''. Let $H^*(N)$ denote the number of indefinite eigenvalues of $N \times N$ matrix $\mathbf{B}_N$. Suppose $N=K_1+K_2$ for $K_1, K_2\in \mathbb{N}^+$, then we have
\[
H^*(N) \geq H^*(K_1) + H^*{K_2} \ ,
\]
where the concerned matrix $\mathbf{B}_N$ is divided into two sub-matrices
\[
\mathbf{B}_{K_1} = \begin{pmatrix}
	0      & \mathbf{V}_{1,2}    & \dots   & \mathbf{V}_{1,K_1} \\
	\mathbf{V}_{2,1} &    0      & \dots   & \mathbf{V}_{2,K_1} \\
	\vdots    & \vdots       & \ddots   & \vdots       \\
	\mathbf{V}_{K_1,1} & \mathbf{V}_{K_1,(K_1-1)}     & \dots    & 0
\end{pmatrix}
\]
and
\[
\mathbf{B}_{K_2} = \begin{pmatrix}
	0      & \mathbf{V}_{(K_1+1),(K_1+2)}    & \dots   & \mathbf{V}_{(K_1+1),N} \\
	\mathbf{V}_{(K_1+2),(K_1+1)} &    0      & \dots   & \mathbf{V}_{(K_1+2),N} \\
	\vdots    & \vdots       & \ddots   & \vdots       \\
	\mathbf{V}_{N,(K_1+1)} & \mathbf{V}_{N,(K_1+2)}     & \dots    & 0
\end{pmatrix} \ .
\]
So on and so forth, we can compute the worst case as $cN\log N$ where $c \in (0,1/2)$. This completes the proof.  $\hfill\square$

\section{The Computations for the Simulation Experiment  in Subsection~\ref{subsec:approximation_ScSNN}}  \label{app:ode_simulation}
This section shows the whole process of solving the differential equations. In the case of two neurons, we have the algebraic equations as follows
\[
\tau_m \frac{\dif \boldsymbol{u}(t)}{\dif t} = - \boldsymbol{u}(t) + \mathbf{V} \ ,
\]
where $\mathbf{V}$ is of the following parameterized form
\[
\mathbf{V}=\begin{pmatrix}
	0&\mathbf{V}_{12}\\ 
	\mathbf{V}_{21}&0
\end{pmatrix} \ .
\]
The first step is to calculate the eigenvalues and eigenvectors of matrix $\mathbf{V}$ according to
\[
\begin{vmatrix}
	\beta\mathbf{E}_2-\mathbf{V}
\end{vmatrix}
=
\begin{vmatrix}
	\beta&-\mathbf{V}_{12}\\ 
	-\mathbf{V}_{21}&\beta
\end{vmatrix} = 0 \ ,
\]
where $\mathbf{E}_2$ is a $2\times2$ unit matrix.

\vspace{0.2 cm}
\noindent\textbf{Two Simple Roots.} Set $\begin{vmatrix} \beta\mathbf{E}_2-\mathbf{V}\end{vmatrix}=0$, we get the eigenvalues $\beta_1=\sqrt{\mathbf{V}_{12}\mathbf{V}_{21}}$ and $\beta_2=\sqrt{\mathbf{V}_{12}\mathbf{V}_{21}}$. Plugging $\beta_1$ into the matrix
\[
\beta_1\mathbf{E}_2-\mathbf{V}=
\begin{pmatrix}
	\sqrt{ \mathbf{V}_{12}\mathbf{V}_{21} }&-\mathbf{V}_{12}\\ 
	-\mathbf{V}_{21} & \sqrt{ \mathbf{V}_{12}\mathbf{V}_{21} }
\end{pmatrix} \ ,
\]
the corresponding eigenvector of $\beta_1$ is $\boldsymbol{\eta}_1 = (\sqrt{\mathbf{V}_{12}},\sqrt{\mathbf{V}_{21}})$. Analogously, the eigenvector of $\beta_2$ is $\boldsymbol{\eta}_2 = (-\sqrt{\mathbf{V}_{12}},\sqrt{\mathbf{V}_{21}})$.

For $\mathbf{M}(\mathbf{V},\tau_m)$, the eigenvalues are $\rho_1 = -1 + \sqrt{ \mathbf{V}_{12}\mathbf{V}_{21} }$, $\rho_2 = -1 - \sqrt{ \mathbf{V}_{12}\mathbf{V}_{21} }$ provided $\tau_m=1$, and the eigenvectors are the same. Notice that we here assume $\rho_1 \neq \rho_2$, i.e., $\mathbf{V}_{12}\mathbf{V}_{21} \neq 0$. Then, we can obtain the general solution of the concerned differential equations as follows
\[
\boldsymbol{u}(t) = C_1\boldsymbol{\eta}_1^T{\e^{\rho_1t}}+C_2\boldsymbol{\eta}_2^T{\e^{\rho_2t}} \ ,
\]
or more explicitly,
\[
\left\{~\begin{aligned}
	u_1(t)&=C_1\sqrt{\mathbf{V}_{12}}{\e^{\rho_1t}}-C_2\sqrt{\mathbf{V}_{12}}{\e^{\rho_2t}} \ ,\\
	u_2(t)&=C_1\sqrt{\mathbf{V}_{21}}{\e^{\rho_1t}}+C_2\sqrt{\mathbf{V}_{21}}{\e^{\rho_2t}} \ ,
\end{aligned}\right.
\]
where $\boldsymbol{u} = (u_1, u_2)^{\top}$ and $C_1, C_2 \in \mathbb{R}$. 

Finally, some special values can be substituted into the expressions above. Given $\mathbf{V}_{12}=1$ and $\mathbf{V}_{21}=1$, we have $\rho_1=0 $ and $\rho_2=-2$, the solution $\boldsymbol{u}$ can be written as
\[
\left\{~\begin{aligned}
	u_1(t)&=C_1-C_2{\e^{-2t}} \ ,\\
	u_2(t)&=C_1+C_2{\e^{-2t}} \ .
\end{aligned}\right.
\] 
Provided the initial point $\boldsymbol{u}(0) = (6,3)$, the ultimate solutions are
\[
\left\{~\begin{aligned}
	u_1(t) &= \frac{3}{2}{\e^{-2t}} + \frac{9}{2} \ , \\
	u_2(t) &= -\frac{3}{2}{\e^{-2t}} + \frac{9}{2} \ .
\end{aligned}\right.
\]

\vspace{0.2 cm}
\noindent\textbf{One Double Root.} If $\mathbf{V}_{12} \mathbf{V}_{21}=0$, we have one double root, that is, $\rho=\rho_1=\rho_2=-1$. For the case of $\mathbf{V}_{12}=0$ and $\mathbf{V}_{21}=1$, we assume that the solution is 
\[
\left\{~\begin{aligned}
	u_1(t) & = (C_1+C_2t){\e^{\rho t}} \ ,\\
	u_2(t) &= (D_1+D_2t){\e^{\rho t}} \ ,
\end{aligned}\right.
\]
where $D_1, D_2 \in \mathbb{R}$. Plugging them into the original formula and use the initial point $\boldsymbol{u}(0) = (3,6)$, the final solutions are
\[
\left\{~\begin{aligned}
	u_1(t) &= 3{\e^{-t}}  \ ,\\
	u_2(t) &= 3t{\e^{-t}}+6{\e^{-t}} \ .
\end{aligned}\right.
\]

\vspace{0.2cm}
Finally, the table below lists all the motion curves of points in our simulation experiment. 
\begin{table}[!htb]
	\centering
	\caption{All the motion curves of points in our simulation experiment.}
	\begin{tabular}{c|l|l|l}
		\toprule
		Values of ($\mathbf{V}_{12}$,$\mathbf{V}_{21}$)  & $\boldsymbol{u}(0)=(3,6)$ & $\boldsymbol{u}(0)=(5,6)$ & $\boldsymbol{u}(0) = (6,3)$ \\
		\midrule
		(0,0)&$	
		\begin{aligned}
			u_1(t)&=3{\e^{-t}} \\
			u_2(t)&=6{\e^{-t}}
		\end{aligned}$
		&$	
		\begin{aligned}
			u_1(t)&=5{\e^{-t}} \\
			u_2(t)&=6{\e^{-t}}
		\end{aligned}$
		&$	
		\begin{aligned}
			u_1(t)&=6{\e^{-t}} \\
			u_2(t)&=3{\e^{-t}}
		\end{aligned}$\\
		\midrule
		(0,1)&$	
		\begin{aligned}
			u_1(t)&=3{\e^{-t}} \\
			u_2(t)&=3t{\e^{-t}}+6{\e^{-t}}
		\end{aligned}$
		&$	
		\begin{aligned}
			u_1(t)&=5{\e^{-t}} \\
			u_2(t)&=5t{\e^{-t}}+6{\e^{-t}}
		\end{aligned}$
		&$	
		\begin{aligned}
			u_1(t)&=6{\e^{-t}} \\
			u_2(t)&=6t{\e^{-t}}+3{\e^{-t}}
		\end{aligned}$\\
		\midrule
		(1,1)&$	
		\begin{aligned}
			u_1(t)&=-\frac{3}{2}{\e^{-2t}}+\frac{9}{2} \\
			u_2(t)&=\frac{3}{2}{\e^{-2t}}+\frac{9}{2} 
		\end{aligned}$
		&$	
		\begin{aligned}
			u_1(t)&=-\frac{1}{2}{\e^{-2t}}+\frac{11}{2} \\
			u_2(t)&=\frac{1}{2}{\e^{-2t}}+\frac{11}{2} 
		\end{aligned}$
		&$	
		\begin{aligned}
			u_1(t)&=\frac{3}{2}{\e^{-2t}}+\frac{9}{2} \\
			u_2(t)&=-\frac{3}{2}{\e^{-2t}}+\frac{9}{2} 
		\end{aligned}$\\
		\midrule
		(4,1)&$	
		\begin{aligned}
			u_1(t)&=\frac{15}{2}{\e^{t}}-\frac{9}{2}{\e^{-3t}} \\
			u_2(t)&=\frac{15}{4}{\e^{t}}+\frac{9}{4}{\e^{-3t}}
		\end{aligned}$
		&$	
		\begin{aligned}
			u_1(t)&=\frac{17}{2}{\e^{t}}-\frac{7}{2}{\e^{-3t}} \\
			u_2(t)&=\frac{17}{4}{\e^{t}}+\frac{7}{4}{\e^{-3t}}
		\end{aligned}$
		&$	
		\begin{aligned}
			u_1(t)&=6{\e^{t}}\\
			u_2(t)&=3{\e^{t}}
		\end{aligned}$\\
		\bottomrule
	\end{tabular}
\end{table}

\section{Full Proof for Theorem~\ref{thm:ScSNN_periodic}}  \label{app:periodic}
This section provides detailed proof for Theorem~\ref{thm:ScSNN_periodic}. Lemma~\ref{lemma:recurrence_upper} establishes based on the perturbation structure between original system~\eqref{eq:algebraic} and perturbed system~\eqref{eq:perturbed}. Lemma~\ref{lemma:fundamental} generalizes the Faa di Bruno's Formula in~\citep{weisstein2003:faa},
\[
\frac{\dif g(f)(t)}{\dif t} = \sum_{\mathcal{S}_r} C_{\mathcal{S}_r} \frac{\dif^{h(r)} g(f)(t) }{\dif t^{h(r)}} \prod_{l=1}^r \left( \frac{\dif^l f(t) }{\dif t^l} \right)^{\alpha_j} \ ,
\]
where $g,f \in \mathcal{C}^K(\mathbb{R}, \mathbb{R})$ and
\[
C_{\mathcal{S}} = \frac{r!}{\alpha_1!(\alpha_2!2!^{\alpha_2}) \dots (\alpha_r!r!^{\alpha_r})} \ ,
\]
for $\mathcal{S}_r$ denotes the set of all $r$-tuples of non-negative integers $\{\alpha_j\}_{j\in[r]}$ that satisfies
\[
\sum_j j\alpha_j = r
\quad\text{and}\quad
h(r) = \sum_j \alpha_j \ .
\]
Informally, we re-formulate the solution function $f$ as $f(t,\epsilon,u): [0,T] \times [-\epsilon_0,\epsilon_0] \times \mathbb{K} \to \mathbb{K}$. Hence, we have
\[
f(t,\epsilon, u) = u + \int_0^t F_0(s,\tilde{u}) \dif s + \sum_{k=0}^K \epsilon^k G_k(t,u) 
+ \epsilon^{K+1} \left[ \int_0^t \mathrm{reset}(s, f(s,\epsilon),\epsilon) \dif s + \mathcal{O}(1) \right] 
\]
for $f(0,\epsilon, \tilde{u}) = \tilde{u}$, especially, 
\[
f(t,\epsilon) = \tilde{u} + \int_0^t F_0(s,\tilde{u}) \dif s + \sum_{k=0}^K \epsilon^k G_k(t,\tilde{u})
+ \epsilon^{K+1} \left[ \int_0^t \mathrm{reset}(s, f(s,\epsilon),\epsilon) \dif s + \mathcal{O}(1) \right] \ ,
\]
where $G_k$ (for $k \in [K]$) is of recursive form as follows
\[
G_k(t,u) = \int_0^t \left[ F_k(s,u) + \mathcal{G} \left( D^{h(r)} F_r(s,u), G_r(s,u) \right) \right] \dif s \ ,
\]
for all $r \in [k-1]$. The Taylor expansion of $F_k(t,f(t,\epsilon,u))$ for $k \in [K-1]$ around $\epsilon = 0$ is given by
\begin{equation}  \label{eq:F_k}
	F_k(t,f(t,\epsilon,u)) = F_k(t,f(t,0,u)) + \epsilon^{K-k+1} \mathcal{O}(1) 
	+ \sum_{r=1}^{K-k} \frac{\epsilon^r}{r!} \left( \frac{\partial^r F_k(t,f(t,\epsilon,u))}{\partial \epsilon^r} \right) \bigg|_{\epsilon=0} \ .
\end{equation}
From the Faa di Bruno's Formula, we calculate the $r$-derivatives of $F_k(t,f(t,\epsilon,u))$ in $\epsilon$
\begin{equation} \label{eq:F_k_der}
	\frac{\partial^r F_k(t,f(t,\epsilon,u))}{\partial \epsilon^r} \bigg|_{\epsilon=0}
	= 
	\sum_{\mathcal{S}_r} C_{\mathcal{S}} r! \frac{\dif^{h(r)} F_k(t,f(t,\epsilon,u)) }{\dif t^{h(r)}} \bigg|_{\epsilon=0} \prod_{l=1}^r  G_l(t,u)^{\alpha_l}   \ ,
\end{equation}
for $k \in [K-1]$, where
\begin{equation} \label{eq:G_k}
	G_k(t,u) = \frac{1}{r!} \left( \frac{\partial^r f(t,\epsilon,u)}{\partial \epsilon^r} \right) \bigg|_{\epsilon=0} 
	= \int_0^t \left[ F_k(s,u) + \mathcal{G} \left( D^{h(r)} F_r(s,u), G_r(s,u) \right) \right] \dif s 
\end{equation}
with
\[
\mathcal{G} =  \sum_{r=1}^k \sum_{\mathcal{S}_r} \frac{D^{h(r)} F_{k-r}(s,u)}{\alpha_1!(\alpha_2!2!^{\alpha_2}) \dots (\alpha_r!r!^{\alpha_r})} \prod_{l=1}^r G_l(s,u)^{\alpha_l} 
\quad\text{for all}\quad
r \in [k-1] \ .
\]
Substituting Eq.~\eqref{eq:F_k_der} into Eq.~\eqref{eq:F_k}, the Taylor expansion of $F_k(t,f(t,\epsilon,u)) $ at $\epsilon = 0$ becomes
\[
F_k(t,f(t,\epsilon,u)) = F_k(t,u) + \epsilon^{K-k+1} \mathcal{O}(1) + \sum_{r=1}^{K-k} \sum_{\mathcal{S}_r} C_{\mathcal{S}} 
\epsilon^r \frac{\dif^{h(r)} F_k(t,f(t,\epsilon,u)) }{\dif t^{h(r)}} \bigg|_{\epsilon=0} \prod_{l=1}^r  G_k(t,u)^{\alpha_j}  
\]
for $k \in [K-1]$. Moreover, the above result holds for the case $k=0$. Further, for $k=K$, one has
\begin{equation}  \label{eq:F_polynomial_1}
	F_k (t,f(t,\epsilon,u)) = F_k(t,u) + \epsilon \mathcal{O}(1) \ . 
\end{equation}
Since the set $[0,T] \times [-\epsilon_0, \epsilon_0] \times \bar{U}$ is compact and $F_k(t,u)$ is locally Lipschitz in $\bar{U}$ with scale $L$
\begin{equation}  \label{eq:F_polynomial_2}
	| F_k (t,f(t,\epsilon,u)) - F_k(t,u) | \leq L |f(t,\epsilon,u) - u| = \mathcal{O}(1) \ .
\end{equation} 

Summing up the above results, we can conclude that
\begin{equation}  \label{eq:f}
	f(t,\epsilon, u) = u + \int_0^t F_0(s,\tilde{u}) \dif s + \sum_{k=0}^K \epsilon^k G_k(t,u)
	+ \epsilon^{K+1} \left[ \int_0^t \mathrm{reset}(s, f(s,\epsilon),\epsilon) \dif s + \mathcal{O}(1) \right]  \ ,
\end{equation}
where $G_k$ (for $k \in [K]$) is of recursive form as follows
\[
G_k(t,u) = \int_0^t \left[ F_k(s,u) + \mathcal{G} \left( D^{h(r)} F_r(s,u), G_r(s,u) \right) \right] \dif s \ ,
\]
for all $r \in [k-1]$, in which
\[
\mathcal{G} =  \sum_{r=1}^k \sum_{\mathcal{S}_r} \frac{D^{h(r)} F_{k-r}(s,u)}{\alpha_1!(\alpha_2!2!^{\alpha_2}) \dots (\alpha_r!r!^{\alpha_r})} \prod_{l=1}^r G_l(s,u)^{\alpha_l}  \ ,
\]
for $\mathcal{S}_r$ denotes the set of all $r$-tuples of non-negative integers $\{\alpha_j\}_{j\in[r]}$ as noted in Section~\ref{sec:notations}, satisfying
\[
\sum_j j\alpha_j = r
\quad\text{and}\quad
h(r) = \sum_j \alpha_j \ .
\]

\vspace{0.2cm}
\noindent\textbf{Finishing the Proof of Theorem~\ref{thm:ScSNN_periodic}.} Let $f(0,\epsilon) = \tilde{u}(0,\epsilon)$, which is abbreviated as $\tilde{u}$. Let $U \subset \mathbb{K}$ be a neighborhood of the critical point $\tilde{u}$ such that $G_k(t,u) \neq 0$ for all $u \in \bar{U}/\tilde{u}$ and the Brouwer degree $d_{B}(G_k,U,\tilde{u}) \neq 0$~\citep{llibre2014:higher}. For each $u \in \bar{U}$, there exists $\epsilon_0>0$ such that the function $f(t,\epsilon)$ is defined on $[0,T] \times [-\epsilon_0,\epsilon_0]$ once $\epsilon \in [-\epsilon_0,\epsilon_0]$. Thus, $f(t,\epsilon): [0,T] \times [-\epsilon_0,\epsilon_0] \to \mathbb{R}$ indicates the solution of system~\eqref{eq:perturbed}, as defined in Lemma~\ref{lemma:fundamental}. From the Existence and Uniqueness Theorem~\citep[Theorem 1.2.4]{sanders2007:hilbert}, the domain of function $f(\cdot,\epsilon)$ can be bounded according to $t \leq \inf(T, d/M(\epsilon))$, where
\[
M(\epsilon) \geq \left| \sum_{k=1}^K \epsilon^k F_k(t,f) + \epsilon^{K+1} \mathrm{reset} (t, \epsilon,u) \right| \ .
\]
Obviously, we can ensure $\inf(T, d/M(\epsilon)) = T$ by taking a sufficiently large $d/M(\epsilon)$ as $\epsilon$ is sufficiently small. On the one hand, based on the continuity of the solution $f(t,\epsilon)$ and the compactness of the set $[0,T] \times [-\epsilon_0,\epsilon_0]$, there exists an image set $\mathbb{K}$ such that $f(t,\epsilon) \in 
\mathbb{K}$, that is, $f(t,\epsilon):[0,T] \times [-\epsilon_0,\epsilon_0] \to  \mathbb{K}$. Informally, we can re-formulate the solution function $f$ as $f(t,\epsilon,u): [0,T] \times [-\epsilon_0,\epsilon_0] \times \mathbb{K} \to \mathbb{K}$ throughout this proof. On the other hand, based on the continuity of the function $\mathrm{reset}$, we have
\[
| \mathrm{reset}(s,\epsilon,f) | \leq \max \{ |\mathrm{reset}(t,\epsilon,u) | \} = N  \ .
\]
for all $ (t,\epsilon,u) \in [0,T] \times [-\epsilon_0,\epsilon_0] \times \mathbb{K}$. Further, we have
\[
\left| \int_0^T \mathrm{reset}(s,\epsilon,f) \dif s \right| \leq \int_0^T |  \mathrm{reset}(s,\epsilon,u) | \dif s = T N \ ,
\]
which implies that
\begin{equation}  \label{eq:O1_app}
	\int_0^T \mathrm{reset}(s,\epsilon,f) \dif s  = \mathcal{O}(1) \ .
\end{equation}
Provided $\epsilon g(u,\epsilon) = f(T,\epsilon,u) - u$, then from Lemma~\ref{lemma:fundamental} and Eq.~\eqref{eq:O1}, we have
\[
g(u,\epsilon) = \sum_{k=1}^{K} \epsilon^{k-1} G_k(T,u) + \epsilon^K \mathcal{O}(1) \ ,
\]
where $u \in \bar{U}/\tilde{u}$. It is self-evident that when $T=2\pi$, it holds that $U \subset \mathbb{K}$ is a neighborhood of the critical point $\tilde{u}$ satisfying
\begin{itemize}
	\item [(1)] $G_k(t,u) \neq 0$ for all $u \in \bar{U}/\tilde{u}$,
	\item [(2)] the Brouwer degree $d_{B}(G_k,U,\tilde{u}) \neq 0$~\citep{llibre2014:higher}.
\end{itemize}
Hence, we have
\[
g(u,\epsilon) = \sum_{k=r}^{K} \epsilon^{k-1} G_k(2\pi,u) + \epsilon^K \mathcal{O}(1) \ ,
\]
for the case that $G_l \equiv 0$ for $l \in [r-1]$ and $r\in[k]$ but $G_r \neq 0$. Thus, it is self-evident that $f(t,\epsilon)$ is an $2\pi$-periodic solution if and only if $g(u,\epsilon) = 0$. From~\citep[Lemma 6]{llibre2014:higher}, we have
\[
d_B(G_r,U,\tilde{u}) = d_B(g(u,\epsilon),U,\tilde{u}) \neq 0 \ ,
\]
for $|\epsilon| >0 $ sufficiently small. Further, from~\citep[Charpter VIII]{crawford1991:bifurcation}, there exists some $u(\epsilon) \in U$ such that $g(u(\epsilon),\epsilon) = 0$. Therefore, we can conclude that $f(t,\epsilon,u(\epsilon))$ is a periodic solution of system~\eqref{eq:perturbed}, and then, pick up a collection of $u(\epsilon)$ such that $u(\epsilon) \to \tilde{u}$ as $\epsilon \to 0$. This completes this proof.  $\hfill\square$

\section{Full Proof for Theorem~\ref{thm:ScSNN_lower}}  \label{app:bound}
We begin the proof of Theorem~\ref{thm:ScSNN_lower} with some useful lemmas.
\begin{lemma} \label{lemma:degree}
	The ScSNN-PIRATE model in Eq.~\eqref{eq:ScSNN} with $n$-order polynomial bifurcation fields coincides with a Hamiltonian system of degree $\textrm{free}(n) = 2^n-1$.
\end{lemma}

\begin{lemma} \label{lemma:recursive}
	The Hamiltonian system led by Eq.~\eqref{eq:ScSNN} with $n$-order polynomial bifurcation fields has at least $P(n)$ limit cycles in which
	\begin{equation} \label{eq:P}
		P(n+1) = P(n) + (\textrm{free}(n) -1 )^2 + \textrm{free}(n)^2 \ .
	\end{equation}
\end{lemma}
Lemma~\ref{lemma:recursive} provides a recursive sequence $P(n)$ relative to the freedom degree $2^n-1$ of system~\eqref{eq:H}, which contributes to the lower bound of $H(n)$.

The basic theory of the perturbation of planar Hamiltonian systems is well known. In general, we can reload Eq.~\eqref{eq:H} as
\[
\left\{\begin{aligned}
	\frac{\dif u_k(t)}{\dif t} &= - \frac{\partial \mathcal{H}(u_k, u_{k'})}{\partial u_{k'}} + \epsilon f_{\epsilon}(u_k, u_{k'}) \ ,\\
	\frac{\dif u_{k'}(t)}{\dif t} &= \frac{\partial \mathcal{H}(u_k, u_{k'})}{\partial u_k} + \epsilon g_{\epsilon}(u_k, u_{k'}) \ .
\end{aligned}\right.  
\]
We are going to show the degree of system Eq.~\eqref{eq:H}. We start this proof with an example of $n=2$
\[
\mathcal{H}(u_k,u_k') = (u_k^2 - 1)^2 + (u_k'^2 - 1)^2 \ .
\]
Thus, the unperturbed system has nine critical points, that is, $(x,y)$ for $x,y \in \{-1,0,1\}$, of which 5 points are non-degenerate, that is, $(\pm 1, \pm 1)$ and $(0,0)$. Therefore, we can claim that there is a polynomial $f_{\epsilon}$ of degree 3, which meets the degree result $2^2-1 =3 $ of Lemma~\ref{lemma:degree}, such that
\[
\left\{\begin{aligned}
	\frac{\dif u_k(t)}{\dif t} &= - \frac{\partial \mathcal{H}(u_k, u_{k'})}{\partial u_{k'}} + \epsilon f_{\epsilon}(u_k, u_{k'}) \\
	\frac{\dif u_{k'}(t)}{\dif t} &= \frac{\partial \mathcal{H}(u_k, u_{k'})}{\partial u_k} 
\end{aligned}\right.  
\]
has limit cycles around critical points $(-1,-1)$, $(0,0)$, or $(1,1)$, if $\epsilon$ is sufficiently small but $\epsilon \neq 0$. This claim is self-evident if provided
\[
f_{\epsilon}(u_k,u_k') = \frac{1}{3}(u_k - u_k')^2 - \epsilon (u_k - u_k') \ .
\]
Next, it suffices to develop the above result to the case of $n$ via mathematical induction. Then we have the following proposition.
\begin{proposition}
	The system Eq.~\eqref{eq:H} has non-degenerate critical points at the origin and at $2^n-2$ other points on each axis, all of which lie within $\{ (x,y) \mid |x| \leq 2^{n-1}  ~\text{and}~  |y| \leq 2^{n-1} \} $.
\end{proposition}
Based on this proposition, we can conclude that
\[
\textrm{free}(n) = 2^{n-1} + 1 \ .
\]
Consider a clear-cut case that
\[
\left\{\begin{aligned}
	\textrm{Poly}(u_k;1) &= u_k \\
	\textrm{Poly}(u_k;n) &= \textrm{Poly}(u_k^2 - 2^{n-2};n-1), \quad\text{for}\quad n \geq 2 \ , \\
\end{aligned}\right.
\]
then system Eq.~\eqref{eq:H} induces a singular transformation
\[
(u_k, u_k')  \mapsto (u_k^2 - 2^{n-2}, u_k'^2 - 2^{n-2}) 
\]
that is,
\[
(u_k, u_k')  \mapsto \left[ u_k^2 - (\textrm{free}(n) -1 ), u_k'^2 - (\textrm{free}(n) -1 ) \right]
\]
for $n \geq 2$. Further, it is easy to calculate the recursive sequence of Eq.~\eqref{eq:P} as follows
\[
P(n+1) = P(n) + (\textrm{free}(n) -1 )^2 + \textrm{free}(n)^2 \ .
\]

\vspace{0.2cm}
\noindent\textbf{Finishing the Proof of Theorem~\ref{thm:ScSNN_lower}}. From Lemma~\ref{lemma:recursive}, the recursive formation of $P(n)$ indicates the lower bound of $H(n)$. Let $P(n) = 4^n Q(n)$. Then Eq.~\eqref{eq:P} becomes 
\[
Q(n+1) = \frac{1}{4} Q(n) + \frac{1}{2} - \frac{3}{2^{n+1}} + \frac{5}{4^{n+1}} \ .
\]
Further, we have $Q(2) = 3/16$ and 
\[
\begin{aligned}
	Q(n) &= \frac{1}{4} Q(n-1) + \frac{1}{2} - \frac{3}{2^n} + \frac{5}{4^n} 
	= \frac{Q(2)}{4^{n-1}} + \frac{n-2}{2} - \frac{3 (1-2^{-{n+2}})}{4} + \frac{5 (1-4^{-(n+2)})}{48}  \\
	&= \frac{Q(2)}{4^{n-1}} + \frac{n}{2} - \left(\frac{16}{5}\right)^{-n} - \frac{5 \cdot 4^{-n}}{3} - \frac{79}{48} 
	= \frac{n}{2} - \left(\frac{16}{5}\right)^{-n} - \frac{ 4^{-n}}{2} - \frac{79}{48}
\end{aligned}
\]
for $n \geq 3$. Since $H(n) \geq P(n) \geq 4^n Q(n)$ and the degree of Eq.~\eqref{eq:H} is $\textrm{free}(n) = 2^{n-1}$, we have
\[
H(2^n-1) \geq 4^{n-1} \left(2n - \frac{35}{6}\right)  + \left( \frac{16}{5} \right)^{n} - \frac{5}{3} \ .
\]
Re-substituting the variable $n$, the above inequality becomes
\[
H(n) \geq \frac{(n+1)^2}{2} \left( \log_2(n+1) - 3 \right) + 3n \ .
\]
Therefore, there exists some constant $C$ such that
\[
H(n) \geq C (n+1)^2 \ln(n+1) \ .
\]
This completes the proof. $\hfill\square$

\section{Full Proof for Corollary~\ref{coroll:example}} \label{app:example}
We begin our analysis with a recall for the concerned example system. We consider a simple case of $N=2$, $n=2$, $ m= 3$, and $K=5$, as follows:
\[
\left\{\begin{aligned}
	\frac{\dif u_1(t)}{\dif t} &= -\frac{u_1(t)}{\tau_m} + u_1(t)^2 u_2(t) + \epsilon ~\textrm{Poly}^1(\boldsymbol{u}(t);m) \ , \\
	\frac{\dif u_2(t)}{\dif t} &= -\frac{u_2(t)}{\tau_m} + u_1(t) u_2(t)^2 + \epsilon ~\textrm{Poly}^2(\boldsymbol{u}(t);m) \ ,
\end{aligned}\right.    
\]
where 
\[
\begin{aligned}
	\textrm{Poly}^i(\boldsymbol{u};3) =&~ \beta^i_{k,1} u_1 + \beta^i_{k,2} u_2 + \beta^i_{k,3} u_1^2 + \beta^i_{k,4} u_1u_2 + \beta^i_{k,5} u_2^2 \\
	&+ \beta^i_{k,6} u_1^3 + \beta^i_{k,7} u_1^2u_2 + \beta^i_{k,8} u_1u_2^2 + \beta^i_{k,9} u_2^3 \ ,
\end{aligned}
\]
for $i \in [N=2]$ and $k \in [K=5]$. Obviously, it is known as a cubic system with $m=3^{\textrm{rd}}$ polynomial perturbations. 

Here, we employ $K=5^{\textrm{th}}$ component to estimate the upper bounds of $H(n)$. From \textbf{Procedure 1-3}, we first convert Eq.~\eqref{eq:example_u} into 
\[
\frac{\partial f(t,\epsilon)}{\partial t} = \sum_{k=0}^5 \epsilon^k F_k(t,f) + \epsilon^{6} \mathrm{reset}(t,\epsilon,f) \ ,
\]
and thus, we have
\[
\frac{\partial f(t,\epsilon)}{\partial t} = F_0 + \sum_{k=1}^5 \epsilon^k F_k(t,f) + \mathcal{O}(\epsilon^6) \ .
\]
Next, we provide the detailed calculation paradigms for $f$ provided $G_k$ and $F_k$ ($k\in[K=5]$). For convenience, we consider two cases either $F_0 \equiv 0$ or $F_0 \not\equiv 0$. Let $\odot$ denote the Hadamard product. We have the sets $\mathcal{S}_r$ for $r \in [K=5]$
\[
\left\{\begin{aligned}
	\mathcal{S}_{1}&=\{1\}, \\
	\mathcal{S}_{2}&=\{(0,1),(2,0)\} \\
	\mathcal{S}_{3}&=\{(0,0,1),(1,1,0),(3,0,0)\}, \\
	\mathcal{S}_{4}&=\{(0,0,0,1),(1,0,1,0),(2,1,0,0),(0,2,0,0),(4,0,0,0)\} , \\
	\mathcal{S}_{5}&=\{(0,0,0,0,1), (1,0,0,1,0), (0,1,1,0,0), (2,0,1,0,0), (3,1,0,0,0), (1,2,0,0,0), (5,0,0,0,0) \} .
\end{aligned}\right.
\]

(i) For the case of $F_0 \equiv 0$, we have
\[
\begin{aligned}
	G_{0}(u) =&~ 0 \\
	G_{1}(u) =&~ \int_{0}^{T} F_{1}(t, u) \dif t \\
	G_{2}(u) =&~ \int_{0}^{T} F_{2}(t, u) \dif s+\frac{\partial F_{1}}{\partial u}(t, u) y_{1}(t, u) d t \\
	G_{3}(u) =&~ \int_{0}^{T}\left(F_{3}(t, u)+\frac{\partial F_{2}}{\partial u}(t, u) y_{1}(t, u)\right) \dif t + \int_{0}^{T}\left(\frac{\partial^{2} F_{1}}{\partial u^{2}}(t, u) y_{1}(t, z)^{2}+\frac{\partial F_{1}}{\partial u}(t, u) y_{2}(t, u)\right) \dif t \\
	G_{4}(u) =&~ \int_{0}^{T}\left(F_{4}(t, u)+\frac{\partial F_{3}}{\partial u}(t, u) y_{1}(t, u)\right) \dif t + \int_{0}^{T}\left(\frac{\partial^{2} F_{2}}{\partial u^{2}}(t, u) y_{1}(t, u)^{2} + \frac{\partial F_{2}}{\partial u}(t, u) y_{2}(t, u)\right) \dif t \\
	&+ \int_{0}^{T} \frac{\partial^{2} F_{1}}{\partial u^{2}}(t, u) y_{1}(t, u) \odot y_{2}(t, u) \dif t + \int_{0}^{T}\left(\frac{\partial^{3} F_{1}}{\partial u^{3}}(t, u) y_{1}(t, u)^{3}+\frac{\partial F_{1}}{\partial u}(t, u) y_{3}(t, u)\right) \dif t \\
	G_{5}(u) =&~ \int_{0}^{T}\left(F_{5}(t, u)+\frac{\partial F_{4}}{\partial u}(t, u) y_{1}(t, u)\right) \dif t \\
	&+ \int_{0}^{T}\left(\frac{\partial^{2} F_{3}}{\partial u^{2}}(t, u) y_{1}(t, u)^{2}+\frac{\partial F_{3}}{\partial u}(t, u) y_{2}(t, u)+\frac{\partial^{2} F_{2}}{\partial u^{2}}(t, u) y_{1}(t, u) \odot y_{2}(t, u)\right) \dif t \\
	&+ \int_{0}^{T}\left(\frac{\partial^{3} F_{2}}{\partial u^{3}}(t, u) y_{1}(t, u)^{3}+\frac{\partial F_{2}}{\partial u}(t, u) y_{3}(t, u)
	+\frac{\partial^{2} F_{1}}{\partial u^{2}}(t, z) y_{1}(t, u) \odot y_{3}(t, z)\right) \dif t \\
	&+ \int_{0}^{T} \frac{\partial^{2} F_{1}}{\partial u^{2}}(t, u) y_{2}(t, u)^{2} \dif t
	+ \int_{0}^{T} \frac{\partial^{3} F_{1}}{\partial u^{3}}(t, u) y_{1}(t, u)^{2} \odot y_{2}(t, u) \dif t \\
	&+ \int_{0}^{T}\left(\frac{\partial^{4} F_{1}}{\partial x^{4}}(t, u) y_{1}(t, u)^{4} + \frac{\partial F_{1}}{\partial u}(t, u) y_{4}(t, u)\right) \dif t  \ ,
\end{aligned}
\]
\newpage
\noindent where
\[
\begin{aligned}
	y_{1}(t, u) =&~ \int_{0}^{s} F_{1}(s, u) \dif s \\
	y_{2}(t, u) =&~ \int_{0}^{s} F_{2}(s, u)+\frac{\partial F_{1}}{\partial u}(s, u) y_{1}(s, u) \dif s \\
	y_{3}(t, u) =&~ \int_{0}^{s} \left( F_{3}(s, u)
	+ \frac{\partial F_{2}}{\partial u}(s, u) y_{1}(t, z) +  \frac{\partial^{2} F_{1}}{\partial u^{2}}(s, u) y_{1}(s, u)^{2}
	+ \frac{\partial F_{1}}{\partial u}(s, u) y_{2}(s, u)\right) \dif s \\
	y_{4}(t, u) =&~ \int_{0}^{s}\left(F_{4}(s, u)
	+ \frac{\partial F_{3}}{\partial x}(s, u) y_{1}(s, u)\right) \dif s 
	+ \int_{0}^{s}\left(\frac{\partial^{2} F_{2}}{\partial u^{2}}(s, u) y_{1}(s, u)^{2}
	+ \frac{\partial F_{2}}{\partial u}(s, u) y_{2}(s, u)\right) \dif s \\
	&+ \int_{0}^{s} \frac{\partial^{2} F_{1}}{\partial u^{2}}(s, u) y_{1}(s, u) \odot y_{2}(s, u) \dif s
	+ \int_{0}^{s}\left(\frac{\partial^{3} F_{1}}{\partial u^{3}}(s, u) y_{1}(s, u)^{3}
	+ \frac{\partial F_{1}}{\partial u}(s, u) y_{3}(s, u)\right) \dif s  \\
	y_{5}(t, u) =&~ \int_{0}^{t}\left(F_{5}(s, u)
	+ \frac{\partial F_{4}}{\partial u}(s, u) y_{1}(s, u)\right) \dif s \\
	&+ \int_{0}^{t}\left(\frac{\partial^{2} F_{3}}{\partial u^{2}}(s, u) y_{1}(s, u)^{2}
	+ \frac{\partial F_{3}}{\partial u}(s, u) y_{2}(s, u)
	+ \frac{\partial^{2} F_{2}}{\partial u^{2}}(s, u) y_{1}(s, u) \odot y_{2}(s, u)\right) \dif s \\
	&+ \int_{0}^{t}\left(\frac{\partial^{3} F_{2}}{\partial u^{3}}(s, u) y_{1}(s, u)^{3}
	+ \frac{\partial F_{2}}{\partial u}(s, u) y_{3}(s, u)
	+ \frac{\partial^{2} F_{1}}{\partial x^{2}}(s, u) y_{1}(s, u) \odot y_{3}(s, z)\right) \dif s \\
	&+ \int_{0}^{t} \frac{\partial^{2} F_{1}}{\partial u^{2}}(s, u) y_{2}(s, u)^{2} \dif s
	+ y^{t} \frac{\partial^{3} F_{1}}{\partial u^{3}}(s, u) y_{1}(s, u)^{2} \odot y_{2}(s, u) \dif s \\
	&+5 \int_{0}^{t}\left(\frac{\partial^{4} F_{1}}{\partial u^{4}}(s, u) y_{1}(s, u)^{4}
	+ \frac{\partial F_{1}}{\partial u}(s, u) y_{4}(s, u)\right) \dif s  \ .
\end{aligned}
\]
\newpage

(ii) For the case of $F_0 \not\equiv 0$, we have
\[
\begin{aligned}
	G_{0}(u) =&~ \int_{0}^{T} F_{0}(t, u) \dif t \\
	G_{1}(u) =&~ \int_{0}^{T} F_{1}(t, u)
	+ \frac{\partial F_{0}}{\partial u}(t, u) y_{1}(t, u) \dif t \\
	G_{2}(u) =&~ \int_{0}^{T}\left(F_{2}(t, u)
	+ \frac{\partial F_{1}}{\partial u}(t, u) y_{1}(t, u)
	+ \frac{\partial^{2} F_{0}}{\partial u^{2}}(t, u) y_{1}(t, u)^{2}
	+ \frac{\partial F_{0}}{\partial u}(t, u) y_{2}(t, u)\right) \dif t \\
	G_{3}(u) =&~ \int_{0}^{T}\left(F_{3}(t, u)
	+ \frac{\partial F_{2}}{\partial u}(t, u) y_{1}(t, u)
	+ \frac{\partial^{2} F_{1}}{\partial u^{2}}(t, u) y_{1}(t, u)^{2}
	+ \frac{\partial F_{1}}{\partial u}(t, u) y_{2}(t, u)\right) \dif t \\
	&+ \int_{0}^{T}\left( \frac{\partial^{2} F_{0}}{\partial u^{2}}(t, u) y_{1}(t, u) \odot y_{2}(t, u)
	+ \frac{\partial^{3} F_{0}}{\partial u^{3}}(t, u) y_{1}(t, u)^{3}
	+ \frac{\partial F_{0}}{\partial u}(t, u) y_{3}(t, u)\right) \dif t \\
	G_{4}(u) =&~ \int_{0}^{T}\left(F_{4}(t, u)
	+ \frac{\partial F_{3}}{\partial u}(t, u) y_{1}(t, u)\right) \dif t \\
	&+ \int_{0}^{T}\left(\frac{\partial^{2} F_{2}}{\partial u^{2}}(t, u) y_{1}(t, u)^{2}
	+ \frac{\partial F_{2}}{\partial u}(t, u) y_{2}(t, u)\right) \dif t \\
	&+ \int_{0}^{T} \frac{\partial^{2} F_{1}}{\partial x^{2}}(t, z) y_{1}(t, z) \odot y_{2}(t, z) d t \\
	&+ \int_{0}^{T}\left(\frac{\partial^{3} F_{1}}{\partial u^{3}}(t, u) y_{1}(t, u)^{3}
	+ \frac{\partial F_{1}}{\partial u}(t, u) y_{3}(t, u)
	+ \frac{\partial^{2} F_{0}}{\partial u^{2}}(t, u) y_{1}(t, u) \odot y_{3}(t, u)\right) \dif t \\
	&+ \int_{0}^{T} \frac{\partial^{2} F_{0}}{\partial u^{2}}(t, u) y_{2}(t, u)^{2} \dif t
	+ \int_{0}^{T} \frac{\partial^{3} F_{0}}{\partial u^{3}}(t, u) y_{1}(t, u)^{2} \odot y_{2}(t, u) \dif t \\
	&+ \int_{0}^{T}\left(\frac{\partial^{4} F_{0}}{\partial u^{4}}(t, u) y_{1}(t, u)^{4}
	+ \frac{\partial F_{0}}{\partial u}(t, u) y_{4}(t, u)\right) \dif t  \ ,
\end{aligned}
\]
where
\[
\begin{aligned}
	y_{1}(t, u) =&~ \int_{0}^{s} F_{1}(s, u) 
	+ \frac{\partial F_{0}}{\partial x}(s, u) y_{1}(s, u) \dif t \\
	y_{2}(t, u) =&~ \int_{0}^{s}\left(2 F_{2}(s, u)
	+ \frac{\partial F_{1}}{\partial u}(s, u) y_{1}(s, u)
	+ \frac{\partial^{2} F_{0}}{\partial u^{2}}(s, u) y_{1}(s, u)^{2}
	+ \frac{\partial F_{0}}{\partial u}(s, u) y_{2}(s, u)\right) \dif t \\
	y_{3}(t, u) =&~ \int_{0}^{s}\left( F_{3}(s, u)
	+ \frac{\partial F_{2}}{\partial u}(s, u) y_{1}(s, u)
	+ \frac{\partial^{2} F_{1}}{\partial u^{2}}(s, u) y_{1}(s, u)^{2}
	+ \frac{\partial F_{1}}{\partial u}(s, u) y_{2}(s, u) \right) \dif t \\
	&+ \int_{0}^{s}\left(\frac{\partial^{2} F_{0}}{\partial u^{2}}(s, u) y_{1}(s, u) \odot y_{2}(s, u)
	+ \frac{\partial^{3} F_{0}}{\partial u^{3}}(s, u) y_{1}(s, u)^{3}
	+ \frac{\partial F_{0}}{\partial u}(s, u) y_{3}(s, u)\right) \dif t 
\end{aligned}
\]
\newpage

\[
\begin{aligned}
	y_{4}(t, u) =&~ \int_{0}^{s}\left(F_{4}(s, u)
	+ \frac{\partial F_{3}}{\partial u}(s, u) y_{1}(s, u)\right) \dif t \\
	&+ \int_{0}^{s}\left(\frac{\partial^{2} F_{2}}{\partial u^{2}}(s, u) y_{1}(s, u)^{2}
	+ \frac{\partial F_{2}}{\partial u}(s, u) y_{2}(s, u)\right) \dif t \\
	&+ \int_{0}^{s} \frac{\partial^{2} F_{1}}{\partial u^{2}}(s, u) y_{1}(s, u) \odot y_{2}(s, u) \dif t \\
	&+ \int_{0}^{s}\left(\frac{\partial^{3} F_{1}}{\partial u^{3}}(s, u) y_{1}(s, u)^{3}
	+ \frac{\partial F_{1}}{\partial u}(s, u) y_{3}(s, u)
	+ \frac{\partial^{2} F_{0}}{\partial u^{2}}(s, u) y_{1}(s, u) \odot y_{3}(s, u)\right) \dif t \\
	&+ \int_{0}^{t} \frac{\partial^{2} F_{0}}{\partial u^{2}}(s, u) y_{2}(s, u)^{2} \dif t
	+ \int_{0}^{t} \frac{\partial^{3} F_{0}}{\partial u^{3}}(s, u) y_{1}(s, u)^{2} \odot y_{2}(s, u) \dif t \\
	&+ \int_{0}^{t}\left(\frac{\partial^{4} F_{0}}{\partial u^{4}}(s, u) y_{1}(s, u)^{4}
	+ \frac{\partial F_{0}}{\partial u}(s, u) y_{4}(s, u)\right) \dif t \ .
\end{aligned}
\]

Recall the concerned example system~\eqref{eq:example_u}, one has
\[
\left\{\begin{aligned}
	F_{1}(t,u) &= u \left(\beta^1_{1,2} + \beta^2_{1,1}\right) \sin t \cos t
	+ u \left(-\beta^1_{1,1} + \beta^2_{1,2}\right) (\sin t)^2 + u \beta^1_{1,1}  \ , \\
	G_1(u) &= \pi u \left( \beta^1_{1,1} + \beta^2_{1,2} \right)  \ .
\end{aligned}\right.
\]
It is observed that the $1^{\textrm{st}}$ component $G_1(u)$ has no positive critical points, and thus, provides no information about the bifurcation solutions once adding perturbations. Further, it is necessary to compute the higher-order components. From \textbf{Procedure 4-12}, we have
\[
\begin{aligned}
	G_{2}(u) &= \frac{\pi u}{2}\big(\pi (\beta^1_{1,1})^{2} + 2 \pi \beta^1_{1,1} \beta^2_{1,2} + \pi (\beta^2_{1,2})^{2} + \beta^1_{1,1} \beta^1_{1,2} - \beta^1_{1,1} \beta^2_{1,1} + \beta^1_{1,2} \beta^2_{1,2} - \beta^2_{1,1} \beta^2_{1,2} \\
	&\quad+ 2 \beta^1_{2,1} + 2 \beta^2_{2,2} \big)  \\
	G_{3}(u) &= \frac{1}{4} \pi u \left[ \left( \beta^1_{1,1} + 3 \beta^1_{1,6} + \beta^1_{1,8}+\beta_{1,7} + 3 \beta^2_{1,9} \right) u^{2} + 4 \left(\beta^1_{3,1} + \beta^2_{3,2} \right) \right] 
	\quad\text{with}\quad
	\beta^2_{2,2} = -\beta^1_{2,1} \\
	G_{4}(u) &= \frac{1}{4} \pi u \left[ C_1 u^{2} + 4\left( \beta^1_{4,1} + \beta^2_{4,2} \right) \right] \\
	&\text{with}\quad 
	\beta^2_{1,7} \gets \beta^1_{1,1} + 3 \beta^1_{1,6} + \beta^1_{1,8} + \beta_{1,7} + 3 \beta^2_{1,9} \beta^2_{1,7} 
	\quad\text{and}\quad
	\beta^2_{3,2} \gets -\beta^1_{3,1} - \beta^2_{3,2} + \beta^2_{3,2} \\
	G_{5}(u) &= \frac{1}{4} \pi u \left[ \left( 2 \beta^1_{1,1} + 2 \beta^1_{1,6} + \beta^1_{1,8} + \beta^2_{1,9} \right) u^{4} + C_2 u^{2} + 4\left( \beta^1_{5,1} + \beta^2_{5,2}\right) \right] \\
	&\text{with}\quad 
	\beta^2_{2,7} \gets - C_1 + \beta^2_{2,7}
	\quad\text{and}\quad
	\beta^2_{4,2} \gets - \beta^1_{3,1} - \beta^2_{3,2} + \beta^2_{4,2} \ ,\\
\end{aligned}
\]
\newpage
\noindent where
\[
\begin{aligned}
	C_1 =&~ 4 \beta^1_{1,1} \beta^1_{1,2} 
	+ 2 \beta^1_{1,1} \beta^1_{1,7}
	+ 2 \beta^1_{1,1} \beta^2_{1,8}
	+ \beta^1_{1,2} \beta^1_{1,8}
	+ 3 \beta^1_{1,2} \beta^2_{1,9}
	+ \beta^1_{1,3} \beta^1_{1,4}
	- 2 \beta^1_{1,3} \beta^2_{1,3} 
	+ \beta^1_{1,4} \beta^1_{1,5} \\
	&+ 2 \beta^1_{1,5} \beta^2_{1,5}
	+ \beta^1_{1,8} \beta^2_{1,1} 
	+ 3 \beta^1_{1,1} \beta^2_{1,9}
	- \beta^2_{1,3} \beta^2_{1,4}
	- \beta^2_{1,4} \beta^2_{1,5}
	+ 4 \beta^1_{2,1}
	+ 3 \beta^1_{2,6} 
	+ \beta^1_{2,8}
	+ \beta^2_{2,7}
	+ 3 \beta^2_{2,9}   \\
	&\quad \\
	C_2 =&~ 4 \beta^1_{1,1} (\beta^1_{1,2})^{2}
	+ 2 \beta^1_{1,1} \beta^1_{1,2} \beta^1_{1,7}
	+ 2 \beta^1_{1,1} \beta^1_{1,2} \beta^2_{1,8}
	+ 2 \beta^1_{1,1} (\beta^1_{1,3})^{2}
	+ 2 \beta^1_{1,1} \beta^1_{1,3} \beta^1_{1,5}
	- \beta^1_{1,1} \beta^1_{1,3} \beta^2_{1,4} \\
	&+ \beta^1_{1,1} (\beta^1_{1,4})^{2} 
	- \beta^1_{1,1} \beta^1_{1,4} \beta^2_{1,3}
	+ \beta^1_{1,1} \beta^1_{1,4} \beta^2_{1,5} 
	+ \beta^1_{1,1} \beta^1_{1,5} \beta^2_{1,4}
	- 2 \beta^1_{1,1} \beta^2_{1,3} \beta^2_{1,5}
	- \beta^1_{1,1} (\beta^2_{1,4})^{2} \\
	&-2 \beta^1_{1,1} (\beta^2_{1,5})^{2}
	+ (\beta^1)_{1,2}^{2} \beta^1_{1,8} 
	+ 3 (\beta^1)_{1,2}^{2} \beta^2_{1,9}
	+ \beta^1_{1,2} \beta^1_{1,3} \beta^1_{1,4}
	+ 2 \beta^1_{1,2} \beta^1_{1,4} \beta^1_{1,5}
	+ 4 \beta^1_{1,2} \beta^1_{1,5} \beta^2_{1,5} \\
	&+ \beta^1_{1,2} \beta^1_{1,8} \beta^2_{1,1}
	+ 3 \beta^1_{1,2} \beta^2_{1,1} \beta^2_{1,9} 
	- \beta^1_{1,2} \beta^2_{1,4} \beta^2_{1,5} 
	+ 2 \beta^1_{1,3} \beta^2_{1,1} \beta^2_{1,3}
	+ \beta^1_{1,4} \beta^1_{1,5} \beta^2_{1,1} 
	+2 \beta^1_{1,5} \beta^2_{1,1} \beta^2_{1,5} \\
	&+ \beta^2_{1,1} \beta^2_{1,3} \beta^2_{1,4}
	+ 4 \beta^1_{1,1} \beta^1_{2,2}
	+ 2 \beta^1_{1,1} \beta^1_{2,7}
	+ 2 \beta^1_{1,1} \beta^2_{2,8}
	+ 4 \beta^1_{1,2} \beta^1_{2,1} 
	+ \beta^1_{1,2} \beta^1_{2,8}
	+ 3 \beta^1_{1,2} \beta^2_{2,9} \\
	&+ \beta^1_{1,3} \beta^1_{2,4}
	- 2 \beta^1_{1,3} \beta^2_{2,3}
	+ \beta^1_{1,4} \beta^1_{2,3}
	+ \beta^1_{1,4} \beta^1_{2,5}
	+ \beta^1_{1,5} \beta^1_{2,4} 
	+ 2 \beta^1_{1,5} \beta^2_{2,5}
	+ 2 \beta^1_{1,7} \beta^1_{2,1} 
	+ \beta^1_{1,8} \beta^1_{2,2} \\
	&+ \beta^1_{1,8} \beta^2_{2,1}
	+ 2 \beta^1_{2,1} \beta^2_{1,8}
	+ 3 \beta^1_{2,2} \beta^2_{1,9}  
	- 2 \beta^1_{2,3} \beta^2_{1,3}
	+ 2 \beta^1_{2,5} \beta^2_{1,5}
	+ \beta^1_{2,8} \beta^2_{1,1}
	+ 3 \beta^2_{1,1} \beta^2_{2,9}
	- \beta^2_{1,3} \beta^2_{2,4}   \\
	&- \beta^2_{1,4} \beta^2_{2,3}
	- \beta^2_{1,4} \beta^2_{2,5} 
	- \beta^2_{1,5} \beta^2_{2,4}
	+ 3 \beta^2_{1,9} \beta^2_{2,1}
	+ 4 \beta^1_{3,1}
	+ 3 \beta^1_{3,6}
	+ \beta^1_{3,8}
	+ \beta^2_{3,7}
	+ 3 \beta^2_{3,9} \ .
\end{aligned}
\]
It is observed that the $5^{\textrm{th}}$ component $G_5(u)$ has at most three positive critical points, which provides support for the existence of the upper bound of $H(n)$ in Corollary~\ref{coroll:example}. This completes the proof. $\hfill\square$

\section{About the Post-synaptic Computations according to Eq.~\eqref{eq:post}} \label{app:post}
Table~\ref{tab:gradients} has listed the post-synaptic computations of conventional surrogate gradients~\citep{li2021:surrogate}, SLAYER~\citep{shrestha2018:slayer}, and our proposed StocSNN, and Figure~\ref{fig:gradients} illustrates the feed-forward and back-propagation computations.

The post-synaptic derivative of Eq.~\eqref{eq:bp_regression} indicates the remediation of the discontinuous and non-differential firing phase led by deterministic feed-forward computations. This derivative, i.e., the derivative of firing function $\textrm{Heaviside}(u - u_{\textrm{reset}})$, is identified as a Dirac-delta function~\footnote{Or equally, the spike sequence is formulated by $\mathbf{I}_j(t) = \sum_{\text{firing}} \delta \left( t-t_j^{\text{firing}} \right)$, where $t_j^{\text{firing}}$ is the spike time of the $j^{\textrm{th}}$ input and $\delta(t)$ is a corresponding Dirac-delta function.}, 
\[
\textrm{Heaviside}(x) = \left\{~ \begin{aligned}
	1 \ ,& \quad x\ \geq 0 \ , \\
	0 \ ,& \quad x <0 \ ,
\end{aligned}\right.
\quad\text{and}\quad
\textrm{Heaviside}'(x) = \left\{~ \begin{aligned}
	\infty \ ,& \quad x = 0 \ , \\
	0 \ ,& \quad x \neq 0 \ ,
\end{aligned}\right.
\]
which is zero almost everywhere except for the threshold that grows to infinity, as shown by the red solid curve in Figure~\ref{fig:gradients}(b)(d). As a consequence, the gradient descent 
\[
\mathbf{W}^l \leftarrow \mathbf{W}^l - \eta \frac{\partial E }{\partial \mathbf{W}^l}
\]
substituted into Eq.~\eqref{eq:w} either freezes the connection weights or updates the connection weights to infinity. In conventional SNNs training algorithms, the post-synaptic derivative is approximated by a smooth surrogate function, such as the rectangular, triangular, and hyperbolic tangent functions~\citep{li2021:surrogate}. It is evident that using surrogate gradients inevitably results in a gap between the feed-forward and back-propagation procedures. Thus, surrogate gradients are asymptotic and biased calculators. 

SLAYER~\citep{shrestha2018:slayer} seeks an alternative gradient to circumvent this issue. As shown in Figure~\ref{fig:gradients}(c)(d), SLAYER utilizes a probability density function to indicate the change of a spiking neuron state such that the proposed probability density function can estimate the expected post-synaptic derivative (also known as surrogate gradient). It has been proved that using probability density functions leads to an asymptotic but unbiased estimator.

\vspace{0.2cm}
\noindent\textbf{Finishing the proof of Theorem~\ref{thm:gradients}}. Here, we leverage the post-synaptic derivative from the perspective of energy back-propagation. It is observed that the pre-synapse receives $f_{\textrm{agg}}(\boldsymbol{s}^{(l-1)}(t))$ at time $t$, and then the post-synapse fires spikes according to the excitation probability $p(u)$. In the whole procedure, the concerned neuron receives the pre-synaptic signals $f_{\textrm{agg}}(\boldsymbol{s}^{(l-1)}(t))$, consumes the integration operations, and fires the output according to $p(u) $; the former two correspond to the pre-synaptic derivative and the latter results in the post-synaptic derivative. Further, we obtain the energy rate as
\[
\frac{\partial p(u) u_{\textrm{firing}}}{\partial u(t) } = \frac{\partial p(u) }{\partial u(t) } u_{\textrm{firing}} \ . 
\]
Inspired by this recognition, we can replace the binary spike $s(t)$ in Eq.~\eqref{eq:bp_regression} by the corresponding excitation probability $p(u)$, thus approximating the post-synaptic derivatives using the element-wise gradients, that is,
\[
\frac{\partial \boldsymbol{s}^{(l)}(t)}{\partial\boldsymbol{u}^{(l)}(t) } \leftarrow \frac{\partial \boldsymbol{p}^{(l)}(t)}{\partial\boldsymbol{u}^{(l)}(t) } \ .
\]

From the perspective of random algorithms, Eq.~\eqref{eq:post} derives a \textbf{non-asymptotic and unbiased} estimator. In Figure~\ref{fig:gradients}(e), the stochastic spiking neuron randomly generates spikes when the membrane potential enter in the red region (possible firing region). Thus, the post-synaptic derivative becomes a pseudo-step function that consists of a Dirac delta function on $[0,u_{\theta}]$ and an expectation derivative on $[u_{\theta}, u_{\textrm{firing}}]$, as shown by the red dotted curve in Figure~\ref{fig:gradients}(f). Thus, our proposed derivative is \textbf{non-asymptotic} as $u_{\theta} \to 0$. 
\begin{itemize}
	\item There is only a small gap between the feed-forward and back-propagation computations. This gap is led by the Dirac delta function defined on $[0,u_{\theta}]$ and controlled by $p_{\theta}$ (corresponding to $u_{\theta}$ and as $u_{\theta} \to 0$). 
	\item Since $\boldsymbol{s}^{(l)}$ is sampled from the Bernoulli distribution with random variable $\boldsymbol{p}^{(l)}$, it holds $\mathbb{E}[\boldsymbol{s}^{(l)}] = \boldsymbol{p}^{(l)}$. Let $\delta \mu$ be a small perturbation that affects the membrane potentials so that $|u - u_{\textrm{firing}}| \leq u_{\theta}$ leads to a change of the spiking neuron state. Thus, we have
	\[
	\mathbb{E} \left[ \frac{\delta \boldsymbol{s}^{(l)}(t)}{\delta \boldsymbol{u}^{(l)}(t)} \right] = \frac{\delta \boldsymbol{p}^{(l)}(t)}{\delta \boldsymbol{u}^{(l)}(t)} \ ,
	\]
	where
	\[
	\left\{~\begin{aligned}
		\delta \boldsymbol{s}^{(l)}(t) &= \boldsymbol{s}^{(l)}(t+\delta t) - \boldsymbol{s}^{(l)}(t)  \ , \\
		\delta \boldsymbol{p}^{(l)}(t) &= \boldsymbol{p}^{(l)}(t + \delta t) - \boldsymbol{p}^{(l)}(t) \ ,
	\end{aligned}\right.
	\]
	and $\delta \boldsymbol{u}^{(l)}(t) \to 0$ as $\delta t \to 0$. Thus, we have
	\[
	\mathbb{E}\left[\frac{\partial \boldsymbol{s}^{(l)}(t)}{\partial\boldsymbol{u}^{(l)}(t) } \right] = \mathbb{E}\left[ \lim_{u_{\theta} \to 0 \atop \delta t \to 0} \frac{\delta \boldsymbol{s}^{(l)}}{\delta \boldsymbol{u}^{(l)}} \right]
	= \lim_{u_{\theta} \to 0 \atop \delta t \to 0} \frac{\delta \boldsymbol{p}^{(l)}(t)}{\delta \boldsymbol{u}^{(l)}(t)}
	= \frac{\partial \boldsymbol{p}^{(l)}(t)}{\partial\boldsymbol{u}^{(l)}(t) } \ ,
	\]
	which implies that Eq.~\eqref{eq:post} is an \textbf{unbiased} estimator for the post-synaptic derivative.
\end{itemize}
This completes the proof. $\hfill\square$

\section{Full Proofs of Theorem~\ref{thm:ua_stoc} and Theorem~\ref{thm:computational_power}}  \label{app:proof_approximation}
The proof idea of these two theorems has its spring in the timing representation of the stochastic spiking neuron model. There exists an inverse transformation $\phi$ between the spike sequences $\mathbf{X} \in \{0,1\}^{M \times T}$ and its timing sequences $\mathbf{T}_X (\mathbb{R}^+)^{M \times T} $. Thus, the universal approximation issue presented in Theorem~\ref{thm:ua_stoc} is equivalent to a new problem that collects expressive functions $f$ to approximate $g' \in \mathcal{C}^r(K,\mathbb{R})$. Moreover, the approximation issue presented in Theorem~\ref{thm:computational_power} equals to finding an apposite function $g'$ such that
\[
\sup_{\phi} \| f(\phi(\mathbf{X})) - g'(\phi(\mathbf{X})) \|_2 < \epsilon 
\]
with arbitrarily high possibility. Here, we define an element distinctness function $g_{\textrm{EDF}} : (\mathbb{R}^+)^M \to \{0, 1\}$ by
\[
g_{\textrm{EDF}} (\mathbf{T}_X) = (g_1, \dots, g_M) \ , \quad
g_k = \left\{\begin{aligned}
	1~~,  &\quad \text{if $\mathbf{T}_{ki} = \mathbf{T}_{kj}$ for $i \neq j$} \ ; \\
	0~~,  &\quad \text{if $| \mathbf{T}_{ki} - \mathbf{T}_{kj} | \geq c \Delta t$ for $i \neq j$} \ ; \\
	p_{\theta}, &\quad \text{otherwise} \ ,
\end{aligned}\right. \quad\text{for $k \in [M]$} \ ,
\]
where $c$ is a scaling constant and $\Delta t$ is a timing threshold. It is obvious that $g_{\textrm{EDF}}$ is an apposite conversion between the rate-based and timing-based encoding, mentioned in Section~\ref{sec:inside}. Thus, the universal approximation issue in Theorem~\ref{thm:ua_stoc} is equivalent to another problem of universally approximating the timing sequences $\mathbf{T}_X \in (\mathbb{R}^+)^{M \times T}$, that is, $g_{\textrm{EDF}} $ is a component of $\phi$. Therefore, it is obvious that approximating $g'$ is a sub-problem of proving Theorem~\ref{thm:computational_power} when one regards $g'$ as a component of $g_{\textrm{EDF}} : (\mathbb{R}^+)^{M \times T} \to \{0, 1\}^M$.

\vspace{0.2cm}
\noindent\textbf{Finishing the proof of Theorem~\ref{thm:ua_stoc}.} Altering to the line of the thought above, we can prove that $g_{\textrm{EDF}}$ is an invertible conversion between $\mathbf{X} \in \{0,1\}^{M \times T}$ and its timing sequences $\mathbf{T}_X (\mathbb{R}^+)^{M \times T} $. Thus, given an apposite $p(u)$, we can obtain that the set of functions expressed by stochastic spiking neurons is dense in $\mathcal{C}^r(K, \mathbb{R})$ where $K$ is a bounded set of $\mathbb{R}^M$. Besides, $\mathcal{C}^r(K,\mathbb{R})$ is dense in $\mathcal{C}^0(K,\mathbb{R})$. According to the transitivity of dense operations, we can finish this proof. $\hfill\square$

\vspace{0.2cm}
\noindent\textbf{Finishing the proof of Theorem~\ref{thm:computational_power}.} Part (i) of Theorem~\ref{thm:computational_power} is self-evident. Notice that $p_{\theta}$ indicates a possibility threshold detailed in Section~\ref{sec:stoc}. Hence, such a function $g_{\textrm{EDF}}$ can be approximated well by a SNN with only one hidden stochastic spiking neuron since the input spike arrives with a temporal distance between $0$ and $c$. 

Part (ii) of Theorem~\ref{thm:computational_power} follows the results of~\citep{maass1996lower}. 

For part (iii) of Theorem~\ref{thm:computational_power}, we consider some set $K_{M-1} \subseteq \mathbb{R}^+$ of the cardinality $M-1$. Let $c > 0$ be sufficiently large to ensure that elements in set $c \cdot K_{M-1}$ have pairwise distances greater than 2. Let $K^*_{M-1}$ be a set with cardinality $M-1$, in which the pairwise distance of elements are greater than $\max(c \cdot K_{M-1}) + 2$. So it suffices to prove that the concerned artificial neural network can partition arbitrary $M$ elements of $K^*_{M-1} \cup c \cdot K_{M-1} $ differently. Let $f_{\textrm{ANN}}$ denote the concerned artificial neural network with $(M-6)/2$ hidden sigmoidal neurons, and $f_{\textrm{ANN}}^{\textrm{c}}$ is a variant of $f_{\textrm{ANN}}$ where all weights connects to input variables are multiplied with $c$. Hence, when one assigns a suitable set of $M-1$ pairwise elements from $K^*_{M-1} \cup c \cdot K_{M-1} $ to the last $M-1$ input variables, $f_{\textrm{ANN}}^{\textrm{c}}$ can approximate any function from $K_{M-1}$ to $\{0,1\}$. In order to shatter arbitrary set $K_{M-1}$ using $f_{\textrm{ANN}}$, there are at least $m+1$ anchor points, that is, hidden spiking neurons. From the results of~\citep{sontag1997shattering} that yield an upper bound of $2(m+1)+1$ for the maximal number $M \in \mathbb{N}^+$ such that every set of $M$ different inputs can be shattered by an artificial neural network with sigmoidal activation of $m$ hidden neurons. Thus, we have 
\[
M - 1 \leq 2((m+1)+1) + 1 \ ,
\quad\text{or equally}\quad
m \geq (M-6)/2 \ ,
\]
which completes this proof. $\hfill\square$

\vspace{0.2cm}
\noindent\textbf{Remark:} Due to the generality of Sontag's results~\citep{sontag1997shattering}, part (iii) of Theorem~\ref{thm:computational_power} (i.e., the lower bound of $(M-6)/2)$ is also valid for all sigmoid-like activations, even if $f_{\textrm{ANN}}$ employs a Heaviside-like function besides sigmoidal activations, which coincides with part (ii) of Theorem~\ref{thm:computational_power}.

\section{Full Proof for Theorem~\ref{thm:generalization_Stoc}} \label{app:proof_generalization_Stoc}
Here, we complete the proof of Theorem~\ref{thm:generalization_Stoc}. Provided Eq.~\eqref{eq:general_radermacher}, the general Rademacher complexity is relevant to not only training samples but also the  excitation probability threshold.

Motivated by the techniques given by Bartlett and Mendelson~\citep{bartlett2002rademacher}, it obviously holds
\[
E(f) \leq \widehat{E}(f) +  \underbrace{ \sup_{\boldsymbol{w}\in\mathcal{W}} \left[ E(f) - \hat{E}(f) \right] }_{R(S_n,\mathbf{P}) } \ .
\]
Let $S'_n$ denote the sample set that the $i^{\textrm{th}}$ sample $(\mathbf{X}_i, y_i)$ is replaced by $(\mathbf{X}'_i, y'_i)$, and correspondingly $\mathbf{P}'$ is the possibility matrix that the $i^{\textrm{th}}$ row vector $\boldsymbol{p}_i$ is replaced by $\boldsymbol{p}'_i$, for $i \in [n]$. For the loss function $\mathscr{L}$ bounded by $C>0$, that is, $|\mathscr{L}| \leq C$, one has
\[
\left\{~\begin{aligned}
	&| R(S_n,\mathbf{P}) - R(S'_n,\mathbf{P}) |  \leq C/n \ , \\
	&| R(S_n, \mathbf{P}) - R(S_n, \mathbf{P}') | \leq C/n \ .
\end{aligned}\right.
\]
From McDiarmid's inequality~\citep{mcdiarmid1989method}, with probability at least $1-\delta$, the following holds
\[
R(S_n, \mathbf{P}) \leq \mathbb{E}_{S_n \in \mathcal{D}, \mathbf{P}} \left[ R(S_n, \mathbf{P}) \right] + C\sqrt{\frac{\ln(2/\delta)}{n}} \ .
\]
It is observed that
\[
R(S_n, \mathbf{P}) = \sup_{\boldsymbol{w} \in \mathcal{W}} \mathbb{E}_{\tilde{S}_n \in \mathcal{D}, \tilde{\mathbf{P}}} \left[ \widehat{E}(f;\tilde{S}_n, \tilde{\mathbf{P}}) - \widehat{E}(f; S_n, \mathbf{P}) \right] \ ,
\]
where $\tilde{S}_n$ is another collection drawn from $\mathcal{D}$ as well as $\tilde{\mathbf{P}}$. Thus, we have
\[
\begin{aligned}
	\mathbb{E}_{S_n \in \mathcal{D}, \mathbf{P}} \left[ R(S_n, \mathbf{P}) \right] 
	&~\leq \mathbb{E} \left[ \sup_{\boldsymbol{w} \in \mathcal{W}} \left[ \widehat{E}(f;\tilde{S}_n, \tilde{\mathbf{P}}) - \widehat{E}(f; S_n, \mathbf{P}) \right] \right] \\
	&~= \mathbb{E} \left[ \sup_{\boldsymbol{w} \in \mathcal{W}} \frac{1}{n} \sum_{i=1}^n \left[ h(\boldsymbol{w}, \tilde{\mathbf{X}}_i, \tilde{y}_i, \tilde{\boldsymbol{p}}_i) - h(\boldsymbol{w}, \mathbf{X}_i, y_i, \boldsymbol{p}_i) \right] \right] \\
	&~\leq 2 \mathbb{E} \left[ \sup_{\boldsymbol{w} \in \mathcal{W}} \frac{1}{n} \sum_{i=1}^n \epsilon_i h(\boldsymbol{w}, \mathbf{X}_i, y_i, \boldsymbol{p}_i) \right] \\
	&~=2 \mathfrak{R}_n( \mathscr{L} \circ \mathcal{F}_{\mathcal{W}} ) \ ,
\end{aligned}
\]
which completes the proof of Eq.~\eqref{eq:Rademacher_a}. By applying McDiarmid's inequality to $\widehat{\mathfrak{R}}_n(\mathscr{L} \circ \mathcal{F}_{\mathcal{W}}, S_n, \mathbf{P})$, we have
\[
\mathfrak{R}_n( \mathscr{L} \circ \mathcal{F}_{\mathcal{W}} ) \leq \widehat{\mathfrak{R}}_n(\mathscr{L} \circ \mathcal{F}_{\mathcal{W}}, S_n, \mathbf{P}) + C \sqrt{\frac{\ln(2/\delta)}{n}} \ ,
\]
which completes the proof of Eq.~\eqref{eq:Rademacher_b}. Therefore, this theorem follows as desired. $\hfill\square$

\section{Full Proof for Theorem~\ref{thm:estimation_for_deep_layer}} \label{app:proof_estimation_for_deep_layer}
Here, we are going to complete the proof of Theorem~\ref{thm:estimation_for_deep_layer}. Before that~\footnote{For convenience, we here omit the superscripts and subscripts as possible.}, we introduce a probability indicate as follows
\[
s_j(t) \sim \text{Bernoulli}(p_j(t))_j \quad\text{with}\quad p_j(t) \sim \textrm{Eq.}~\eqref{eq:p} \ .
\]
Thus, we can convert $s^{(l)}(t)$ of Eq.~\eqref{eq:StocSNN} into 
\[
\boldsymbol{s}^l(t) = \boldsymbol{I}^l(t) = \left( \boldsymbol{I}^l_1(t), \cdots, \boldsymbol{I}^l_{d_l}(t) \right)^{\top} \ .
\]

We begin the proof of Theorem~\ref{thm:estimation_for_deep_layer} with several useful lemmas as follows.
\begin{lemma}[\citep{ledoux1991probability}] \label{lemma:dropout_1}
	Let $\mathcal{F}$ denote a bounded function space from $\mathcal{X}$ to $\mathcal{Y}$, and $\phi: \mathbb{R} \times \mathbb{R} \to \mathbb{R}$ is a Lipschitz function with constant $C_n>0$ and $\phi(0)=0$. For $(\mathbf{X}_i,y_i) \in \mathcal{X} \times \mathcal{Y}$ $(i \in [n])$, we have
	\[
	\mathbb{E}_{\epsilon_i, i \in [n]} \left[ \sup_{f \in \mathcal{F}} \frac{1}{n} \sum_{i=1}^n \epsilon_i \phi( f(\mathbf{X}_i), y_i ) \right] 
	\leq
	C_n~ \mathbb{E}_{\epsilon_i, i \in[n]} \left[ \sup_{f \in \mathcal{F}} \frac{1}{n} \sum_{i=1}^n \epsilon_i f(\mathbf{X}_i) \right] \ .    
	\]
\end{lemma}

\begin{lemma} [\citep{gao2016dropout}] \label{lemma:dropout_2}
	Suppose that $\mathscr{L}(\cdot, y)$ is a Lipschitz function with constant $C_n>0$, then we have
	\[
	\mathfrak{R}_n (\mathscr{L} \circ \mathcal{F}_{\mathcal{W}}) \leq C_n~ \mathfrak{R}_n( \mathcal{F}_{\mathcal{W}} ) \ .
	\]
\end{lemma}
Lemma~\ref{lemma:dropout_2} provides an intuitive way for estimating $\mathfrak{R}_n (\mathscr{L} \circ \mathcal{F}_{\mathcal{W}})$ from $\mathfrak{R}_n (\mathcal{F}_{\mathcal{W}})$.

\begin{lemma}  \label{lemma:dropout_3}
	Let $\boldsymbol{\alpha}$, $\boldsymbol{\alpha}'$, and $\boldsymbol{\alpha}''$ be three $L$-dimensional indicator vectors, of which the $l$-th element $\boldsymbol{\alpha}_l$, $\boldsymbol{\alpha}'_l$, and $\boldsymbol{\alpha}''_l$ are drawn i.i.d. from Bernoulli distribution $\textrm{Bernoulli}(p^{(l)})$ for $l \in [L]$. Then for $\mathbf{X} \in \mathcal{X}$, one has
	\[
	\begin{aligned}
		& \mathbb{E}_{\boldsymbol{\alpha}} \left[ \| \mathbf{X} \odot \boldsymbol{\alpha} \|_2 \right] \leq p_0^{1/2} \cdot \|\mathbf{X}\|_2 \ , \\
		& \mathbb{E}_{\boldsymbol{\alpha},\boldsymbol{\alpha}'} \left[ \| \mathbf{X} \odot \boldsymbol{\alpha} \odot \boldsymbol{\alpha}' \|_2 \right] \leq p_0 \cdot \|\mathbf{X}\|_2 \ , \\
		& \mathbb{E}_{\boldsymbol{\alpha},\boldsymbol{\alpha}'} \left[ \prod_{l\in[L]} \alpha_l \left\| \mathbf{X} \odot \boldsymbol{\alpha}' \right\|_2 \right] \leq  p_0^{(L+1)/2} \| \mathbf{X}\|_2  \ , \\
		& \mathbb{E}_{\boldsymbol{\alpha},\boldsymbol{\alpha}',\boldsymbol{\alpha}''}     \left[ \prod_{l\in[L]} \alpha_l \left\| \mathbf{X} \odot \boldsymbol{\alpha}' \odot \boldsymbol{\alpha}'' \right\| \right] \leq p_0^{(L+2)/2} \| \mathbf{X}\|_2  \ , \\
	\end{aligned}
	\]
	where $p_0 = \max_{l\in[L]} \{ p^{(l)} \}$.
\end{lemma}

\vspace{0.2cm}
\noindent\textbf{Finishing the proof of Theorem~\ref{thm:estimation_for_deep_layer}.} We first show the estimation results for one-hidden-layer StocSNN, i.e.,
\[
\mathfrak{R}^{\textrm{one}}_n(\mathcal{F}_{\mathcal{W}}) \leq \frac{C_n C_X C_w^2 ~p_{\textrm{max}}}{\sqrt{n}} \ ,
\]
if $u_{\textrm{reset}} = 0$, $\|\boldsymbol{w} \| \leq \| \mathbf{W} \| \leq C_w$ for $\mathbf{W} \in \mathcal{W}$, and $\|\boldsymbol{x} \| \leq \| \mathbf{X} \| \leq C_X$ for $\mathbf{X} \in \mathcal{X}$, where $\mathcal{F}_{\mathcal{W}}^{\textrm{one}}$ denotes the function space of StocSNN with one hidden layer. For $i\in[n]$, we define an element-wise vector $\Delta \boldsymbol{u} $ as follows
\[
\Delta \boldsymbol{u}  = \left( \boldsymbol{1} - \boldsymbol{s}  \right) \odot \left( \tau \boldsymbol{u}  + \boldsymbol{w} \boldsymbol{x}  \right)  + \boldsymbol{s}  \cdot u_{\text{reset}} \ ,
\]	
where $\boldsymbol{x} $, $\boldsymbol{u} $, and $\boldsymbol{s} $ is the input, membrane, and spike vectors, respectively. According to $u_{\textrm{reset}} = 0$ and $\boldsymbol{s}^{(i)} \in \{0,1\}^m$ that is drawn i.i.d. from Bernoulli($\boldsymbol{p}^{(i)} $) where $m$ denotes the number of hidden spiking neurons and the superscript $i$ denotes the $i^{\textrm{th}}$ instance, we have
\begin{equation}  \label{eq:hat}
	\begin{aligned}
		\widehat{\mathfrak{R}}^{\textrm{one}}_n(\mathscr{L} \circ \mathcal{F}_{\mathcal{W}}, S_n, \mathbf{P})
		& = \frac{1}{n} \mathbb{E}_{\epsilon_i, i \in [n]} \left[ \sup_{\boldsymbol{w} \in \mathcal{W}} \left\langle \boldsymbol{w}, \sum_{i=1}^n \epsilon_i \boldsymbol{s}^{(i)}   \odot \Delta\boldsymbol{u}^{(i)}   \right\rangle \right] \\
		&\leq C_w \mathbb{E}_{\epsilon_i, i \in [n]} \left[ \sup_{\boldsymbol{w} \in \mathcal{W}} \left\langle \frac{\boldsymbol{w}}{\|\boldsymbol{w}\|_1}, \frac{1}{n} \sum_{i=1}^n \epsilon_i \boldsymbol{s}^{(i)}   \odot \Delta\boldsymbol{u}^{(i)}   \right\rangle \right] \\
		&\leq C_w \mathbb{E}_{\epsilon_i, i \in [n]} \left[ \sup_{\boldsymbol{w} \in \mathcal{W}} \left\langle \frac{\boldsymbol{w}}{\|\boldsymbol{w}\|_1}, \frac{1}{n} \sum_{i=1}^n \epsilon_i \boldsymbol{s}^{(i)}   \odot \left( \boldsymbol{1} - \boldsymbol{s}^{(i)}   \right) \odot \left( \tau \boldsymbol{u}  + \boldsymbol{w} \boldsymbol{x}^{(i)}   \right) \right\rangle \right] \\
		&\leq \frac{C_nC_w}{n} \mathbb{E}_{\epsilon_i, i \in [n]} \left[ \sup_{\boldsymbol{w} \in \mathcal{W}} \sum_{i=1}^n \epsilon_i \left\langle \boldsymbol{w} \odot \boldsymbol{s}^{(i)}  , \boldsymbol{x}^{(i)}   \odot \boldsymbol{s}^{(i)}  \right\rangle \right]  \ ,
	\end{aligned}
\end{equation}
where $0 \leq C_n, C_w$ and the last inequality hold from~\citep[Lemma 1]{gao2016dropout} and Lemma~\ref{lemma:dropout_1}. Since $\mathbb{E} [ \epsilon_i\epsilon_k ] = 0$ for $i \neq k$ and $\mathbb{E} [ \epsilon_i\epsilon_i ] = 1$ for $i \in [n]$, we have
\begin{equation}  \label{eq:w}
	\mathbb{E}_{\epsilon_i, i \in [n]} \left[ \sup_{\boldsymbol{w} \in \mathcal{W}} \sum_{i=1}^n \epsilon_i \left\langle \boldsymbol{w} \odot \boldsymbol{s}^{(i)}  , \boldsymbol{x}^{(i)}   \odot \boldsymbol{s}^{(i)}   \right\rangle \right]
	\leq
	C_w \left( \sum_{i=1}^n p_{\textrm{max}}^{(i)}  \left\langle \boldsymbol{x}^{(i)}   \odot \boldsymbol{s}^{(i)}  , \boldsymbol{x}^{(i)}   \odot \boldsymbol{s}^{(i)}   \right\rangle  \right)^{1/2} \ ,
\end{equation}
where $p_{\textrm{max}}^{(i)}  = 1 - \max\{ p^{(i)}, p_{\theta}\} $ denotes the possibility threshold of the $i^{\textrm{th}}$ instance. According to Lemma~\ref{lemma:dropout_3}, we have
\[
\widehat{\mathfrak{R}}^{\textrm{one}}_n(\mathscr{L} \circ \mathcal{F}_{\mathcal{W}}, S_n, \mathbf{P})
\leq
\frac{C_nC_w}{n} C_w \left( \sum_{i=1}^n p_i  \left\langle \boldsymbol{x}^{(i)}   \odot \boldsymbol{s}^{(i)}  , \boldsymbol{x}^{(i)}   \odot \boldsymbol{s}^{(i)}   \right\rangle  \right)^{1/2}
\leq \frac{C_n C_X C_w^2 ~p_{\textrm{max}}}{\sqrt{n}} \ ,
\]
where $p_{\textrm{max}} = \max_{i\in[n]} \{ p_{\textrm{max}}^{(i)} \}$. Notice that the above formula holds according to the sampling of probability indicate $\boldsymbol{I}_j^l(t)$. If one sets $p_{\theta} = 1$ (corresponding to $p_{\textrm{max}} = 1$), that is, any membrane potential may generate spikes, then our bound can be relaxed to the conventional studies in artificial neural networks~\citep{wan2013:dropout}. And $p_{\theta} = 0$ makes the aforementioned bound meaningless.

Next, we extend the above result to the deep StocSNN. The proof processes for single-layer and multi-layer StocSNN is almost the same. The only difference is that $\widehat{\mathfrak{R}}^{\textrm{deep}}_n(\mathscr{L} \circ \mathcal{F}_{\mathcal{W}}, S_n, \mathbf{P})$ of deep StocSNN is unfolded layer by layer. By exploiting Eqs.~\eqref{eq:hat} and~\eqref{eq:w}, one has
\[
\left\{\begin{aligned}
	\widehat{\mathfrak{R}}^{\textrm{deep}}_n(\mathscr{L} &\circ \mathcal{F}_{\mathcal{W}}, S_n, \mathbf{P}) \leq \frac{C_nC_L}{n} \mathbb{E}_{\epsilon_i, i \in [n]} \left[ \sup_{\boldsymbol{w}^L} \sum_{i=1}^n \epsilon_i \left\langle \boldsymbol{w}^L \odot \boldsymbol{s}^{(l)}, \boldsymbol{s}^{(l-1)} \odot \boldsymbol{s}^{(l)} \right\rangle \right]  \ , \\
	\mathbb{E}_{\epsilon_i, i \in [n]} &\left[ \sup_{\boldsymbol{w}^{l+1}} \sum_{i=1}^n \epsilon_i \left\langle \boldsymbol{w}^{(l+1)} \odot \boldsymbol{s}^{(l+1,i)}, \boldsymbol{s}^{(l,i)} \odot \boldsymbol{s}^{(l+1,i)} \right\rangle \right] \\
	&\leq
	C_n C_{l+1} \mathbb{E}_{\epsilon_i, i \in [n]} \left[ \sup_{\boldsymbol{w}^{l}} \sum_{i=1}^n \epsilon_i ~p_{\textrm{max}}^{(l+1,i)} \left\langle \boldsymbol{w}^{(l)} \odot \boldsymbol{s}^{(l,i)}, \boldsymbol{s}^{(l-1,i)} \odot \boldsymbol{s}^{(l,i)} \right\rangle \right]  \\
	&\leq
	C_n C_{l+1}~\max_{i}\{ p_{\textrm{max}}^{(l+1,i)} \} ~\mathbb{E}_{\epsilon_i, i \in [n]} \left[ \sup_{\boldsymbol{w}^{l}} \sum_{i=1}^n \epsilon_i \left\langle \boldsymbol{w}^{l} \odot \boldsymbol{s}^{(l,i)}, \boldsymbol{s}^{(l-1,i)} \odot \boldsymbol{s}^{(l,i)} \right\rangle \right] \ , \\
	\mathbb{E}_{\epsilon_i, i \in [n]} &\left[ \sup_{\boldsymbol{w}^1} \sum_{i=1}^n \epsilon_i \left\langle \boldsymbol{w}^1 \odot \boldsymbol{s}^{(1,i)}, \boldsymbol{x}  \odot \boldsymbol{s}^{(1,i)} \right\rangle \right]
	\leq
	C_1 \left( \sum_{i=1}^n p_{\textrm{max}}^{(1,i)} \left\langle \boldsymbol{x}  \odot \boldsymbol{s}^{(1,i)}, \boldsymbol{x}  \odot \boldsymbol{s}^{(1,i)} \right\rangle  \right)^{1/2} \ ,
\end{aligned}\right.
\]
where $p_{\textrm{max}}^{(l,i)}  = 1 - \max\{ p^{(l,i)}, p_{\theta}\} $ in which the superscripts $l$ and $i$ denote the layer and instance index, respectively. Provided $p_{\textrm{max}} = \max_{i\in[n],l \in [L]} \{ p_{\textrm{max}}^{(l,i)} \}$, this proof follows as desired. $\hfill\square$

\newpage

\bibliography{JMrefs}
\bibliographystyle{plain}

\end{document}